\documentclass[nohyperref]{article}
\usepackage{graphicx}
\usepackage{tkz-tab}
\usepackage{pgfplots}
\usepackage{verbatim}
\usetikzlibrary{positioning,decorations.pathreplacing,decorations.markings,calc,arrows,hobby,backgrounds,patterns}
\usepackage{microtype}
\usepackage{graphicx}
\usepackage{subcaption}
\usepackage{adjustbox}
\usepackage{dsfont}
\usepackage{booktabs} 
\usepackage{hyperref}
\usepackage[hyphenbreaks]{breakurl}
\usepackage{tablefootnote}

\usepackage[accepted]{icml2023}

\usepackage{amsmath}
\usepackage{amssymb}
\usepackage{mathtools}
\usepackage{amsthm}
\usepackage{adjustbox}
\usepackage{tikz}
\usetikzlibrary{positioning,decorations.pathreplacing,decorations.markings,calc,arrows,hobby,backgrounds,patterns,calligraphy}

\usepackage[capitalize,noabbrev]{cleveref}

\theoremstyle{plain}
\newtheorem{theorem}{Theorem}[section]
\newtheorem{proposition}[theorem]{Proposition}

\theoremstyle{definition}

\theoremstyle{remark}

\usepackage{graphicx}
\usepackage{stackengine}

\usepackage{sepfootnotes}
\usepackage{hyperref}

\DeclareFontFamily{U}{mathx}{\hyphenchar\font45}
\DeclareFontShape{U}{mathx}{m}{n}{%
<-6> mathx5
<6-7> mathx6
<7-8> mathx7
<8-9> mathx8
<9-10> mathx9
<10-12> mathx10
<12-> mathx12
}{}

\DeclareSymbolFont{mathx}{U}{mathx}{m}{n}
\DeclareFontSubstitution{U}{mathx}{m}{n}

\DeclareMathSymbol{\bigovoid}{\mathop}{mathx}{"EC}

\newcommand{\BigO}{\mathop{\stackinset{c}{}{c}{}{ \scalebox{1.1}{$\bigovoid$}}{ \scalebox{1.15}{$\bigovoid$}}}}

\newcommand{\Conv}{%
  \mathop{\!\scalebox{2}{\raisebox{-0.2ex}{$\circledast$}}
  }
}
\newcommand{\conv}{%
  \mathop{\scalebox{1.5}{\raisebox{-0.1ex}{$\circledast$}}
  }
}

\usepackage{xcolor}
\usepackage{pifont}
\usepackage[textsize=tiny]{todonotes}
\usepackage{multirow}
\usepackage{pgfplots}
\pgfplotsset{compat=1.7}
\usetikzlibrary{positioning, decorations.pathreplacing, decorations.markings, calc,arrows, pgfplots.groupplots, decorations.pathreplacing, decorations.markings, patterns}
\usepackage{wrapfig}

\icmltitlerunning{Efficient Latency-Aware CNN Depth Compression via Two-Stage Dynamic Programming}

\begin{document}

\twocolumn[
\icmltitle{Efficient Latency-Aware CNN Depth Compression via \\Two-Stage Dynamic Programming}



\icmlsetsymbol{equal}{*}

\begin{icmlauthorlist}
\icmlauthor{Jinuk Kim}{equal,snu,nprc}
\icmlauthor{Yeonwoo Jeong}{equal,snu,nprc} 
\icmlauthor{Deokjae Lee}{snu,nprc}
\icmlauthor{Hyun Oh Song}{snu,nprc}
\end{icmlauthorlist}

\icmlaffiliation{snu}{
Department of Computer Science and Engineering, Seoul National University}
\icmlaffiliation{nprc}{
Neural Processing Research Center}

\icmlcorrespondingauthor{Hyun Oh Song}{hyunoh@snu.ac.kr} 

\icmlkeywords{Machine Learning, ICML}

\vskip 0.3in
]


\printAffiliationsAndNotice{\icmlEqualContribution} 

\begin{abstract}
Recent works on neural network pruning advocate that reducing the depth of the network is more effective in reducing run-time memory usage and accelerating inference latency than reducing the width of the network through channel pruning. In this regard, some recent works propose depth compression algorithms that merge convolution layers. However, the existing algorithms have a constricted search space and rely on human-engineered heuristics. In this paper, we propose a novel depth compression algorithm which targets general convolution operations. We propose a subset selection problem that replaces inefficient activation layers with identity functions and optimally merges consecutive convolution operations into shallow equivalent convolution operations for efficient end-to-end inference latency. Since the proposed subset selection problem is NP-hard, we formulate a surrogate optimization problem that can be solved exactly via two-stage dynamic programming within a few seconds. We evaluate our methods and baselines by TensorRT for a fair inference latency comparison. Our method outperforms the baseline method with higher accuracy and faster inference speed in MobileNetV2 on the ImageNet dataset. Specifically, we achieve $1.41\times$ speed-up with $0.11$\%p accuracy gain in MobileNetV2-1.0 on the ImageNet. 

\end{abstract}
\newcommand{\red}[1]{\textcolor{red}{#1}}
\newcommand{\blue}[1]{\textcolor{blue}{#1}}

\section{Introduction}
Deep learning with Convolutional Neural Network (CNN) has achieved outstanding results in various fields such as image classification, object detection, image segmentation, and generation \citep{efficientnet,detection,segmentation,generation}. However, the success of CNNs in such fields is accompanied by the challenge of increased complexity and inference latency. For real-world applications, accelerating the inference latency of CNNs is of great practical importance, especially when deploying the models on edge devices with limited resources.

To this end, a line of research called channel pruning has been introduced to remove unnecessary channels in CNNs to accelerate the wall-clock time in the edge device while preserving the performance of the CNNs \citep{wen2016learning,chipnet,halp}. However, with the advancement of hardware technology for parallel computation, channel pruning which reduces the width of neural networks has become less effective than removing entire layers in terms of latency \citep{layerpruning,layerpruning2,layerpruning3,depthshrinker}.

In contrast, layer pruning, which prunes entire layers, has been proposed to reduce the depth of neural networks. Layer pruning also significantly reduces the run-time memory usage and achieves effective speed-up in many edge devices compared to channel pruning \citep{layerpruning3}. However, layer pruning is more aggressive than channel pruning in terms of reducing the number of parameters and FLOPs, thereby resulting in a more severe accuracy drop compared to channel pruning methods. Instead of naively removing an entire layer, \citet{depthshrinker} present a depth compression algorithm called DepthShrinker which integrates layers by replacing inefficient consecutive depth-wise convolution and point-wise convolution with an efficient dense convolution operation in MobileNetV2 \citep{mobilenet}. This compression algorithm results in depth reduction with low run-time memory usage and fast inference latency similar to layer pruning. However, the depth compression algorithm does not suffer from a commensurate accuracy drop.

\newcommand\grid[3]{
    \def\p{#1}
    \def\xx{#2}
    \def\yy{#3}
  \draw[fill,\c,opacity=0.9] (\p) rectangle ++(\xx,\yy);
  \draw[gray!60] ($(\p) + (\xx/3,0)$) -- ($(\p) + (\xx/3,\yy)$);
  \draw[gray!60] ($(\p) + (\xx/3*2,0)$) -- ($(\p) + (\xx/3*2,\yy)$);
  \draw[gray!60] ($(\p) + (0,\yy/3)$) -- ($(\p) + (\xx,\yy/3)$);
  \draw[gray!60] ($(\p) + (0,\yy/3*2)$) -- ($(\p) + (\xx,\yy/3*2)$);
  \draw (\p) rectangle ++(\xx,\yy);
}
\newcommand\gridd[3]{
    \def\p{#1}
    \def\xx{#2}
    \def\yy{#3}
  \draw[fill,\c,opacity=0.9] (\p) rectangle ++(\xx,\yy);
  \draw[gray!60] ($(\p) + (\xx/5,0)$) -- ($(\p) + (\xx/5,\yy)$);
  \draw[gray!60] ($(\p) + (\xx/5*2,0)$) -- ($(\p) + (\xx/5*2,\yy)$);
  \draw[gray!60] ($(\p) + (\xx/5*3,0)$) -- ($(\p) + (\xx/5*3,\yy)$);
  \draw[gray!60] ($(\p) + (\xx/5*4,0)$) -- ($(\p) + (\xx/5*4,\yy)$);
  \draw[gray!60] ($(\p) + (0,\yy/5)$) -- ($(\p) + (\xx,\yy/5)$);
  \draw[gray!60] ($(\p) + (0,\yy/5*2)$) -- ($(\p) + (\xx,\yy/5*2)$);
  \draw[gray!60] ($(\p) + (0,\yy/5*3)$) -- ($(\p) + (\xx,\yy/5*3)$);
  \draw[gray!60] ($(\p) + (0,\yy/5*4)$) -- ($(\p) + (\xx,\yy/5*4)$);
  \draw (\p) rectangle ++(\xx,\yy);
}
\newcommand\griddd[3]{
    \def\p{#1}
    \def\xx{#2}
    \def\yy{#3}
  \draw[fill,\c,opacity=0.9] (\p) rectangle ++(\xx,\yy);
  \draw[gray!60] ($(\p) + (\xx/7,0)$) -- ($(\p) + (\xx/7,\yy)$);
  \draw[gray!60] ($(\p) + (\xx/7*2,0)$) -- ($(\p) + (\xx/7*2,\yy)$);
  \draw[gray!60] ($(\p) + (\xx/7*3,0)$) -- ($(\p) + (\xx/7*3,\yy)$);
  \draw[gray!60] ($(\p) + (\xx/7*4,0)$) -- ($(\p) + (\xx/7*4,\yy)$);
  \draw[gray!60] ($(\p) + (\xx/7*5,0)$) -- ($(\p) + (\xx/7*5,\yy)$);
  \draw[gray!60] ($(\p) + (\xx/7*6,0)$) -- ($(\p) + (\xx/7*6,\yy)$);
  \draw[gray!60] ($(\p) + (0,\yy/7)$) -- ($(\p) + (\xx,\yy/7)$);
  \draw[gray!60] ($(\p) + (0,\yy/7*2)$) -- ($(\p) + (\xx,\yy/7*2)$);
  \draw[gray!60] ($(\p) + (0,\yy/7*3)$) -- ($(\p) + (\xx,\yy/7*3)$);
  \draw[gray!60] ($(\p) + (0,\yy/7*4)$) -- ($(\p) + (\xx,\yy/7*4)$);
  \draw[gray!60] ($(\p) + (0,\yy/7*5)$) -- ($(\p) + (\xx,\yy/7*5)$);
  \draw[gray!60] ($(\p) + (0,\yy/7*6)$) -- ($(\p) + (\xx,\yy/7*6)$);
  \draw (\p) rectangle ++(\xx,\yy);
}

\newcommand\fm[5]{
    \def\nin{#1}
    \def\convid{#2}
    \def\xx{#3}
    \def\yy{#4}
    \def\ccc{#5}
        \foreach \i in {0,1,...,\nin} {
        \begin{scope}[  
            yshift={1*\channelwidth*\i},every node/.append style={
                 yslant=0,xslant=1,rotate=\rot},yslant=0,xslant=1,rotate=\rot
              ]              
              \coordinate (A\convid\i) at (\xx,\yy);
              \coordinate (DA\convid\i) at ($(\xx,\yy)+(\fmw,\fmh)$);
        \end{scope}
        }
    \draw[\ccc] (A\convid0) -- (A\convid\nin);
    \draw[\ccc] ($(A\convid0) + (\fmw,0)$) -- ($(A\convid\nin) + (\fmw,0)$);
    \draw[\ccc] (DA\convid0) -- (DA\convid\nin);
        \foreach \i in {0,1,...,\nin} {
        \begin{scope}[  
            yshift={1.5*\channelwidth*\i},every node/.append style={
                 yslant=0,xslant=1,rotate=\rot},yslant=0,xslant=1,rotate=\rot
              ]              
              \draw[fill,\ccc,opacity=0.8] (A\convid\i) rectangle ++(\fmw,\fmh);
              \ifnum\i=0
              \draw[red!40] (A\convid\i) rectangle ++(\fmw,\fmh);
              \else
            \ifnum\i=\nin
              \draw (A\convid\i) rectangle ++(\fmw,\fmh);            
            \else 
              \draw[red!40] (A\convid\i) rectangle ++(\fmw,\fmh);           
            \fi
            \fi
        \end{scope}
        }
    \draw[fill,\ccc,opacity=0.7] (A\convid0) -- (A\convid\nin) -- ($(A\convid\nin)+(\fmw,0)$) -- ($(A\convid0)+(\fmw,0)$) --cycle; 
    \draw[fill,\ccc,opacity=0.7] (DA\convid0) -- (DA\convid\nin) -- ($(A\convid\nin)+(\fmw,0)$) -- ($(A\convid0)+(\fmw,0)$) --cycle; 
    \draw[black] (DA\convid0) -- (DA\convid\nin) -- ($(A\convid\nin)+(\fmw,0)$) -- ($(A\convid0)+(\fmw,0)$) --cycle; 
    \draw[black] (A\convid0) -- (A\convid\nin) -- ($(A\convid\nin)+(\fmw,0)$) -- ($(A\convid0)+(\fmw,0)$) --cycle; 
    \foreach \i in {0,1,...,\nin} {
        \begin{scope}[  
            yshift={1.5*\channelwidth*\i},every node/.append style={
                 yslant=0,xslant=1,rotate=\rot},yslant=0,xslant=1,rotate=\rot
              ]              
              \draw[red!80!black!60] ($(A\convid\i)+(\fmw,0)$) -- ($(A\convid\i)+(\fmw,\fmh)$);
              \draw[red!80!black!60] (A\convid\i) -- ($(A\convid\i)+(\fmw,0.0)$);
        \end{scope}
        }
        \begin{scope}[  
            yshift={-100},every node/.append style={
                 yslant=0,xslant=1,rotate=\rot},yslant=0,xslant=1,rotate=\rot
              ]              
              \draw[black] ($(A\convid\nin)+(\fmw,0)$) -- ($(A\convid\nin)+(\fmw,\fmh)$);
              \draw[black] (A\convid\nin) -- ($(A\convid\nin)+(\fmw,0.0)$);
              \draw[black] ($(A\convid0)+(\fmw,0)$) -- ($(A\convid0)+(\fmw,\fmh)$);
              \draw[black] (A\convid0) -- ($(A\convid0)+(\fmw,0.0)$);
        \end{scope}
}

\newcommand\ggfm[5]{
    \def\nin{#1}
    \def\convid{#2}
    \def\xx{#3}
    \def\yy{#4}
    \def\margin{#5}
    \def\ccc{red!20}
        \foreach \i in {0,1,...,\nin} {
        \begin{scope}[  
            yshift={\margin + 1*\channelwidth*\i},every node/.append style={
                 yslant=0,xslant=1,rotate=\rot},yslant=0,xslant=1,rotate=\rot
              ]              
              \coordinate (A\convid\i) at (\xx,\yy);
              \coordinate (DA\convid\i) at ($(\xx,\yy)+(1,0.7)$);
        \end{scope}
        }
    \draw[\ccc] (A\convid0) -- (A\convid\nin);
    \draw[\ccc] ($(A\convid0) + (1,0)$) -- ($(A\convid\nin) + (1,0)$);
    \draw[\ccc] (DA\convid0) -- (DA\convid\nin);
        \foreach \i in {0,1,...,\nin} {
        \begin{scope}[  
            yshift={\margin + 1.5*\channelwidth*\i},every node/.append style={
                 yslant=0,xslant=1,rotate=\rot},yslant=0,xslant=1,rotate=\rot
              ]              
              \draw[fill,\ccc,opacity=0.7] (A\convid\i) rectangle ++(1,0.7);
              \ifnum\i=0
              \draw[red!40] (A\convid\i) rectangle ++(1,0.7);
              \else
            \ifnum\i=\nin
              \draw (A\convid\i) rectangle ++(1,0.7);            
            \else 
              \draw[red!40] (A\convid\i) rectangle ++(1,0.7);           
            \fi
            \fi
        \end{scope}
        }
    \draw[fill,\ccc,opacity=0.7] (A\convid0) -- (A\convid\nin) -- ($(A\convid\nin)+(1,0)$) -- ($(A\convid0)+(1,0)$) --cycle; 
    \draw[fill,\ccc,opacity=0.7] (DA\convid0) -- (DA\convid\nin) -- ($(A\convid\nin)+(1,0)$) -- ($(A\convid0)+(1,0)$) --cycle; 
    \draw[black] (DA\convid0) -- (DA\convid\nin) -- ($(A\convid\nin)+(1,0)$) -- ($(A\convid0)+(1,0)$) --cycle; 
    \draw[black] (A\convid0) -- (A\convid\nin) -- ($(A\convid\nin)+(1,0)$) -- ($(A\convid0)+(1,0)$) --cycle; 
    \foreach \i in {0,1,...,\nin} {
        \begin{scope}[  
            yshift={\margin + 1.5*\channelwidth*\i},every node/.append style={
                 yslant=0,xslant=1,rotate=\rot},yslant=0,xslant=1,rotate=\rot
              ]              
              \draw[red!80!black!60] ($(A\convid\i)+(1,0)$) -- ($(A\convid\i)+(1,0.7)$);
              \draw[red!80!black!60] (A\convid\i) -- ($(A\convid\i)+(1,0.0)$);
        \end{scope}
        }
        \begin{scope}[  
            yshift={\margin},every node/.append style={
                 yslant=0,xslant=1,rotate=\rot},yslant=0,xslant=1,rotate=\rot
              ]              
              \draw[black] ($(A\convid\nin)+(1,0)$) -- ($(A\convid\nin)+(1,0.7)$);
              \draw[black] (A\convid\nin) -- ($(A\convid\nin)+(1,0.0)$);
              \draw[black] ($(A\convid0)+(1,0)$) -- ($(A\convid0)+(1,0.7)$);
              \draw[black] (A\convid0) -- ($(A\convid0)+(1,0.0)$);
        \end{scope}
}
\newcommand\gfm[5]{
    \def\nin{#1}
    \def\convid{#2}
    \def\xx{#3}
    \def\yy{#4}
    \def\margin{#5}
    \def\ccc{green!20}
        \foreach \i in {0,1,...,\nin} {
        \begin{scope}[  
            yshift={\margin + 1*\channelwidth*\i},every node/.append style={
                 yslant=0,xslant=1,rotate=\rot},yslant=0,xslant=1,rotate=\rot
              ]              
              \coordinate (A\convid\i) at (\xx,\yy);
              \coordinate (DA\convid\i) at ($(\xx,\yy)+(1,0.7)$);
        \end{scope}
        }
    \draw[\ccc] (A\convid0) -- (A\convid\nin);
    \draw[\ccc] ($(A\convid0) + (1,0)$) -- ($(A\convid\nin) + (1,0)$);
    \draw[\ccc] (DA\convid0) -- (DA\convid\nin);
        \foreach \i in {0,1,...,\nin} {
        \begin{scope}[  
            yshift={\margin + 1.5*\channelwidth*\i},every node/.append style={
                 yslant=0,xslant=1,rotate=\rot},yslant=0,xslant=1,rotate=\rot
              ]              
              \draw[fill,\ccc,opacity=0.7] (A\convid\i) rectangle ++(1,0.7);
              \ifnum\i=0
              \draw[green!80!black!40] (A\convid\i) rectangle ++(1,0.7);
              \else
            \ifnum\i=\nin
              \draw (A\convid\i) rectangle ++(1,0.7);            
            \else 
              \draw[green!80!black!40] (A\convid\i) rectangle ++(1,0.7);           
            \fi
            \fi
        \end{scope}
        }
    \draw[fill,\ccc,opacity=0.7] (A\convid0) -- (A\convid\nin) -- ($(A\convid\nin)+(1,0)$) -- ($(A\convid0)+(1,0)$) --cycle; 
    \draw[fill,\ccc,opacity=0.7] (DA\convid0) -- (DA\convid\nin) -- ($(A\convid\nin)+(1,0)$) -- ($(A\convid0)+(1,0)$) --cycle; 
    \draw[black] (DA\convid0) -- (DA\convid\nin) -- ($(A\convid\nin)+(1,0)$) -- ($(A\convid0)+(1,0)$) --cycle; 
    \draw[black] (A\convid0) -- (A\convid\nin) -- ($(A\convid\nin)+(1,0)$) -- ($(A\convid0)+(1,0)$) --cycle; 
    \foreach \i in {0,1,...,\nin} {
        \begin{scope}[  
            yshift={\margin + 1.5*\channelwidth*\i},every node/.append style={
                 yslant=0,xslant=1,rotate=\rot},yslant=0,xslant=1,rotate=\rot
              ]              
              \draw[green!80!black!100] ($(A\convid\i)+(1,0)$) -- ($(A\convid\i)+(1,0.7)$);
              \draw[green!80!black!100] (A\convid\i) -- ($(A\convid\i)+(1,0.0)$);
        \end{scope}
        }
        \begin{scope}[  
            yshift={\margin},every node/.append style={
                 yslant=0,xslant=1,rotate=\rot},yslant=0,xslant=1,rotate=\rot
              ]              
              \draw[black] ($(A\convid\nin)+(1,0)$) -- ($(A\convid\nin)+(1,0.7)$);
              \draw[black] (A\convid\nin) -- ($(A\convid\nin)+(1,0.0)$);
              \draw[black] ($(A\convid0)+(1,0)$) -- ($(A\convid0)+(1,0.7)$);
              \draw[black] (A\convid0) -- ($(A\convid0)+(1,0.0)$);
        \end{scope}
}
\newcommand\bfm[5]{
    \def\nin{#1}
    \def\convid{#2}
    \def\xx{#3}
    \def\yy{#4}
    \def\ccc{#5}
        \foreach \i in {0,1,...,\nin} {
        \begin{scope}[  
            yshift={-100+1*\channelwidth*\i},every node/.append style={
                 yslant=0,xslant=1,rotate=\rot},yslant=0,xslant=1,rotate=\rot
              ]              
              \coordinate (BA\convid\i) at (\xx,\yy);
              \coordinate (DBA\convid\i) at ($(\xx,\yy)+(1,0.7)$);
        \end{scope}
        }
    \draw[\ccc] (BA\convid0) -- (BA\convid\nin);
    \draw[\ccc] ($(BA\convid0) + (1,0)$) -- ($(BA\convid\nin) + (1,0)$);
    \draw[\ccc] (DBA\convid0) -- (DBA\convid\nin);
        \foreach \i in {0,1,...,\nin} {
        \begin{scope}[  
            yshift={-100+1.5*\channelwidth*\i},every node/.append style={
                 yslant=0,xslant=1,rotate=\rot},yslant=0,xslant=1,rotate=\rot
              ]              
              \draw[fill,\ccc,opacity=0.8] (BA\convid\i) rectangle ++(1,0.7);
              \ifnum\i=0
              \draw[red!40] (BA\convid\i) rectangle ++(1,0.7);
              \else
            \ifnum\i=\nin
              \draw (BA\convid\i) rectangle ++(1,0.7);            
            \else 
              \draw[red!40] (BA\convid\i) rectangle ++(1,0.7);           
            \fi
            \fi
        \end{scope}
        }
    \draw[fill,\ccc,opacity=0.7] (BA\convid0) -- (BA\convid\nin) -- ($(BA\convid\nin)+(1,0)$) -- ($(BA\convid0)+(1,0)$) --cycle; 
    \draw[fill,\ccc,opacity=0.7] (DBA\convid0) -- (DBA\convid\nin) -- ($(BA\convid\nin)+(1,0)$) -- ($(BA\convid0)+(1,0)$) --cycle; 
    \draw[black] (DBA\convid0) -- (DBA\convid\nin) -- ($(BA\convid\nin)+(1,0)$) -- ($(BA\convid0)+(1,0)$) --cycle; 
    \draw[black] (BA\convid0) -- (BA\convid\nin) -- ($(BA\convid\nin)+(1,0)$) -- ($(BA\convid0)+(1,0)$) --cycle; 
    \foreach \i in {0,1,...,\nin} {
        \begin{scope}[  
            yshift={-100+1.5*\channelwidth*\i},every node/.append style={
                 yslant=0,xslant=1,rotate=\rot},yslant=0,xslant=1,rotate=\rot
              ]              
              \draw[red!80!black!60] ($(BA\convid\i)+(1,0)$) -- ($(BA\convid\i)+(1,0.7)$);
              \draw[red!80!black!60] (BA\convid\i) -- ($(BA\convid\i)+(1,0.0)$);
        \end{scope}
        }
        \begin{scope}[  
            yshift={-100},every node/.append style={
                 yslant=0,xslant=1,rotate=\rot},yslant=0,xslant=1,rotate=\rot
              ]              
              \draw[black] ($(BA\convid\nin)+(1,0)$) -- ($(BA\convid\nin)+(1,0.7)$);
              \draw[black] (BA\convid\nin) -- ($(BA\convid\nin)+(1,0.0)$);
              \draw[black] ($(BA\convid0)+(1,0)$) -- ($(BA\convid0)+(1,0.7)$);
              \draw[black] (BA\convid0) -- ($(BA\convid0)+(1,0.0)$);
        \end{scope}
}

\newcommand\ffm[5]{
    \def\nin{#1}
    \def\convid{#2}
    \def\xx{#3}
    \def\yy{#4}
    \def\ccc{#5}
        \foreach \i in {0,1,...,\nin} {
        \begin{scope}[  
            yshift={-200+1*\channelwidth*\i},every node/.append style={
                 yslant=0,xslant=1,rotate=\rot},yslant=0,xslant=1,rotate=\rot
              ]              
              \coordinate (BA\convid\i) at (\xx,\yy);
              \coordinate (DBA\convid\i) at ($(\xx,\yy)+(1,0.7)$);
        \end{scope}
        }
    \draw[\ccc] (BA\convid0) -- (BA\convid\nin);
    \draw[\ccc] ($(BA\convid0) + (1,0)$) -- ($(BA\convid\nin) + (1,0)$);
    \draw[\ccc] (DBA\convid0) -- (DBA\convid\nin);
        \foreach \i in {0,1,...,\nin} {
        \begin{scope}[  
            yshift={-200+1.5*\channelwidth*\i},every node/.append style={
                 yslant=0,xslant=1,rotate=\rot},yslant=0,xslant=1,rotate=\rot
              ]              
              \draw[fill,\ccc,opacity=0.8] (BA\convid\i) rectangle ++(1,0.7);
              \ifnum\i=0
              \draw[red!40] (BA\convid\i) rectangle ++(1,0.7);
              \else
            \ifnum\i=\nin
              \draw (BA\convid\i) rectangle ++(1,0.7);            
            \else 
              \draw[red!40] (BA\convid\i) rectangle ++(1,0.7);           
            \fi
            \fi
        \end{scope}
        }
    \draw[fill,\ccc,opacity=0.7] (BA\convid0) -- (BA\convid\nin) -- ($(BA\convid\nin)+(1,0)$) -- ($(BA\convid0)+(1,0)$) --cycle; 
    \draw[fill,\ccc,opacity=0.7] (DBA\convid0) -- (DBA\convid\nin) -- ($(BA\convid\nin)+(1,0)$) -- ($(BA\convid0)+(1,0)$) --cycle; 
    \draw[black] (DBA\convid0) -- (DBA\convid\nin) -- ($(BA\convid\nin)+(1,0)$) -- ($(BA\convid0)+(1,0)$) --cycle; 
    \draw[black] (BA\convid0) -- (BA\convid\nin) -- ($(BA\convid\nin)+(1,0)$) -- ($(BA\convid0)+(1,0)$) --cycle; 
    \foreach \i in {0,1,...,\nin} {
        \begin{scope}[  
            yshift={-200+1.5*\channelwidth*\i},every node/.append style={
                 yslant=0,xslant=1,rotate=\rot},yslant=0,xslant=1,rotate=\rot
              ]              
              \draw[red!80!black!60] ($(BA\convid\i)+(1,0)$) -- ($(BA\convid\i)+(1,0.7)$);
              \draw[red!80!black!60] (BA\convid\i) -- ($(BA\convid\i)+(1,0.0)$);
        \end{scope}
        }
        \begin{scope}[  
            yshift={-200},every node/.append style={
                 yslant=0,xslant=1,rotate=\rot},yslant=0,xslant=1,rotate=\rot
              ]              
              \draw[black] ($(BA\convid\nin)+(1,0)$) -- ($(BA\convid\nin)+(1,0.7)$);
              \draw[black] (BA\convid\nin) -- ($(BA\convid\nin)+(1,0.0)$);
              \draw[black] ($(BA\convid0)+(1,0)$) -- ($(BA\convid0)+(1,0.7)$);
              \draw[black] (BA\convid0) -- ($(BA\convid0)+(1,0.0)$);
        \end{scope}
}
\newcommand\bbconv[4]{
    \def\nin{#1}
    \def\convid{#2}
    \def\xx{#3}
    \def\yy{#4}
    \foreach \i in {0,1,...,\nin} {
        \begin{scope}[  
            yshift={\channelwidth*\i},every node/.append style={
                 yslant=0,xslant=1,rotate=\rot},yslant=0,xslant=1,rotate=\rot
              ]              
                \coordinate (B\convid\i) at ($(\xx,\yy)+(0,0)$);
                \coordinate (DB\convid\i) at ($(\xx,\yy)+(0.5,0.35)$);
              \grid{B\convid\i}{0.5}{0.35}
        \end{scope}
    }

    \draw[fill,gray!10,opacity=0.8] (B\convid0) -- (B\convid\nin) -- ($(B\convid\nin)+(0.5,0)$) -- ($(B\convid0)+(0.5,0)$) --cycle; 
    \draw[fill,gray!10,opacity=0.8] (DB\convid0) -- (DB\convid\nin) -- ($(B\convid\nin)+(0.5,0)$) -- ($(B\convid0)+(0.5,0)$) --cycle; 
    \foreach \i in {0,1,...,\nin} {
        \begin{scope}[  
            yshift={\channelwidth*\i},every node/.append style={
                 yslant=0,xslant=1,rotate=\rot},yslant=0,xslant=1,rotate=\rot
              ]              
            \draw[gray!80] (B\convid\i) -- ($(B\convid\i)+(0.5,0)$);
            \draw[gray!80] (DB\convid\i) -- ($(B\convid\i)+(0.5,0)$);
        \end{scope}
    }
    \draw[black] (B\convid0) -- (B\convid\nin) -- ($(B\convid\nin)+(0.5,0)$) -- ($(B\convid0)+(0.5,0)$) --cycle; 
    \draw[black] (DB\convid0) -- (DB\convid\nin) -- ($(B\convid\nin)+(0.5,0)$) -- ($(B\convid0)+(0.5,0)$) --cycle;

}

\newcommand\bbbconv[4]{
    \def\nin{#1}
    \def\convid{#2}
    \def\xx{#3}
    \def\yy{#4}
    \foreach \i in {0,1,...,\nin} {
        \begin{scope}[  
            yshift={100+\channelwidth*\i},every node/.append style={
                 yslant=0,xslant=1,rotate=\rot},yslant=0,xslant=1,rotate=\rot
              ]              
                \coordinate (B\convid\i) at ($(\xx,\yy)+(0,0)$);
                \coordinate (DB\convid\i) at ($(\xx,\yy)+(0.5,0.35)$);
              \grid{B\convid\i}{0.5}{0.35}
        \end{scope}
    }

    \draw[fill,gray!10,opacity=0.8] (B\convid0) -- (B\convid\nin) -- ($(B\convid\nin)+(0.5,0)$) -- ($(B\convid0)+(0.5,0)$) --cycle; 
    \draw[fill,gray!10,opacity=0.8] (DB\convid0) -- (DB\convid\nin) -- ($(B\convid\nin)+(0.5,0)$) -- ($(B\convid0)+(0.5,0)$) --cycle; 
    \foreach \i in {0,1,...,\nin} {
        \begin{scope}[  
            yshift={100+\channelwidth*\i},every node/.append style={
                 yslant=0,xslant=1,rotate=\rot},yslant=0,xslant=1,rotate=\rot
              ]              
            \draw[gray!80] (B\convid\i) -- ($(B\convid\i)+(0.5,0)$);
            \draw[gray!80] (DB\convid\i) -- ($(B\convid\i)+(0.5,0)$);
        \end{scope}
    }
    \draw[black] (B\convid0) -- (B\convid\nin) -- ($(B\convid\nin)+(0.5,0)$) -- ($(B\convid0)+(0.5,0)$) --cycle; 
    \draw[black] (DB\convid0) -- (DB\convid\nin) -- ($(B\convid\nin)+(0.5,0)$) -- ($(B\convid0)+(0.5,0)$) --cycle;

}
\newcommand\bconv[4]{
    \def\nin{#1}
    \def\convid{#2}
    \def\xx{#3}
    \def\yy{#4}
    \foreach \i in {0,1,...,\nin} {
        \begin{scope}[  
            yshift={-100+\channelwidth*\i},every node/.append style={
                 yslant=0,xslant=1,rotate=\rot},yslant=0,xslant=1,rotate=\rot
              ]              
                \coordinate (EB\convid\i) at ($(\xx,\yy)+(0,0)$);
                \coordinate (DEB\convid\i) at ($(\xx,\yy)+(0.5,0.35)$);
              \grid{EB\convid\i}{0.5}{0.35}
        \end{scope}
    }

    \draw[fill,gray!10,opacity=0.8] (EB\convid0) -- (EB\convid\nin) -- ($(EB\convid\nin)+(0.5,0)$) -- ($(EB\convid0)+(0.5,0)$) --cycle; 
    \draw[fill,gray!10,opacity=0.8] (DEB\convid0) -- (DEB\convid\nin) -- ($(EB\convid\nin)+(0.5,0)$) -- ($(EB\convid0)+(0.5,0)$) --cycle; 
    \foreach \i in {0,1,...,\nin} {
        \begin{scope}[  
            yshift={-100+\channelwidth*\i},every node/.append style={
                 yslant=0,xslant=1,rotate=\rot},yslant=0,xslant=1,rotate=\rot
              ]              
            \draw[gray!80] (EB\convid\i) -- ($(EB\convid\i)+(0.5,0)$);
            \draw[gray!80] (DEB\convid\i) -- ($(EB\convid\i)+(0.5,0)$);
        \end{scope}
    }
    \draw[black] (EB\convid0) -- (EB\convid\nin) -- ($(EB\convid\nin)+(0.5,0)$) -- ($(EB\convid0)+(0.5,0)$) --cycle; 
    \draw[black] (DEB\convid0) -- (DEB\convid\nin) -- ($(EB\convid\nin)+(0.5,0)$) -- ($(EB\convid0)+(0.5,0)$) --cycle; 
}

\newcommand\bconvv[4]{
    \def\nin{#1}
    \def\convid{#2}
    \def\xx{#3}
    \def\yy{#4}
    \foreach \i in {0,1,...,\nin} {
        \begin{scope}[  
            yshift={-100+\channelwidth*\i},every node/.append style={
                 yslant=0,xslant=1,rotate=\rot},yslant=0,xslant=1,rotate=\rot
              ]              
                \coordinate (EB\convid\i) at ($(\xx,\yy)+(0,0)$);
                \coordinate (DEB\convid\i) at ($(\xx,\yy)+(0.5*5/3,0.35*5/3)$);
              \gridd{EB\convid\i}{0.5*5/3}{0.35*5/3}
        \end{scope}
    }

    \draw[fill,gray!10,opacity=0.8] (EB\convid0) -- (EB\convid\nin) -- ($(EB\convid\nin)+(0.5*5/3,0)$) -- ($(EB\convid0)+(0.5*5/3,0)$) --cycle; 
    \draw[fill,gray!10,opacity=0.8] (DEB\convid0) -- (DEB\convid\nin) -- ($(EB\convid\nin)+(0.5*5/3,0)$) -- ($(EB\convid0)+(0.5*5/3,0)$) --cycle; 
    \foreach \i in {0,1,...,\nin} {
        \begin{scope}[  
            yshift={-100+\channelwidth*\i},every node/.append style={
                 yslant=0,xslant=1,rotate=\rot},yslant=0,xslant=1,rotate=\rot
              ]              
            \draw[gray!80] (EB\convid\i) -- ($(EB\convid\i)+(0.5*5/3,0)$);
            \draw[gray!80] (DEB\convid\i) -- ($(EB\convid\i)+(0.5*5/3,0)$);
        \end{scope}
    }
    \draw[black] (EB\convid0) -- (EB\convid\nin) -- ($(EB\convid\nin)+(0.5*5/3,0)$) -- ($(EB\convid0)+(0.5*5/3,0)$) --cycle; 
    \draw[black] (DEB\convid0) -- (DEB\convid\nin) -- ($(EB\convid\nin)+(0.5*5/3,0)$) -- ($(EB\convid0)+(0.5*5/3,0)$) --cycle; 
}
\newcommand\relu[2]{
        \def\nodename{#1}
        \def\writename{#2}
        \draw[blue!60!black!100, thick] (\nodename) circle [radius=0.35];
        \draw[blue!60!black!100, thick] ($(\nodename) + (-0.35,0)$) -- (\nodename);
        \draw[blue!60!black!100, thick] ($(\nodename) + (0.35*1.414/2,0.35*1.414/2)$) -- (\nodename);
        \node[blue!60!black!100] at ($(\nodename)+(0,-0.9)$) {ReLU};
        \node[blue!60!black!100] at ($(\nodename)+(0,+0.7)$) {\writename};
}
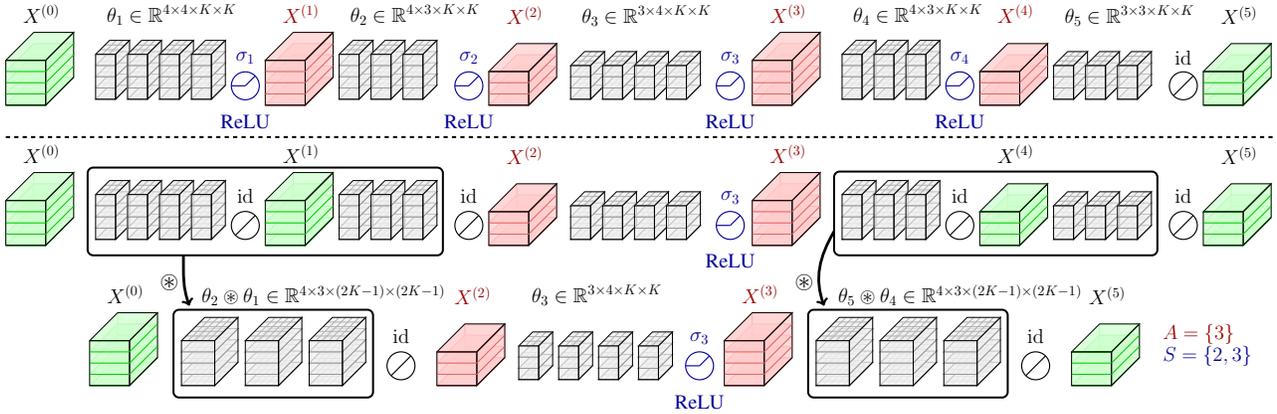
\begin{figure*}[ht]
    \centering
\def\rot{0}
\def\dd{0.8}
\def\ggg{0.6}
\def\c{gray!10}
\def\d{green!20}
\def\channelwidth{8}
\def\fmw{1.0}
\def\fmh{0.7}
\resizebox{2.07\columnwidth}{!}{
\pgfdeclarelayer{fig}\pgfdeclarelayer{bg}
\pgfsetlayers{bg,fig}  
\begin{tikzpicture}[decoration={brace},font=\Large][scale=1,every node/.style={minimum size=1cm},on grid]
\begin{pgfonlayer}{fig}
        \gfm{4}{0}{-2.1}{0.0}{100}
        \bbbconv{4}{0}{0.0}{0.15}
        \bbbconv{4}{1}{0.8}{0.15}
        \bbbconv{4}{2}{1.6}{0.15}
        \bbbconv{4}{3}{2.4}{0.15}
        \ggfm{4}{1}{4.4}{0.0}{100}
        \bbbconv{4}{4}{6.1}{0.15}
        \bbbconv{4}{5}{6.9}{0.15}
        \bbbconv{4}{6}{7.7}{0.15}
        \ggfm{3}{2}{10.0}{0.0}{100}
        \bbbconv{3}{7}{11.9}{0.15}
        \bbbconv{3}{8}{12.7}{0.15}
        \bbbconv{3}{9}{13.5}{0.15}
        \bbbconv{3}{10}{14.3}{0.15}
        \ggfm{4}{3}{16.6}{0.0}{100}
        \bbbconv{4}{11}{18.7}{0.15}
        \bbbconv{4}{12}{19.5}{0.15}
        \bbbconv{4}{13}{20.3}{0.15}
        \ggfm{3}{4}{22.3}{0.0}{100}
        \bbbconv{3}{14}{24.0}{0.15}
        \bbbconv{3}{15}{24.8}{0.15}
        \bbbconv{3}{16}{25.6}{0.15}
        \gfm{3}{5}{27.9}{0.0}{100}
        \coordinate (id) at (3.9,4);
        \coordinate (id2) at (9.5,4);
        \coordinate (id3) at (21.8,4);
        \coordinate (act) at (16.05, 4);
        \coordinate (act2) at (27.4, 4);
        \relu{id}{$\sigma_1$}
        \relu{id2}{$\sigma_2$}
        \relu{act}{$\sigma_3$}
        \relu{id3}{$\sigma_4$}
            
        \draw[black!100, thick] (act2) circle [radius=0.35];
        \draw[black!100, thick] ($(act2) + (0.35*1.414/2,0.35*1.414/2)$) -- ($(act2) + (-0.35*1.414/2,-0.35*1.414/2)$);
        \node[black!100] at ($(act2)+(0,+0.7)$) {$\mathrm{id}$};

    \node at ($(B24)+(0.3,1.)$) {$\theta_1\in\mathbb{R}^{4\times 4\times K\times K}$};
    \node at ($(B54)+(1.1,1.)$) {$\theta_2\in\mathbb{R}^{4\times 3\times K\times K}$};
    \node at ($(B93)+(0.3,1.25)$) {$\theta_3\in\mathbb{R}^{3\times 4\times K \times K}$};
    \node at ($(B124)+(1.1,1.)$) {$\theta_4\in\mathbb{R}^{4\times 3\times K \times K}$};
    \node at ($(B153)+(1.1,1.25)$) {$\theta_5\in\mathbb{R}^{3\times 3\times K \times K}$};

    \node[black!100] at ($(A03)+(0.9,1.45)$) {$X^{(0)}$};
    \node[red!60!black!100] at ($(A13)+(0.9,1.45)$) {$X^{(1)}$};
    \node[red!60!black!100] at ($(A22)+(0.9,1.7)$) {$X^{(2)}$};
    \node[red!60!black!100] at ($(A33)+(0.9,1.45)$) {$X^{(3)}$};
    \node[red!60!black!100] at ($(A43)+(0.9,1.45)$) {$X^{(4)}$};
    \node[black!100] at ($(A52)+(0.9,1.7)$) {$X^{(5)}$};
    \draw[dashed, line width=0.5mm] ($(A00)+(0,-0.8)$) -- ($(A50)+(1.8,-0.8)$);
    \end{pgfonlayer}
    \begin{pgfonlayer}{fig}
        \gfm{4}{0}{-2.1}{0.0}{0}
        \bbconv{4}{0}{0.0}{0.15}
        \bbconv{4}{1}{0.8}{0.15}
        \bbconv{4}{2}{1.6}{0.15}
        \bbconv{4}{3}{2.4}{0.15}
        \gfm{4}{1}{4.4}{0.0}{0}
        \bbconv{4}{4}{6.1}{0.15}
        \bbconv{4}{5}{6.9}{0.15}
        \bbconv{4}{6}{7.7}{0.15}
        \fm{3}{2}{10.0}{0.0}{red!20}
        \bbconv{3}{7}{11.9}{0.15}
        \bbconv{3}{8}{12.7}{0.15}
        \bbconv{3}{9}{13.5}{0.15}
        \bbconv{3}{10}{14.3}{0.15}
        \fm{4}{3}{16.6}{0.0}{red!20}
        \bbconv{4}{11}{18.7}{0.15}
        \bbconv{4}{12}{19.5}{0.15}
        \bbconv{4}{13}{20.3}{0.15}
        \gfm{3}{4}{22.3}{0.0}{0}
        \bbconv{3}{14}{24.0}{0.15}
        \bbconv{3}{15}{24.8}{0.15}
        \bbconv{3}{16}{25.6}{0.15}
        \gfm{3}{5}{27.9}{0.0}{0}
        \coordinate (id) at (3.9,0.5);
        \coordinate (id2) at (9.5,0.5);
        \coordinate (id3) at (21.8,0.5);
        \coordinate (act) at (16.05, 0.5);
        \coordinate (act2) at (27.4, 0.5);
        \draw[black, thick] (id) circle [radius=0.35];
        \draw[black, thick] ($(id) + (0.35*1.414/2,0.35*1.414/2)$) --($(id) + (-0.35*1.414/2,-0.35*1.414/2)$);
        \node[black] at ($(id)+(0,+0.75)$) {$\mathrm{id}$};
        \draw[black, thick] (id2) circle [radius=0.35];
        \draw[black, thick] ($(id2) + (0.35*1.414/2,0.35*1.414/2)$) --($(id2) + (-0.35*1.414/2,-0.35*1.414/2)$);
        \node[black] at ($(id2)+(0,+0.75)$) {$\mathrm{id}$};
        \draw[black, thick] (id3) circle [radius=0.35];
        \draw[black, thick] ($(id3) + (0.35*1.414/2,0.35*1.414/2)$) --($(id3) + (-0.35*1.414/2,-0.35*1.414/2)$);
        \node[black] at ($(id3)+(0,+0.75)$) {$\mathrm{id}$};

        \draw[blue!60!black!100, thick] (act) circle [radius=0.35];
        \draw[blue!60!black!100, thick] ($(act) + (-0.35,0)$) -- (act);
        \draw[blue!60!black!100, thick] ($(act) + (0.35*1.414/2,0.35*1.414/2)$) -- (act);
        \node[blue!60!black!100] at ($(act)+(0,-0.9)$) {ReLU};
        \node[blue!60!black!100] at ($(act)+(0,+0.7)$) {$\sigma_3$};
        \draw[black, thick] (act2) circle [radius=0.35];
        \draw[black, thick] ($(act2) + (0.35*1.414/2,0.35*1.414/2)$) --($(act2) + (-0.35*1.414/2,-0.35*1.414/2)$);
        \node[black] at ($(act2)+(0,+0.75)$) {$\mathrm{id}$};
        \draw[rounded corners,line width=0.5mm,inner sep=20pt, black!100] ($(B60)+(1.0,-0.4)$) rectangle ($(B00)+(-0.2,1.8)$) {};
        \draw[rounded corners,line width=0.5mm,inner sep=20pt, black!100] ($(B160)+(1.0,-0.4)$) rectangle ($(B110)+(-0.2,1.7)$) {};

    \node[black!100] at ($(A03)+(0.9,1.45)$) {$X^{(0)}$};
    \node[black!100] at ($(A13)+(0.9,1.45)$) {$X^{(1)}$};
    \node[red!60!black!100] at ($(A22)+(0.9,1.7)$) {$X^{(2)}$};
    \node[red!60!black!100] at ($(A33)+(0.9,1.45)$) {$X^{(3)}$};
    \node[black!100] at ($(A43)+(0.9,1.45)$) {$X^{(4)}$};
    \node[black!100] at ($(A52)+(0.9,1.7)$) {$X^{(5)}$};
    \end{pgfonlayer}
 \begin{pgfonlayer}{bg}
    \draw[green!80!black!40] ($(DA00)+(-1,0)$) -- ($(DA04)+(-1,0)$) ;
    \draw[green!80!black!40] ($(DA10)+(-1,0)$) -- ($(DA14)+(-1,0)$) ;
    \draw[red!80!black!60] ($(DA20)+(-1,0)$) -- ($(DA23)+(-1,0)$) ;
    \draw[red!80!black!60] ($(DA30)+(-1,0)$) -- ($(DA34)+(-1,0)$) ;
    \draw[green!80!black!40] ($(DA40)+(-1,0)$) -- ($(DA43)+(-1,0)$) ;
    \draw[green!80!black!40] ($(DA50)+(-1,0)$) -- ($(DA53)+(-1,0)$) ;
    \end{pgfonlayer}
\begin{pgfonlayer}{fig}
    \gfm{4}{0}{0}{0.0}{-100}
    \bconvv{4}{1}{2.3}{0.0}
    \bconvv{4}{2}{3.9}{0.0}
    \bconvv{4}{3}{5.5}{0.0}
    \bfm{3}{1}{8.7}{0.0}{red!20}
    \bconv{3}{4}{10.6}{0.15}
    \bconv{3}{5}{11.6}{0.15}
    \bconv{3}{6}{12.6}{0.15}
    \bconv{3}{7}{13.6}{0.15}
    \bfm{4}{2}{15.9}{0.0}{red!20}
    \bconvv{4}{8}{18.2}{0.0}
    \bconvv{4}{9}{19.8}{0.0}
    \bconvv{4}{10}{21.4}{0.0}
    \gfm{3}{3}{24.6}{0.0}{-100}
        \coordinate (act) at (15.28, -3.02);
        \coordinate (act2) at (23.7, -3.02);
        \coordinate (id) at (7.8,-3.02);
        \draw[black, thick] (id) circle [radius=0.35];
        \draw[black, thick] ($(id) + (0.35*1.414/2,0.35*1.414/2)$) --($(id) + (-0.35*1.414/2,-0.35*1.414/2)$);
        \node[black] at ($(id)+(0,+0.75)$) {$\mathrm{id}$};
        
        \draw[blue!60!black!100, thick] (act) circle [radius=0.35];
        \draw[blue!60!black!100, thick] ($(act) + (-0.35,0)$) -- (act);
        \draw[blue!60!black!100, thick] ($(act) + (0.35*1.414/2,0.35*1.414/2)$) -- (act);
        \node[blue!60!black!100] at ($(act)+(0,-0.9)$) {ReLU};
        \node[blue!60!black!100] at ($(act)+(0,+0.7)$) {$\sigma_3$};
        
        \draw[black, thick] (act2) circle [radius=0.35];
        \draw[black, thick] ($(act2) + (0.35*1.414/2,0.35*1.414/2)$) --($(act2) + (-0.35*1.414/2,-0.35*1.414/2)$);
        \node[black] at ($(act2)+(0,+0.75)$) {$\mathrm{id}$};

    \draw[rounded corners,line width=0.5mm,inner sep=20pt, black!100] ($(EB30)+(1.6,-0.3)$) rectangle ($(EB10)+(-0.2,1.9)$) {};
    \draw[rounded corners,line width=0.5mm,inner sep=20pt, black!100] ($(EB100)+(1.6,-0.3)$) rectangle ($(EB80)+(-0.2,1.9)$) {};

    \node () at ($(B60)!0.5!(B00) + (1.85,-1.4)$) {$\theta_2 \circledast \theta_1\in\mathbb{R}^{4\times 3\times(2K-1)\times(2K-1)}$};
    \node () at ($(B100)!0.5!(B70) + (-0.55,-1.4)$) {$\theta_3\in\mathbb{R}^{3\times 4\times K\times K}$};
    \node () at ($(B160)!0.5!(B110) + (-0.5,-1.4)$) {$\theta_5 \circledast \theta_4\in\mathbb{R}^{4\times 3\times(2K-1)\times(2K-1)}$};
    \draw[->, line width=2pt,black!100] ($(B00)+(2.2,-0.4)$) to [out=-90,in=110] ($(EB10)+(0.2,2.0)$);
    \draw[->, line width=2pt,black!100] ($(B110)+(-0.2,0.2)$) to [out=240,in=110] ($(EB80)+(0.2,2.0)$);
    \node[black!100] () at ($(EB80)+(-0.3,2.6)$) {\LARGE $\circledast$};
    \node[black!100] () at ($(EB10)+(-0.3,2.6)$) {\LARGE $\circledast$};
    
    \node[black!100] at ($(A03)+(0.9,1.45)$) {$X^{(0)}$};
    \node[red!60!black!100] at ($(BA12)+(0.9,1.7)$) {$X^{(2)}$};
    \node[red!60!black!100] at ($(BA23)+(0.9,1.45)$) {$X^{(3)}$};
    \node[black!100] at ($(A32)+(0.9,1.7)$) {$X^{(5)}$};
    \end{pgfonlayer} 
    \begin{pgfonlayer}{bg}
    \draw[green!80!black!40] ($(DA00)+(-1,0)$) -- ($(DA04)+(-1,0)$) ;
    \draw[red!80!black!60] ($(DBA10)+(-1,0)$) -- ($(DBA13)+(-1,0)$) ;
    \draw[red!80!black!60] ($(DBA20)+(-1,0)$) -- ($(DBA24)+(-1,0)$) ;
    \draw[green!80!black!40] ($(DA30)+(-1,0)$) -- ($(DA33)+(-1,0)$) ;
    \node[align=left] () at (28,-2.5) {\textcolor{red!60!black!100}{$A=\{3\}$}\\\textcolor{blue!60!black!100}{$S=\{2,3\}$}};
    \end{pgfonlayer}
\end{tikzpicture}
}

\caption{Illustration of depth compression for a five-layer CNN, $\BigO_{l=1}^5 \sigma_l \circ f_{\theta_l}$. The original network is
on the default setting when $A=S=\{1,2,3,4\}$ (above). When $A=\{3\}$ and $S=\{2,3\}$, the activation layer not in $A$ is replaced with identity functions (middle). Then, the network is merged into the shallow network which functions identically (below). 
}
\label{fig:conv}
\vspace{-0.5em}
\end{figure*}

Although DepthShrinker has shown promising results in reducing the depth of the network while preserving the performance, the method is limited to constricted search space as it only considers merging within the Inverted Residual Block \citep{depthshrinker, mobilenet}. Furthermore, the method relies on human-engineered heuristics for layer merging which is unlikely to scale to other architectures. To this end, we introduce a novel optimization-based framework for general convolution merging framework that is not restricted to the design of the network and does not rely on manually designed heuristics. We formulate a depth compression optimization problem that replaces inefficient activation layers with identity functions and optimally merges consecutive convolution operations for optimal latency.

Our optimization problem is NP-Hard and its objective requires a prohibitively exhaustive training of the neural network. Thus, we formulate a surrogate optimization problem by approximating the objective as the linear sum of the accuracy change induced by each network block. The surrogate optimization problem can be exactly optimized via dynamic programming on a given network architecture with a given latency. Furthermore, we evaluate the latency of the network with TensorRT for a fair comparison \citep{tensorrt}. Our experiments show that the proposed method outperforms the baseline method in both the accuracy and the inference latency in MobileNetV2 on ImageNet dataset. We release the code at {\small\url{https://github.com/snu-mllab/Efficient-CNN-Depth-Compression}}.


\section{Related Works}

\paragraph{Channel Pruning} 
Channel pruning originally aims to reduce computation FLOPs by removing less important channels \citep{pfec,fpgm,sfp, metaprun,amc,gate,apoz,perfmax}. Specifically, \citet{knapsack} formulate a knapsack problem for channel pruning with an explicit FLOPs constraint. For practitioners, however, end-to-end inference wall-clock time is the most important metric. In light of this, \citet{halp} build a latency lookup table and proposes a knapsack problem for channel pruning with a latency constraint.

\paragraph{Network Morphism} 
Our work is partially inspired by network morphism which morphs a trained parent network into a child network that functions identically \citep{net2net,netmorph}. Here, the child network is larger than the parent network and is finetuned after morphing. Instead, we aim to find the parent network where some activation layers are removed, thereby morphing the parent network into the child network which has a faster inference time and functions almost identically to the parent network.

\paragraph{Depth Reduction} 
There are two lines of research that reduce the depth of neural networks: layer-pruning and depth compression. In layer pruning, \citet{layerpruning} and \citet{layerpruning2} evaluate the importance of layers by the amount of discriminative information in each feature map. In depth compression, DepthShrinker points out the inefficiency of depth-wise convolutions during inference in the edge device and proposes a depth compression algorithm that replaces inefficient consecutive depth-wise convolution and point-wise convolution inside the Inverted Residual Block with an efficient dense convolution \citep{depthshrinker,depthconv,mobilenet}. We generalize depth compression space to cover any general convolution operations. Also, while DepthShrinker requires full training of the network during the search phase to identify the 
unnecessary activations, our method employs importance evaluation which can be efficiently computed in an embarrasingly parallel fashion. Furthermore, we propose a novel two-stage dynamic programming algorithm which simultaneously finds the optimal set of selected activation layers and the optimal set of layers to be merged in a few seconds.

\paragraph{TensorRT} 
Choosing the appropriate implementation of the network to measure the inference latency is crucial for a fair comparison. For instance, a batch normalization (BN) module can be fused into the preceding convolution layer without compromising accuracy and accelerating the inference latency. In this regard, we utilize TensorRT to optimize trained network architectures with various techniques such as BN fusion, precision calibration and dynamic memory management \citep{tensorrt}. 

\section{Preliminary} \label{prelim}
Consider a $L$-layer CNN which consists of alternating convolution layer $f_{\theta_l}$ and activation layer $\sigma_l$ with the layer index $l \in [L]$. Each convolution layer is parametrized by convolution kernel parameter $\theta_l\in \mathbb{R}^{ C_{l-1} \times C_{l}\times K_l \times K_l}$, where $C_{l-1}, C_l, K_l$ represent the number of input channels, the number of output channels, and the kernel size, respectively. The CNN can be represented as a composite function $\BigO_{l=1}^L \sigma_l \circ f_{\theta_l} : \mathbb{R}^{H_0\times W_0 \times K_0 \times K_0}\to \mathbb{R}^{H_L \times W_L \times K_L\times K_L}$, where $H_l,W_l$ are the height and width of $l$-th feature map and $\BigO$ denotes an iterated function composition. We set the last activation layer ($\sigma_L$) to identity function ($\mathrm{id}$).

Note, any consecutive convolution operations can be replaced by an equivalent convolution operation with a larger kernel due to the associative property. We denote this process as \emph{merging}. For example, consider two consecutive convolutional layers, $f_{\theta_1}$ and $f_{\theta_2}$, applied to an input image $X$ (\emph{i.e.} $f_{\theta_2}(f_{\theta_1}(X))$). This can also be computed using an equivalent \emph{merged} convolutional layer $f_{\theta_2 \circledast \theta_1}$ where $\circledast$ denotes convolution with proper padding. Further merging details can be found in \Cref{subsec:mergedetail}.




\section{Method} \label{method}
We first formulate an optimal subset selection problem for depth compression under a given latency constraint. Subsequently, we propose a surrogate objective for the objective in the optimal subset selection problem and formulate a corresponding surrogate optimization problem, which can be exactly solved via two-stage dynamic programming (DP).

\subsection{Optimal Subset Selection Problem for Depth Compression}
Any neighboring convolutional layers can be merged into an equivalent convolutional layer, often resulting in a latency speed-up from the depth compression of the CNN. In this regard, we aim to optimize the replacement of a subset of activation layers with $\mathrm{id}$ in order to reduce the latency of the resulting network while preserving its performance. 

However, merging every consecutive series of convolutional layers into a single large layer may not be the optimal merge in terms of latency. In certain cases, it is possible that merging certain convolutional layers has a detrimental effect on the latency of a network. To illustrate, consider merging two consecutive $1\times1$ convolutional layers, with the first layer having $100$ input channels and $1$ output channel and the second layer having $1$ input channel and $100$ output channels. Then the merged convolutional layer results in a $1\times1$ convolution with $100$ input channels and $100$ output channels. This merge significantly increases the latency of a merged convolutional layer, thereby canceling out any benefits gained from the depth compression.

To address this, we propose two \emph{ordered set} variables, $A$ and $S$ to be simultaneously optimized. $A$ indicates the layer indices where the activation layer is kept intact and not replaced with an identity function, and $S$ indicates the layer indices where we do not merge. It is important to note that $S$ includes $A$, since the activation layers that are not $\mathrm{id}$ can not be merged. \Cref{fig:conv} illustrates how network is merged according to $S$ and $A$. Our goal is to optimize for the ordered set $A$ and $S$ in order to reduce the latency of the resulting network while preserving its performance.

Thus, our objective can be formulated as follows: \footnotesize
\begin{subequations}
\begin{align}
    &\underset{A\subseteq S \subseteq [L-1]}{\mathrm{maximize}} ~~\underset{\theta}{\mathrm{max}}\;\mathrm{Acc}\left( 
    \BigO_{l=1}^L \left( \mathds{1}_A(l)\sigma_l + \left(1-\mathds{1}_A(l)\right) \mathrm{id} \right) \circ f_{\theta_l} \right)\label{eq:opt1_obj}\\
    &\mathrm{subject\ to\ }\nonumber \\
    &\qquad T\left( \BigO_{i=1}^{|S|+1} \left( \mathds{1}_A(s_i)\sigma_{s_i} + \left(1-\mathds{1}_A(s_i)\right) \mathrm{id} \right) \circ f_{\hat{\theta}_i} \right) < T_0 \label{eq:opt1_const}\\\
    &\qquad \hat{\theta}_i= \Conv_{l=s_{i\!-\!1}\!+\!1}^{s_i} \theta_l, ~ \forall i \in [|S|+1] ~~\text{and} ~~s_{|S|+1}=L, ~s_0=0,\nonumber
\end{align}
\label{eq:opt1}
\end{subequations}
\normalsize
\hspace{-0.2em}where $(s_i)_{i=1}^{|S|}$ denotes the elements of $S$ and $\mathrm{Acc}(\cdot)$ and $T(\cdot)$ denote the accuracy and latency of the network, respectively. The objective in \Cref{eq:opt1_obj} describes the accuracy of the network where the activation layer is replaced by $\mathrm{id}$ if the layer index is not in $A$. The constraint in \Cref{eq:opt1_const} describes the latency constraint of the network, which is merged according to $S$. Note, the networks in \Cref{eq:opt1_obj} and  \Cref{eq:opt1_const} function identically.

We can simplify the constraint in \Cref{eq:opt1_const} by expressing the total latency of the network as the sum of the latency of each merged convolution layer as each layer is sequentially connected. Additionally, we ignore the latency from each activation layer since the latency incurred by activation layers is negligible\footnote{Deactivating $50$ ReLUs in MobileNetV2 results in less than a 1\% change in end-to-end inference time on \textit{RTX 2080 Ti} in TensorRT format.
}. Then, the total latency of the merged network in the constraint can be simplified as follows:

\scriptsize
\begin{align*}
    &T\left( \BigO_{i=1}^{|S|+1} \left( \mathds{1}_A(s_i)\sigma_l + \left(1-\mathds{1}_A(s_i)\right) \mathrm{id} \right) \circ f_{\hat{\theta}_i} \right)\\
    &\approx\sum_{i=1}^{|S|+1} T\left( \left( \mathds{1}_A(s_i)\sigma_l + \left(1-\mathds{1}_A(s_i)\right) \mathrm{id} \right) \circ f_{\hat{\theta}_i} \right)\approx\sum_{i=1}^{|S|+1} T\left( f_{\hat{\theta}_i} \right).
\end{align*}
\normalsize

As a shorthand, we denote $T(f_{\theta'})$ 
where $\theta'= \conv_{l=i+1}^{j} \theta_l$ as $T[i,j]$. Note that $f_{\theta'}$ is the merged convolution operation equivalent to $\bigovoid_{l=i+1}^j f_{\theta_l}$. Thus, the latency constraint, \Cref{eq:opt1_const}, can be compactly expressed as $\sum_{s_{i\!-\!1},s_i \in \{0\}\cup S \cup \{L\}} T[s_{i\!-\!1},s_i]<T_0$.


\subsection{Formulation with the Surrogate Objective}
\label{sec:dp}
Directly optimizing \Cref{eq:opt1} requires training of the whole network for all combinations of $A$ and $S$ which is NP-hard. Therefore, we propose a surrogate objective for the objective in \Cref{eq:opt1_obj}. Through this approach, \Cref{eq:opt1} can be reformulated into an optimal subset selection problem which can be exactly solved via DP. 

The network in \Cref{eq:opt1_obj} can be equivalently represented as $\BigO_{j=1}^{|A|+1} \sigma_{a_j} \circ \left( \BigO_{l=a_{j\!-\!1}+1}^{a_j} f_{\theta_l} \right)$ where $a_0=0, a_{|A|+1}=L$, and $(a_j)_{j=1}^{|A|}$ denote the elements of $A$ in the ascending order. We can observe that $A$ partitions the network into contiguous network blocks, $\BigO_{l=a_{i\!-\!1}\!+\!1}^{a_i} f_{\theta_l}$. Therefore, \Cref{eq:opt1_obj} can be reformulated as the accuracy change from the original network as follows:
\footnotesize
\begin{align}
\label{eq:optacc}
\underset{A\subseteq S \subseteq [L-1]}{\mathrm{maximize}}& \max_{\theta}~\mathrm{Acc}\left(\BigO_{j=1}^{|A|+1} \sigma_{a_j} \circ \left( \BigO_{l=a_{j\!-\!1}+1}^{a_j} f_{\theta_l} \right)\right)\nonumber\\
&-\max_\theta
\mathrm{Acc}~\left(\BigO_{l=1}^L \sigma_l \circ f_{\theta_l} \right),
\end{align}
\normalsizewhere $ \max_\theta\mathrm{Acc}~\left(\BigO_{l=1}^L \sigma_l \circ f_{\theta_l} \right)$ is the accuracy of the original network which is a constant. 

Each contiguous block results in an accuracy change from the original network. However, the exact estimation of the accuracy change resulting from all possible combinations of contiguous network blocks remains impractical due to the exponential number of possible combinations of contiguous network blocks and the requirement of training the neural network for each one. Therefore, we propose the sum of the accuracy change caused by each contiguous network block, $\BigO_{l=a_{i\!-\!1}\!+\!1}^{a_i} f_{\theta_l}$ as a proxy for the accuracy change resulting from contiguous network blocks in \Cref{eq:optacc}.

We denote $I[i,j]$ as the accuracy change when the activation layers between the $i\!+\!1$-th and $j$-th layers in the original network are replaced with $\mathrm{id}$. Concretely, 

\vspace{-2em}
\scriptsize
\begin{align}
\label{eq:imp}
  I[i, j] &\coloneqq \max_\theta
          \mathrm{Acc}\left(
          \underbrace{\BigO_{l=j+1}^{L} \sigma_l \circ f_{\theta_l}}_{\text{$j\!+\!1$ to$L$ layers}}
          \circ \underbrace{\sigma_j
          \circ \BigO_{l=i+1}^{j} f_{\theta_l}}_{\text{$i\!+\!1$ to $j$ layers}}
          \circ \underbrace{\BigO_{l=1}^{i}  \sigma_l \circ f_{\theta_l} }_{\text{$1$
        to $i$ layers}}\right)\nonumber\\
        &- \max_\theta
 \mathrm{Acc}\left( 
    \BigO_{l=1}^L \sigma_l \circ f_{\theta_l} \right).
\end{align}
\normalsizeNote that computing $I[\cdot, \cdot]$ can be efficiently done in embarrassingly parallel fashion. We define the surrogate objective for \Cref{eq:opt1_obj} as $\sum_{a_{j-1},a_j \in \{0\}\cup A\cup \{L\}} I[a_{j-1}, a_{j}]$. Then the optimization problem in \Cref{eq:opt1} becomes \begin{align}
    \label{eq:opt3}
&\underset{A\subseteq S \subseteq [L-1]}{\mathrm{maximize}} \;\; \sum_{a_{j\!-\!1},a_j \in \{0\}\cup A\cup \{L\}} I[a_{j\!-\!1}, a_{j}]\\
&\mathrm{subject\ to\ } ~~ \sum_{s_{i\!-\!1},s_i \in \{0\}\cup S\cup \{L\}} T[s_{i\!-\!1}, s_{i}] < T_0. \nonumber
\end{align}

\subsection{Optimization via Dynamic Programming}
We first define an ordered set for the indices to be merged for the optimal inference time for the contiguous network block between $k\!+\!1$-th layer and $l$-th layer as $S_\text{opt}[k,l]$ and the optimal inference time as $T_\text{opt}[k,l]$. Concretely, \begin{subequations}
\begin{align}
    T_{\text{opt}}[k,l] &\coloneqq \min_{S \subseteq \{ k\!+\!1,\ldots,l\!-\!1\}} \sum_{s_{i\!-\!1},s_i \in \{k\}\cup S\cup \{l\}} T[s_{i\!-\!1}, s_{i}] \label{eq:timedp1}\\
    S_{\text{opt}}[k,l] &\coloneqq \underset{S \subseteq \{ k\!+\!1,\ldots,l\!-\!1\}}{\mathrm{argmin}}  \sum_{s_{i\!-\!1},s_i \in \{k\}\cup S\cup \{l\}} T[s_{i\!-\!1}, s_{i}]\label{eq:timedp2}.
\end{align}
\end{subequations}For the base case, $T_\text{opt}[k,k] = 0$ and $S_\text{opt}[k,k]=\emptyset$. Then, $T_\text{opt}[k,l]$ and $S_\text{opt}[k,l]$ can be computed via dynamic programming algorithm as described in \Cref{alg:timeDP}.

We formulate a sub-optimization problem of \Cref{eq:opt3} with respect to an  intermediate layer index, $l\leq L$ and a latency constraint, $t>T_\text{opt}[0,l]$: \begin{align}
    \label{eq:opt4}
 &\underset{A\subseteq S \subseteq [l\!-\!1]}{\mathrm{maximize}} \;\; \sum_{a_{j\!-\!1},a_j \in \{0\}\cup A\cup \{l\}} I[a_{j\!-\!1}, a_{j}]\\
&\mathrm{subject\ to\ } ~~ \sum_{s_{i\!-\!1},s_i \in \{0\}\cup S\cup \{l\}} T[s_{i\!-\!1}, s_{i}] < t. \nonumber
\end{align}

 Then, we define the optimal ordered sets $A$ and $S$ in the sub-optimization problem as $A[l,t]$ and $S[l,t]$. Here, $A[l,t]$ indicates activation layers to keep until layer $l$ given the latency budget $t$, and $S[l,t]$ indicates layers to merge until layer $l$ given the latency budget $t$. Then, $A[L,T_0]$ and $S[L,T_0]$ represent the optimal set $A$ and $S$ of the surrogate optimization problem, \Cref{eq:opt3}, respectively.

For the base case, we set $A[0,t]=S[0,t]=\emptyset$. Then, we compute the ordered sets $A[l,t]$ and $S[l,t]$ according to the dynamic programming (DP) recurrence relation defined by 

\footnotesize
\begin{subequations}
\begin{align}
    A[l, t] &=  A[k, t\!- \!T_\text{opt}[k,l]] \cup \{k:k>0\}
    \label{eq:dprecura}\\
    S[l, t] &= S[k, t\!- \!T_\text{opt}[k,l]] \cup \{k:k>0\}\cup S_\text{opt}[k,l],
    \label{eq:dprecurs}\\
    k&= \underset{0\leq k'<l}{\mathrm{argmax}} ~ 
    \sum_{a_{j\!-\!1},a_j \in \{0\}\cup A[k', t\!-\!T_\text{opt}[k',l]]\cup \{k',l\}} I[a_{j\!-\!1}, a_{j}]
\nonumber\\
    &\quad \mathrm{subject\ to\ } ~~T_\text{opt}[0,k'] +T_\text{opt}[k',l]< t,\nonumber
\end{align}
\label{eq:dprecur}
\end{subequations}
\normalsize\hspace{-0.3em}where $k$ is the maximum element of $A[l,t]$. Therefore, $A[l,t]$ is an empty set when $k\!=\!0$. \Cref{fig:dp} illustrates DP computation example in detail.

The $j$-th activation layer of our target network for depth compression is either the $j$-th activation layer in the vanilla network ($\sigma_j$) or an identity function ($\mathrm{id}$). Thus, if $\sigma_j=\mathrm{id}$, the $j$-th activation layer in the target network is inherently an identity function. For instance, in MobileNetV2, the identity function serves as the activation layer at the end of each Inverted Residual Block, and the corresponding activation layers in our target network are bound to be the identity functions \citep{mobilenet}. On the other hand, non-linear activation layers at the end of the Inverted Residual Block can improve the performance of the networks compressed from MobileNetV2 \citep{depthshrinker}. To this end, we incorporate the network blocks that have non-linear activation layers at the end of the Inverted Residual Block into the DP formulation. We provide a detailed explanation in \Cref{app:imp-ext}.

\begin{algorithm}[t]
\caption{Finding Optimal Latency with DP}
\label{alg:timeDP}
\begin{algorithmic}
\INPUT $T, L$ 
\STATE Initialize $T_{\text{opt}}[k, l] \leftarrow 0, S_{\text{opt}}[k, l] \leftarrow \emptyset$ for $0\leq k\leq l \leq L$
\FOR{$l = 1$ {\bfseries to} $L$}
    \FOR{$k = 0$ {\bfseries to} $l\!-\!1$}
        \STATE $m \leftarrow 
        \underset{k\leq m'< l}{ \mathrm{argmin} } 
        \left(T_{\text{opt}}[k, m']+ T[m', l] \right)$
        \STATE $T_{\text{opt}}[k, l] \leftarrow T_{\text{opt}}[k, m] + T[m, l]$
        \IF{$m \notin \{k\}$}
        \STATE $S_{\text{opt}}[k, l] \leftarrow S_{\text{opt}}[k, m] \cup \{m\}$
        \ENDIF
    \ENDFOR
\ENDFOR
\OUTPUT $T_{\text{opt}}$, $S_{\text{opt}}$
\end{algorithmic}
\end{algorithm}

\begin{algorithm}[t]
\caption{Solving the Surrogate Objective with DP}
\label{alg:optDP}
\begin{algorithmic}
\INPUT $T_0, L, T, I$ 
\STATE Initialize $D[l,t] \leftarrow  0, A[l, t] \leftarrow \emptyset, S[l, t] \leftarrow \emptyset \quad \forall t,l$
\STATE $T_{\text{opt}}, S_{\text{opt}} \leftarrow \text{\Cref{alg:timeDP} }(T, L)$
\FOR{$l = 1$ {\bfseries to} $L$}
\FOR{$t=T_{\text{opt}}[0,l]+1$ {\bfseries to} $T_0$}
    \STATE $k \leftarrow  \underset{0\leq k' <l}{\mathrm{argmax}} \left(
    D[k', t - T_{\text{opt}}[k', l]] + I(k', l) \right)$
    \STATE $\qquad \mathrm{subject\ to\ }  ~T_\text{opt}[0,k'] +T_\text{opt}[k',l]< t$
    \STATE $t_{\text{last}} \leftarrow T_{\text{opt}}[k, l]$
    \STATE $D[l, t] \leftarrow D[k, t -t_{\text{last}}] + I[k, l]$
    \STATE $A[l, t] \leftarrow A[k, t- t_{\text{last}}] \cup \{k:k>0\}$
    \STATE $S[l, t] \leftarrow S[k, t- t_{\text{last}}] \cup \{k:k>0\} \cup S_{\text{opt}}[k, l]$
\ENDFOR 
\ENDFOR
\OUTPUT $A[L, T_0], S[L, T_0]$
\end{algorithmic}
\end{algorithm}

\subsection{Theoretical Analysis}
\Cref{prop:dp} shows that \Cref{eq:dprecur} exactly computes $A[l,t]$ and $S[l,t]$. The detailed procedure for implementing the DP recurrence relation can be found in \Cref{alg:optDP}\footnote{We denote $\sum_{a_{j\!-\!1},a_j \in \{0\}\cup A[l,t]\cup \{l\}} I[a_{j\!-\!1}, a_{j}]$ as $D[l,t]$ in \Cref{alg:optDP} for brevity.}. At the start of \Cref{alg:optDP}, we compute the ordered set for the indices to be merged for the optimal inference time for the contiguous network block between $k\!+\!1$-th layer and $l$-th layer and the optimal inference time at \Cref{alg:timeDP} where the time complexity for the DP recurrences is $\mathcal{O}(L^3)$. In \Cref{alg:optDP}, the time complexity for the DP recurrences is $\mathcal{O}(L^2T_0)$, thus the total time complexity is $\mathcal{O}(L^3 + L^2T_0)$. 

\newcommand\tablestructure[4]{
    \def\W{#1}
    \def\H{#2}
    \def\IDID{#3}
    \def\ccc{#4}
    
    \foreach \x in {0,...,\W} {
        \foreach \y in {0,...,\H} {
            \coordinate (\IDID\x\y) at (\xx + \x * \gW, \yy - \y * \gH);
        }   
    }
    \draw[fill, \ccc] (\IDID00) rectangle (\IDID1\H);
    \draw[fill, \ccc] (\IDID00) rectangle (\IDID\W1);
    \foreach \x in {2,...,\W}{
        \foreach \y in {\x,...,\H}{
            \draw ($(\IDID\x\y) + (-\gW, +\gH)$) -- (\IDID\x\y);
        }
    }
    \draw ($(\IDID00) + (0, 0)$) -- ($(\IDID00) + (\gW, -\gH)$);
    \foreach \x in {2,...,\W}{
        \draw ($(\IDID\x0)+ (-\gW,0)$) -- ($(\IDID\x\W)+ (-\gW,0)$);
    }
    \foreach \y in {2,...,\H}{
        \draw ($(\IDID0\y)+ (0,\gH)$) -- ($(\IDID\H\y)+ (0,\gH)$);
    }
}
\newcommand\tablestructureTP[4]{
    \def\W{#1}
    \def\H{#2}
    \def\IDID{#3}
    \def\ccc{#4}
    
    \foreach \y in {0,...,\W} {
        \foreach \x in {0,...,\H} {
            \coordinate (\IDID\y\x) at (\xx + \x * \gW, \yy - \y * \gH);
        }   
    }
    \draw[fill, \ccc] (\IDID00) rectangle (\IDID1\H);
    \draw[fill, \ccc] (\IDID00) rectangle (\IDID\W1);
    \foreach \y in {2,...,\W}{
        \foreach \x in {\y,...,\H}{
            \ifnum\x=\y
            \node () at ($(\IDID\y\x) + (-0.5*\gW, +0.5*\gH)$) {0};
            \else
            \draw ($(\IDID\y\x) + (-\gW, +\gH)$) -- (\IDID\y\x);
            \fi
        }
    }
    \draw ($(\IDID00) + (0, 0)$) -- ($(\IDID00) + (\gW, -\gH)$);
    \foreach \y in {2,...,\W}{
        \draw ($(\IDID\y0)+ (0,\gH)$) -- ($(\IDID\y\W)+ (0,\gH)$);
    }
    \foreach \x in {2,...,\H}{
        \draw ($(\IDID0\x)+ (-\gW,0)$) -- ($(\IDID\H\x)+ (-\gW,0)$);
    }
}
\newcommand\putvalue[4]{
\def\i{#1}
\def\j{#2}
\def\CURID{#3}
\def\val{#4}
\node () at ($(\CURID\i\j) + (0.5*\gW, -0.5*\gH)$) {\val};
}
\newcommand\putvaluevdots[4]{
\def\i{#1}
\def\j{#2}
\def\CURID{#3}
\def\val{#4}
\node () at ($(\CURID\i\j) + (0.5*\gW, -0.5*\gH-0.1)$) {\val};
}
\newcommand\putvalueunderline[4]{
\def\i{#1}
\def\j{#2}
\def\CURID{#3}
\def\val{#4}
\node () at ($(\CURID\i\j) + (0.5*\gW, -0.5*\gH-0.04)$) {\val};
}
\newcommand\putvalueTP[4]{
\def\i{#1}
\def\j{#2}
\def\CURID{#3}
\def\val{#4}
\node () at ($(\CURID\j\i) + (0.5*\gW, -0.5*\gH)$) {\val};
}
\newcommand\tablestructureC[4]{
    \def\W{#1}
    \def\H{#2}
    \def\IDID{#3}
    \def\ccc{#4}
    
    \foreach \x in {0,...,\W} {
        \foreach \y in {0,...,\H} {
            \coordinate (\IDID\x\y) at (\xx + \x * \gW, \yy - \y * \gH);
        }   
    }
    \draw[fill, \ccc] (\IDID00) rectangle (\IDID1\H);
    \draw[fill, \ccc] (\IDID00) rectangle (\IDID\W1);
    \draw ($(\IDID00) + (0, 0)$) -- ($(\IDID00) + (\gW, -\gH)$);
    \foreach \x in {2,...,\W}{
        \draw ($(\IDID\x0)+ (-\gW,0)$) -- ($(\IDID\x\H)+ (-\gW,0)$);
    }
    \foreach \y in {2,...,\H}{
        \draw ($(\IDID0\y)+ (0,\gH)$) -- ($(\IDID\W\y)+ (0,\gH)$);
    }
}
\newcommand\tablestructureCTP[4]{
    \def\W{#1}
    \def\H{#2}
    \def\IDID{#3}
    \def\ccc{#4}
    
    \foreach \y in {0,1,2,3} {
        \foreach \x in {0,...,\H} {
            \coordinate (\IDID\y\x) at (\xx + \x * \gW, \yy - \y * \gH);
        }   
    }
    \foreach \y in {4,5} {
        \foreach \x in {0,...,\H} {
            \coordinate (\IDID\y\x) at (\xx + \x * \gW, \yy - \y * \gH-0.4);
        }   
    }
    \foreach \y in {6,7} {
        \foreach \x in {0,...,\H} {
            \coordinate (\IDID\y\x) at (\xx + \x * \gW, \yy - \y * \gH-0.8);
        }   
    }
    \foreach \y in {8,9} {
        \foreach \x in {0,...,\H} {
            \coordinate (\IDID\y\x) at (\xx + \x * \gW, \yy - \y * \gH-1.2);
        }   
    }
    
    \draw[fill, \ccc] (\IDID00) rectangle (\IDID1\H);
    \draw[fill, \ccc] (\IDID00) rectangle (\IDID\W1);
    \draw ($(\IDID00) + (0, 0)$) -- ($(\IDID00) + (\gW, -\gH)$);
    \foreach \y in {1,...,\W}{
        \ifnum\y<\W
        \draw ($(\IDID\y0)$) -- ($(\IDID\y\H)$);
        \fi
    }
    \foreach \x in {2,...,\H}{
        \draw ($(\IDID0\x)+ (-\gW,0)$) -- ($(\IDID\W\x)+ (-\gW,0)$);
    }
}
\newcommand\tableA[2]{
\def\xx{0}
\def\yy{3.8}
\tablestructureTP{5}{5}{A}{gray!20}
\putvalue{1}{0}{A}{0}
\putvalue{2}{0}{A}{1}
\putvalue{3}{0}{A}{2}
\putvalue{4}{0}{A}{3}
\putvalue{0}{1}{A}{0}
\putvalue{0}{2}{A}{1}
\putvalue{0}{3}{A}{2}
\putvalue{0}{4}{A}{3}

\putvalue{2}{1}{A}{9}
\putvalue{3}{1}{A}{12}
\putvalue{3}{2}{A}{6}
\putvalue{4}{1}{A}{\textcolor{green!50!black!100}{20}}
\putvalue{4}{2}{A}{\textcolor{blue!80!black!100}{14}}
\putvalue{4}{3}{A}{\textcolor{red}{11}}
}
\newcommand\tableB[2]{
\def\xx{4.5}
\def\yy{3.8}
\tablestructureTP{5}{5}{B}{gray!20}
\putvalue{1}{0}{B}{0}
\putvalue{2}{0}{B}{1}
\putvalue{3}{0}{B}{2}
\putvalue{4}{0}{B}{3}
\putvalue{0}{1}{B}{0}
\putvalue{0}{2}{B}{1}
\putvalue{0}{3}{B}{2}
\putvalue{0}{4}{B}{3}

\putvalue{2}{1}{B}{0.5}
\putvalue{3}{1}{B}{0.9}
\putvalue{3}{2}{B}{0.3}
\putvalue{4}{1}{B}{\textcolor{green!50!black!100}{1.8}}
\putvalue{4}{2}{B}{\textcolor{blue!80!black!100}{1.4}}
\putvalue{4}{3}{B}{\textcolor{red}{0.7}}
}
\newcommand\tableC[2]{
\def\xx{0}
\def\yy{0}
\tablestructureCTP{9}{5}{C}{red!10}
\putvalueTP{1}{0}{C}{0}
\putvalueTP{2}{0}{C}{1}
\foreach \k in {0,...,4}{
    \ifnum\k>1
    \putvalue{1}{\k}{C}{0}
    \putvalue{2}{\k}{C}{0}
    \fi
    \ifnum\k=1
    \putvalue{1}{1}{C}{0}
    \putvalue{2}{1}{C}{\textcolor{green!50!black!100}{0}}
    \fi
    \putvaluevdots{3}{\k}{C}{$\vdots$}
    \putvaluevdots{5}{\k}{C}{$\vdots$}
    \putvaluevdots{7}{\k}{C}{$\vdots$}
}
\putvalue{1}{0}{C}{0}
\putvalueunderline{2}{0}{C}{\textcolor{green!50!black!100}{\underline{1}}}
\putvalueunderline{4}{0}{C}{\textcolor{blue!80!black!100}{\underline{7}}}
\putvalueunderline{6}{0}{C}{\textcolor{red}{\underline{10}}}
\putvalue{8}{0}{C}{21}
\putvalue{0}{1}{C}{0}
\putvalue{0}{2}{C}{1}
\putvalue{0}{3}{C}{2}
\putvalue{0}{4}{C}{3}

\putvalue{4}{1}{C}{0}
\putvalue{4}{2}{C}{\textcolor{blue!80!black!100}{0.5}}
\putvalue{4}{3}{C}{0}
\putvalue{4}{4}{C}{0}
\putvalue{6}{1}{C}{0}
\putvalue{6}{2}{C}{0.5}
\putvalue{6}{3}{C}{\textcolor{red}{0.8}}
\putvalue{6}{4}{C}{0}
\putvalue{8}{1}{C}{0}
\putvalue{8}{2}{C}{0.5}
\putvalue{8}{3}{C}{0.9}
\putvalue{8}{4}{C}{\textbf{?}}

}

\newcommand\tableD[2]{
\def\xx{4.5}
\def\yy{0}
\tablestructureCTP{9}{5}{D}{red!10}
\putvalueTP{1}{0}{D}{0}
\putvalueTP{2}{0}{D}{1}
\foreach \k in {0,...,4}{
    \ifnum\k>1
    \putvalue{1}{\k}{D}{$\emptyset$}
    \putvalue{2}{\k}{D}{$\emptyset$}
    \fi
    \ifnum\k=1
    \putvalue{1}{1}{D}{$\emptyset$}
    \putvalue{2}{1}{D}{\textcolor{green!50!black!100}{$\emptyset$}}
    \fi
    \putvaluevdots{3}{\k}{D}{$\vdots$}
    \putvaluevdots{5}{\k}{D}{$\vdots$}
    \putvaluevdots{7}{\k}{D}{$\vdots$}
}
\putvalue{1}{0}{D}{0}
\putvalueunderline{2}{0}{D}{\textcolor{green!50!black!100}{\underline{1}}}
\putvalueunderline{4}{0}{D}{\textcolor{blue!80!black!100}{\underline{7}}}
\putvalueunderline{6}{0}{D}{\textcolor{red}{\underline{10}}}
\putvalue{8}{0}{D}{21}
\putvalue{0}{1}{D}{0}
\putvalue{0}{2}{D}{1}
\putvalue{0}{3}{D}{2}
\putvalue{0}{4}{D}{3}

\putvalue{4}{1}{D}{$\emptyset$}
\putvalue{4}{2}{D}{\textcolor{blue!80!black!100}{$\emptyset$}}
\putvalue{4}{3}{D}{$\emptyset$}
\putvalue{4}{4}{D}{$\emptyset$}
\putvalue{6}{1}{D}{$\emptyset$}
\putvalue{6}{2}{D}{$\emptyset$}
\putvalue{6}{3}{D}{\textcolor{red}{$\{1\}$}}
\putvalue{6}{4}{D}{$\emptyset$}
\putvalue{8}{1}{D}{$\emptyset$}
\putvalue{8}{2}{D}{$\emptyset$}
\putvalue{8}{3}{D}{$\emptyset$}
\putvalue{8}{4}{D}{\textbf{?}}

}

\newcommand\colorrec[2]{
\def\p{#1}
\def\ccc{#2}
\draw[\ccc,line width=2pt] ($(\p)+(1pt, 1pt)$) rectangle ($($(\p)+(\gW, \gH)$) + (-1pt,-1pt)$);
}
\newcommand\colorrecdash[2]{
\def\p{#1}
\def\ccc{#2}
\draw[\ccc,dashed, line width=2pt] ($(\p)+(1pt, 1pt)$) rectangle ($($(\p)+(\gW, \gH)$) + (-1pt,-1pt)$);
}
\begin{figure}[t]
\resizebox{\columnwidth}{!}{
\begin{tikzpicture}
\def\gW{0.8}
\def\gH{0.6}

\tableA{0}{0}
\tableB{0}{0}
\tableC{0}{0}
\tableD{0}{0}
\node () at ($(A00) + (0.2*\gW, -0.7*\gH)$) {$l$};
\node () at ($(A00) + (0.7*\gW, -0.25*\gH)$) {$k$};
\node () at ($(B00) + (0.2*\gW, -0.7*\gH)$) {$l$};
\node () at ($(B00) + (0.7*\gW, -0.25*\gH)$) {$k$};
\node () at ($(C00) + (0.2*\gW, -0.7*\gH)$) {$t$};
\node () at ($(C00) + (0.7*\gW, -0.25*\gH)$) {$l$};
\node () at ($(D00) + (0.2*\gW, -0.7*\gH)$) {$t$};
\node () at ($(D00) + (0.7*\gW, -0.25*\gH)$) {$l$};

\node () at ($(A02) + (0.5*\gW, 0.5*\gH)$) {\textcolor{black}{$T_\text{opt}[k,l]$}};
\node () at ($(B02) + (0.5*\gW, 0.5*\gH)$) {\textcolor{black}{$I[k,l]$}};
\node () at ($(C02) + (0.5*\gW, 0.5*\gH)$) {\textcolor{black}{$D[l,t]$}};
\node () at ($(D02) + (0.5*\gW, 0.5*\gH)$) {\textcolor{black}{$A[l,t]$}};

\node () at ($(C60) + (-1.4*\gW, -0.5*\gH-0.015)$) {\footnotesize $\textcolor{red}{\underline{21\!-\!T_\text{opt}[2,3]}}$};
\node () at ($(C40) + (-1.4*\gW, -0.5*\gH-0.015)$) {\footnotesize $\textcolor{blue!80!black!100}{\underline{21\!-\!T_\text{opt}[1,3]}}$};
\node () at ($(C20) + (-1.4*\gW, -0.5*\gH-0.015)$) {\footnotesize $\textcolor{green!50!black!100}{\underline{21\!-\!T_\text{opt}[0,3]}}$};
\colorrec{C31}{green!80!black!100}
\colorrec{C52}{blue}
\colorrec{C73}{red}
\colorrec{C94}{black}
\colorrec{D31}{green!80!black!100}
\colorrec{D52}{blue}
\colorrec{D73}{red}
\colorrec{D94}{black}
\colorrec{B53}{red}
\colorrec{B52}{blue}
\colorrec{B51}{green!80!black!100}
\colorrec{A53}{red}
\colorrec{A52}{blue}
\colorrec{A51}{green!80!black!100}

\node[align=left,execute at begin node=\setlength{\baselineskip}{1.6em}] () at (2.4,-7.5) {
$D[3,21] \,= \mathop{\mathrm{max}}\limits_{k'\in\{0,1,2\}}\left(D[k',\underline{21\!-\!T_\text{opt}[k',3]}]+I[k',3]\right)$
\\
$~~~~~~~~~~~\,~~=~~~~\textcolor{blue!80!black!100}{D[1,\underline{7}] + I[1,3]} ~=~1.9$
};

\node (M) at (-0.8, -8.9) {};
\node (X) at (0.5, -8.9) {};
\node (Y) at (1.8, -8.9) {};
\node (Z) at (3.1, -8.9) {};
\node (U) at (7, -8.9) {};
\draw[rounded corners,line width=0.3mm,inner sep=20pt, fill=blue!10] ($(X) + (-0.35,-0.4)$) rectangle  ($(Y) + (0.35,0.4)$);
\draw[rounded corners,line width=0.3mm,inner sep=20pt, fill=blue!10] ($(Z) + (-0.35,-0.4)$) rectangle  ($(U) + (0.35,0.4)$);
\node (M) at (-0.8, -8.9) {\large $X$};
\node (X) at (0.5, -8.9) {\large $f_{\theta_1}$};
\node (Y) at (1.8, -8.9) {\large $\sigma_1$};
\node (Z) at (3.1, -8.9) {\large $f_{\theta_2}$};
\node (W) at (4.4, -8.9) {\large $\mathrm{id}$};
\node (V) at (5.7, -8.9) {\large $f_{\theta_3}$};
\node (U) at (7, -8.9) {\large $\sigma_3$};
\node (T) at (8.3, -8.9) {};
\node[black!100] () at ($($(M)!0.5!(X)$) + (0,0)$) {$\rightarrow$};
\node[black!100] () at ($($(X)!0.5!(Y)$) + (0,0)$) {$\rightarrow$};
\node[black!100] () at ($($(Y)!0.5!(Z)$) + (0,0)$) {$\rightarrow$};
\node[black!100] () at ($($(Z)!0.5!(W)$) + (0,0)$) {$\rightarrow$};
\node[black!100] () at ($($(W)!0.5!(V)$) + (0,0)$) {$\rightarrow$};
\node[black!100] () at ($($(V)!0.5!(U)$) + (0,0)$) {$\rightarrow$};
\node[black!100] () at ($($(U)!0.5!(T)$) + (0,0)$) {$\rightarrow$};
\node[black!100] () at ($(T) + (0,0)$) {$\cdots$};
\draw[rounded corners,line width=0.3mm,inner sep=20pt, blue!80!black!100] ($(X) + (-0.35,-0.4)$) rectangle  ($(Y) + (0.35,0.4)$);
\draw[rounded corners,line width=0.3mm,inner sep=20pt, blue!80!black!100] ($(Z) + (-0.35,-0.4)$) rectangle  ($(U) + (0.35,0.4)$);

\node[black!100] () at ($($(Y)!0.5!(Z)$) + (-1.55 ,-0.8)$) {~~$A[3,21]\,=~~~~\textcolor{blue!80!black!100}{A[1,
\underline{7}] \cup \{1\}}~=~\{1\}$};
\draw[line width=1.6pt, ->] ($(C52)+(0.5*\gW,0)$) to[out=-90, in=130] ($(C94)+(0,\gH)$);
\draw[line width=1.6pt, ->] ($(B52)+(0*\gW,0*\gH)$) to[out=-170, in=80] ($(C94)+(\gW,\gH)$);
\end{tikzpicture}
}

\caption{Illustration of DP table construction based on the recurrence relation (\Cref{eq:dprecur}). $D[3, 21]$ is computed upon the solutions of the previous sub-problems ($D[0,1]$, $D[1,7]$, $D[2, 10]$).}
\label{fig:dp}
\end{figure}
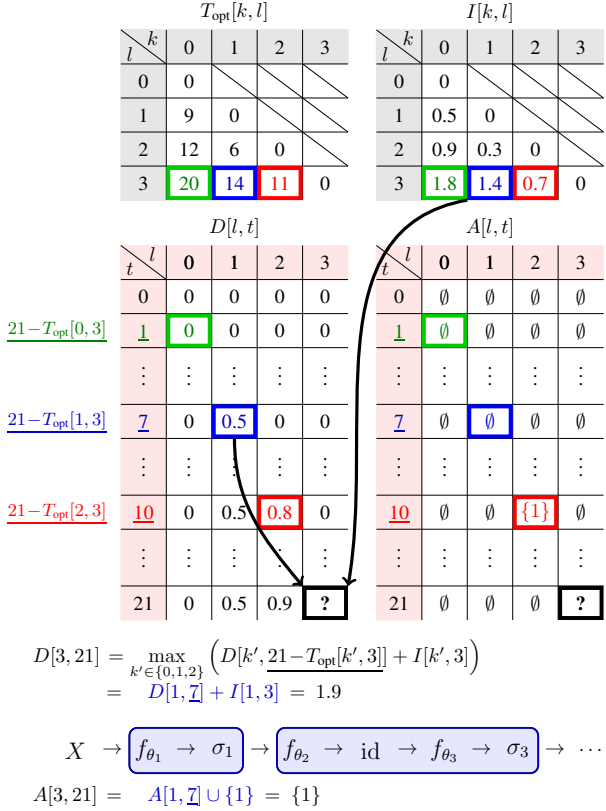
\begin{proposition}
$A[l,t]$ and $S[l,t]$ computed from the DP recurrence relations, \Cref{eq:dprecur} are the optimal sets $A$ and $S$ of \Cref{eq:opt4}, respectively.
\label{prop:dp}
\end{proposition}
\begin{proof}
Refer to \Cref{app:proof}.
\end{proof}
For a given set of the optimal indices where activation layers are not replaced with identity functions, $A[l,t]$ and the network replaced according to $A[l,t]$, \Cref{prop:timedp} shows that $S[l,t]$ is the optimal $S$ which merges the network into the optimal structure in terms of latency.
\begin{proposition}
$S[l,t]$ computed from the DP recurrence relations, \Cref{eq:dprecur} is the optimal $S$ which minimizes the latency of the network when $A[l,t]$ is fixed. Concretely, $S[l,t]$ is the optimal $S$ of the optimization problem:
\begin{align}
    \label{eq:opts}
    \underset{A[l,t]\subseteq S \subseteq [l\!-\!1]}{\mathrm{min}} \sum_{s_{i\!-\!1},s_i \in \{0\}\cup S\cup \{l\}} T[s_{i\!-\!1}, s_{i}] . 
\end{align} 
\label{prop:timedp}
\end{proposition}
\begin{proof}
Refer to \Cref{app:proof}.
\end{proof}


\section{Experimental Results}
We evaluate our method on various datasets and network architectures. 
We first introduce implementation details for the overall process in the experiments. Then, we present an evaluation of our method on various scales of networks and datasets to demonstrate its effectiveness. Furthermore, we conduct ablation studies on the search space of our proposed method and the ordered set $S$.

\subsection{Implementation Details}
\label{subsec:detail}

\paragraph{Measurement} We first evaluate the latency of each contiguous network block, $T[i,j]$, individually. The latency of the network is subject to the format which it is implemented on. We utilize TensorRT to convert the network into its optimal form and measure the latency for a fair comparison.

Then, we measure the change of the accuracy incurred by each contiguous network block, $I[i,j]$, for \Cref{eq:imp}. As the number of possible contiguous network blocks is of the order of $N^2$, where $N$ is the number of activations, we need to train $\mathcal{O}(N^2)$ networks to obtain the accuracy change of every contiguous network block. For efficiency, we approximate the first term in \Cref{eq:imp} using the accuracy of the network trained for a few epochs after replacing the activation layers between the $i\!+\!1$-th and $j$-th layers with identity functions. Specific details on evaluating the importance of each block and the methodologies used to normalize the importance values can be found in the \Cref{app:imp}.

\paragraph{Dynamic Programming} Given the latency of each contiguous network block, $T[i,j]$, and the accuracy change caused by each contiguous network block, $I[i,j]$, we can solve \Cref{eq:opt3} for the time constraint $T_0$ with \Cref{alg:optDP}. In \Cref{alg:optDP}, we assume the time constraint $T_0$ and time index $t$ to be integers. In practice, we multiply every occurrence of $t$ and $T_0$ by a constant factor and round the multiplied values to integer.

\paragraph{Finetune and Merge} After obtaining the optimal ordered sets $A$ and $S$ in \Cref{eq:opt3}, we replace the activation layers not present in $A$ with identity functions. 
In order to exactly merge the network in the inference phase, it is necessary to ensure that sufficient padding is applied to the first convolution layers within the target contiguous network blocks to be merged. To this end, we reorder the zero padding according to the set $S$ first, then finetune the network until convergence. We detail this padding reordering technique in \Cref{sup:padreorder}\footnote{We apply the same padding reordering technique when we reproduce the baseline work, DepthShrinker.}.
At the test time, we merge the finetuned network following $S$ and evaluate the latency.

During finetuning, we follow the identical training protocol with the DepthShrinker for finetuning \citep{depthshrinker}. In detail, we finetune the network for 180 epochs using cosine learning rate decay with the SGD optimizer. We further adopt label smoothing, random erasing and RandAugment following \citet{depthshrinker}, except in the case of MobileNetV2-1.0 on ImageNet where additional augmentation did not improve performance \citep{labelsmoothing,randerase,randaug}.

\paragraph{Evaluation}

We employ \textit{RTX2080 Ti} GPU when evaluating the latency of each contiguous network block. Then, we evaluate the end-to-end inference latency of merged architectures on various GPUs including \textit{TITAN Xp}, \textit{RTX2080 Ti}, \textit{RTX 3090}, and \textit{Tesla V100}. Also, we evaluate the inference latency of the networks in two distinct formats: 1) TensorRT exported model (FP32) and 2) PyTorch model \citep{tensorrt, pytorch}. To ensure a fair comparison, we fuse the batch normalization (BN) modules with the previous convolution layers when we measure latency in the PyTorch format, as the depth compression algorithm results in a different number of BN modules.

\subsection{Depth Compression Results}
We apply our depth compression method to the MobileNetV2 architecture on ImageNet-100 and ImageNet dataset, starting from the public pretrained weight \citep{mobilenet,imagenetsubset,imagenet}.

\subsubsection{ImageNet-100}

\begin{table}[t]
\caption{Accuracy and latency of compressed architectures applied to MobileNetV2-1.0 and MobileNetV2-1.4 on ImageNet-100 dataset. Compression methods use the latency information of \textit{RTX 2080 Ti} and is measured on \textit{RTX 2080 Ti} with batch size of 128. We report the average accuracy of the three runs of finetuning.} 
\vspace{0.5em}
\centering
\begin{adjustbox}{max width=1.0\columnwidth}
\begin{tabular}{lccc}
\toprule
                            && \normalsize{TensorRT}
                            & \footnotesize{\textit{w/o}} \normalsize{TensorRT }\\
                        Network     & Acc (\%)& Lat. (ms)
                            & Lat. (ms) \\
 \cmidrule(r){1-2}\cmidrule(r){3-3} \cmidrule(r){4-4}
       MBV2-1.0  & 87.58 & 19.25 & 40.61 \\
 \cmidrule(r){1-2}\cmidrule(r){3-3} \cmidrule(r){4-4}
    DS-A-1.0  & 87.58 & 14.74 &  27.59  \\
    Ours& \textbf{87.69}  & \textbf{12.53} & \textbf{23.02} \\
 \cmidrule(r){1-2}\cmidrule(r){3-3} \cmidrule(r){4-4}
    DS-B-1.0  & 87.31 & 12.33 &  22.99  \\
    Ours& \textbf{87.45}  & \textbf{12.11} & \textbf{22.29} \\
 \cmidrule(r){1-2}\cmidrule(r){3-3} \cmidrule(r){4-4}
    DS-C-1.0 & 85.92 & 11.20 &  20.76 \\
    Ours      & \textbf{86.73} & \textbf{11.14} & \textbf{20.62} \\
 \cmidrule(r){1-2}\cmidrule(r){3-3} \cmidrule(r){4-4}
    DS-D-1.0 & 85.30 & 10.49 &  18.78 \\
    Ours      & \textbf{85.91} & \textbf{9.62} & \textbf{16.82} \\
\cmidrule[0.5pt]{1-4} \\ [-3.5ex]
\cmidrule[0.5pt]{1-4}
     MBV2-1.4  & 88.88 & 29.94 & 61.68 \\
 \cmidrule(r){1-2}\cmidrule(r){3-3} \cmidrule(r){4-4}
    DS-A-1.4  & 88.01 & 19.61 & 35.06 \\
    Ours& \textbf{88.41}  & \textbf{19.48} & \textbf{34.01} \\
 \cmidrule(r){1-2}\cmidrule(r){3-3} \cmidrule(r){4-4}
    DS-B-1.4  & 86.99 & 19.21 & 31.63 \\
    Ours& \textbf{87.58}  & \textbf{18.22} & \textbf{30.77} \\
 \cmidrule(r){1-2}\cmidrule(r){3-3} \cmidrule(r){4-4}
    DS-C-1.4 & 86.73 & 17.47 & 29.73 \\
    DS-D-1.4 & 86.05 & 17.50 & 27.99 \\
    Ours      & \textbf{87.18} & \textbf{16.26} & \textbf{27.42} \\
 \cmidrule(r){1-2}\cmidrule(r){3-3} \cmidrule(r){4-4}
    DS-E-1.4 & 85.29 & 15.67 & 26.08 \\
    Ours      & \textbf{85.93} & \textbf{14.65} & \textbf{22.96} \\
\bottomrule
\end{tabular}
\end{adjustbox}
\label{tab:small-in100-mbv2-1.4}
\end{table}

\begin{table}[t]
\caption{Accuracy and latency of compressed architectures applied to MobileNetV2-1.0 on ImageNet dataset. 
Compression methods use the latency information of  \textit{RTX 2080 Ti} and is measured on \textit{RTX 2080 Ti} with batch size of 128. $\dagger$ denotes the accuracy of the pretrained weight used in DepthShrinker, and we use the same pretrained weight for a fair comparison.}
\label{tab:small-in-mbv2-1.0}
\vspace{0.5em}
\centering

\begin{adjustbox}{max width=1.0\columnwidth}
\begin{tabular}{lccc}
\toprule
                            && \normalsize{TensorRT}
                            & \footnotesize{\textit{w/o}} \normalsize{TensorRT
}\\
                        Network     & Acc (\%)& Lat. (ms)
                            & Lat. (ms)  \\
 \cmidrule(r){1-2}\cmidrule(r){3-3} \cmidrule(r){4-4}
    MBV2-1.0  & 72.89$^\dagger$ & 19.26 & 40.71 \\
 \cmidrule(r){1-2}\cmidrule(r){3-3} \cmidrule(r){4-4}
    DS-A-1.0  & 72.37 & 14.82 & 27.53  \\
    Ours& \textbf{72.83}  & \textbf{13.67} & \textbf{25.09}  \\
 \cmidrule(r){1-2}\cmidrule(r){3-3} \cmidrule(r){4-4}
    DS-B-1.0  & 71.96 & 12.42 & 22.92 \\
    Ours& \textbf{72.13}  & \textbf{12.38} & \textbf{21.74} \\
 \cmidrule(r){1-2}\cmidrule(r){3-3} \cmidrule(r){4-4}
    DS-C-1.0  & 70.87 & 11.28 & 20.77  \\
    Ours      & \textbf{71.44} & \textbf{10.90} & \textbf{19.75}  \\
 \cmidrule(r){1-2}\cmidrule(r){3-3} \cmidrule(r){4-4}
    DS-D-1.0  & 69.43 & 10.53 & 18.82 \\
    Ours      & \textbf{70.65} & \textbf{9.88} & \textbf{16.55}  \\

\bottomrule
\end{tabular}
\end{adjustbox}
\end{table}

\begin{table*}[t]
\caption{Accuracy and latency of compressed architectures applied to MobileNetV2-1.4 on ImageNet dataset. Compression methods use the latency information of \textit{RTX 2080 Ti}. The latency of the compressed network architecture is measured on \textit{TITAN Xp}, \textit{RTX 2080 Ti}, \textit{RTX 3090}, and  \textit{Tesla V100} with batch size of 128. $\dagger$ denotes the accuracy of the pretrained weight used in DepthShrinker, and we use the same pretrained weight for a fair comparison.}
\vspace{0.5em}
\label{tab:in-mbv2-1.4}
\centering
\begin{adjustbox}{max width=2.0\columnwidth}
\begin{tabular}{lcccccc}
\toprule
    & 
                            & \multicolumn{4}{c}{TensorRT Latency (ms)}
                            & \footnotesize{\textit{w/o} }\normalsize{TensorRT (ms)}\\
                            \cmidrule(lr){3-6}
                            \cmidrule(lr){7-7}
                            Network & Acc (\%)& \emph{TITAN Xp} & \emph{RTX 2080 Ti} & \emph{RTX 3090} 
                            & \emph{Tesla V100} & \emph{RTX 2080 Ti} \\
 \cmidrule(r){1-2}\cmidrule(r){3-6} \cmidrule(r){7-7}
    MobileNetV2-1.4  & 76.28$^\dagger$ & 42.13 &  29.93 & 20.79 & 24.35 & 61.64 \\
 \cmidrule(r){1-2}\cmidrule(r){3-6} \cmidrule(r){7-7}
    MBV2-1.4-DS-A  &  74.42  & 26.87 & 19.62 & 13.54 & 16.13 & 35.05 \\
    Ours & \textbf{74.68} & \textbf{25.77} & \textbf{18.63} & \textbf{12.88} & \textbf{15.89} & \textbf{32.35} \\
 \cmidrule(r){1-2}\cmidrule(r){3-6} \cmidrule(r){7-7}
    MBV2-1.4-DS-B  &  74.06  & 25.30 & 19.20 & 13.21 & 15.82 & 31.63 \\
    Ours& \textbf{74.19} & \textbf{24.69} & \textbf{18.10} & \textbf{12.32} & \textbf{15.15} & \textbf{31.34} \\
 \cmidrule(r){1-2}\cmidrule(r){3-6} \cmidrule(r){7-7}
    MBV2-1.4-DS-C & 73.30 & 23.97 & 17.48 & 12.07 &  14.48 & 29.69 \\
    MBV2-1.4-DS-D & 72.99 & 22.81 & 17.51 & 12.01 & 14.30 & 27.93 \\
    Ours& \textbf{73.46} & \textbf{22.40} & \textbf{16.39} & \textbf{11.15} & \textbf{13.64} & \textbf{27.54} \\
 \cmidrule(r){1-2}\cmidrule(r){3-6} \cmidrule(r){7-7}
    MBV2-1.4-DS-E & 72.34 & 21.04 & 15.71 & 10.81 & 12.97 & 26.01 \\
    Ours      & \textbf{72.57} & \textbf{20.49} & \textbf{15.03} & \textbf{10.29} & \textbf{12.86} & \textbf{25.84} \\
\bottomrule
\end{tabular}
\end{adjustbox}
\end{table*}

We first experiment with our depth compression method on the ImageNet-100 dataset, which is a subset of ImageNet consisting of 100 classes. We bring the list of the subclasses from \citet{imagenetsubset}. The size of the image is preprocessed to $224 \times 224$ and the dataset contains approximately 1200 images per class. We apply our depth compression method to both MobileNetV2-1.0 and MobileNetV2-1.4 starting from the pretrained weight and compare it to the architectures proposed in DepthShrinker.

When implementing the DepthShrinker on the ImageNet-100 dataset, we bring the architectures in DepthShrinker and finetune from the pretrained weight after substituting the last classifier to match the number of classes \citep{depthshrinker}. Then we measure the latency of the merged network.

\Cref{tab:small-in100-mbv2-1.4} summarizes the depth compression results in MobileNetV2-1.0 and MobileNetV2-1.4. Our method consistently outperforms the baseline at every compression ratio in MobileNetV2-1.0 and MobileNetV2-1.4. In particular, we achieve $1.08 \times$ speedup in TensorRT compiled format with $1.13$\%p higher accuracy compared to DS-D-1.4. Also, we achieve $1.18 \times$ speedup with $0.11$\%p higher accuracy in TensorRT compiled format compared to DS-A-1.0.


Additionally, we evaluate the wall-clock inference time on various GPU platforms other than \textit{RTX 2080 Ti}. The comprehensive result of the latency on different GPUs can be found in \Cref{app:gpus}. Furthermore, we reproduce the full searching stage of DepthShrinker on top of the ImageNet-100 dataset and compare our method against the resulting architecture which we also provide the results in \Cref{app:in100-ds}.

\subsubsection{ImageNet}

We apply our depth compression method to MobileNetV2-1.0 and MobileNetV2-1.4 on the full ImageNet dataset \citep{imagenet} and compare with the architectures proposed in DepthShrinker \citep{depthshrinker}\footnote{DepthShrinker's official implementation omits merging the first Inverted Residual Block; following their paper, we merge it if their pattern removes the activation in this block.}. Note that every method uses the latency information of the \textit{RTX 2080 Ti} with TensorRT and is measured on different model formats and GPU platforms.

We use the same pretrained weight with DepthShrinker for a fair comparison and report the accuracy of it for the vanilla network. It is worth noting that this accuracy value differs from the one reported by \citet{depthshrinker} because they reported the accuracy of the vanilla network from their baseline work instead of the pretrained weight they started from. We report the accuracy of the pretrained weight to precisely convey the effect of the compression methods.

\Cref{tab:small-in-mbv2-1.0} demonstrates that our method consistently outperforms the baseline in MobileNetV2-1.0 architecture on the ImageNet dataset. Specifically, our method attains $1.08 \times$ speedup with $0.46$\%p higher accuracy compared to DS-A-1.0. We present the comprehensive table including the latency on different GPUs in \Cref{app:gpus}.

\Cref{tab:in-mbv2-1.4} shows the result of applying our method to MobileNetV2-1.4. The result demonstrates that our method outperforms the baseline method in every compression ratio and across all model formats and GPU platforms. In particular, our method achieves $1.07 \times$ speedup in TensorRT compiled format with higher accuracy compared to MBV2-1.4-DS-C. Compared to the pretrained network, our compressed network achieves $1.61 \times$ speedup in TensorRT compiled format and $1.91 \times$ speedup without TensorRT with $1.60$\%p accuracy drop.

We further present the results of applying knowledge distillation from the pretrained weight when we finetune the compressed networks. \Cref{tab:dist-mbv2} shows that adopting the knowledge distillation technique further boosts the accuracy of the compressed networks. Specifically for MobileNetV2-1.0, our method achieves 1.41$\times$ speedup in TensorRT format and 1.62$\times$ speedup in PyTorch without losing accuracy from the pretrained weight.

\begin{table}[t]
\caption{Accuracy and latency of compressed architectures applied to MobileNetV2 on ImageNet dataset adopting knowledge distillation technique. 
Compression methods use the latency information of  \textit{RTX 2080 Ti} and the latency is measured on \textit{RTX 2080 Ti} with batch size of 128. $\dagger$ denotes the accuracy of the pretrained weight used in DepthShrinker, and we use the same pretrained weight for a fair comparison.}
\label{tab:dist-mbv2}
\vspace{0.5em}
\centering

\begin{adjustbox}{max width=1.0\columnwidth}
\begin{tabular}{lccc}
\toprule
                            && \normalsize{TensorRT}
                            & \footnotesize{\textit{w/o}} \normalsize{TensorRT
}\\
                        Network     & Acc (\%)& Lat. (ms)
                            & Lat. (ms)  \\
 \cmidrule(r){1-2}\cmidrule(r){3-3} \cmidrule(r){4-4}
    MBV2-1.0  & 72.89$^\dagger$ & 19.26 & 40.71 \\
 \cmidrule(r){1-2}\cmidrule(r){3-3} \cmidrule(r){4-4}
    DS-A-1.0  & 72.76 & 14.82 & 27.53  \\
    Ours& \textbf{73.00}  & \textbf{13.67} & \textbf{25.09}  \\
\midrule \\ [-3.5ex]
\midrule
    MBV2-1.4  & 76.28$^\dagger$ & 29.93 & 61.64 \\
 \cmidrule(r){1-2}\cmidrule(r){3-3} \cmidrule(r){4-4}
    DS-A-1.4  & 75.08 & 19.62 & 35.05 \\
    Ours& \textbf{75.16}  & \textbf{18.80} & \textbf{32.78} \\
\bottomrule
\end{tabular}
\end{adjustbox}
\end{table}
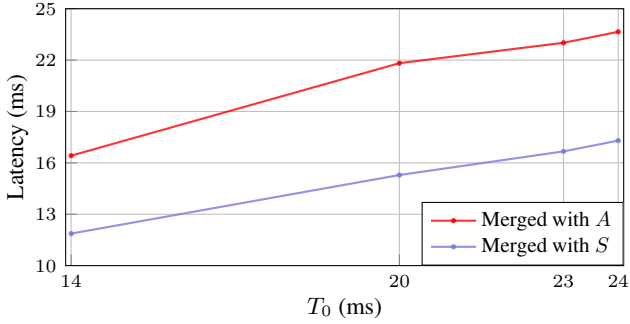
\begin{figure}[t]
\begin{tikzpicture}[define rgb/.code={\definecolor{mycolor}{RGB}{#1}},
                    rgb color/.style={define rgb={#1},mycolor}]

\definecolor{gr}{RGB}{60,160,100}
\definecolor{or}{RGB}{200,140,80}
\definecolor{bl}{RGB}{120,120,220}
\definecolor{yl}{RGB}{200,200,100}
\definecolor{pp}{RGB}{200,150,240}

\begin{groupplot}[
        group style={columns=1, horizontal sep=1.05cm, 
        vertical sep=0.0cm},
        ]

\nextgroupplot[
            width=9cm,
            height=5cm,
            every axis plot/.append style={thick},
            grid=major,
            scaled ticks = false,
            xlabel near ticks,
            ylabel near ticks,
            tick pos=left,
            tick label style={font=\scriptsize},
            xlabel shift=-0.1cm,         
            ylabel shift=-0.15cm,
            label style={font=\small},
            xlabel style={align=center},
            xlabel={$T_0$ (ms)},
            ylabel={Latency (ms)},
            xmin=13.9,
            xmax=24.1,
            xtick={14.0,20.0, 23.0, 24.0, 27.0},
            ytick={10,13,16,19,22,25},
            ymin=10,
            ymax=25,
            legend cell align=left,
            legend style={at={(1,0)},anchor=south east, nodes={scale=0.8}},
            ]

\addplot[red, opacity=0.8, mark=*, mark size=0.6pt] table [y=act, col sep=comma]{data/abl_merge.csv};\addlegendentry{Merged with $A$}

\addplot[bl, opacity=0.8, mark=*, mark size=0.6pt] table [y=opt, col sep=comma]{data/abl_merge.csv};\addlegendentry{Merged with $S$}
\end{groupplot}
\end{tikzpicture}
    \caption{Latency comparison between the network that is merged according to $A$ and the network that is merged according to $S$ for different time constraint $T_0$. $A$ and $S$ are the optimal solutions of \Cref{eq:opt3} where $I[\cdot, \cdot]$ and $T[\cdot,\cdot]$ are evaluated for MobileNetV2-1.0 on ImageNet dataset.
    }
    \label{fig:ablmerge}
    \vspace{-0.5em}
\end{figure}

\begin{figure}[t]
\resizebox{\columnwidth}{!}{
\begin{tikzpicture}[font=\large]
\def\gW{0.7}
\def\gH{0.6}

\coordinate (A) at (0,0);
\coordinate (ACT0) at ($(A)+(0,1.5)$);
\coordinate (B) at ($(A)+(0,-1.8)$);
\coordinate (ACT1) at ($(B)+(0,0.9)$);
\coordinate (C) at ($(B)+(0,-2.0)$);
\coordinate (ACT2) at ($(C)+(0,1.0)$);
\coordinate (D) at ($(C)+(0,-1.8)$);
\coordinate (ACT3) at ($(D)+(0,0.9)$);
\coordinate (E) at ($(D)+(0,-1.8)$);
\coordinate (ACT4) at ($(E)+(0,0.9)$);
\coordinate (ACT5) at ($(E)+(0,-1.3)$);
\draw[rounded corners, dash pattern=on 4pt off 4pt,line width=0.3mm,inner sep=20pt, fill=blue!5] 
($(B) + (-2.5,-0.6)$)  rectangle ($(A) + (2.5,0.6)$) ;
\draw[rounded corners, dash pattern=on 4pt off 4pt,line width=0.3mm,inner sep=20pt, fill=blue!5] 
($(E) + (-2.5,-0.6)$)  rectangle ($(C) + (2.5,0.6)$) ;
\node[blue!60!black!100,align=left,execute at begin node=\setlength{\baselineskip}{0.8em}] () at ($(ACT2) + (-2.35,-0.05)$) {\normalsize IR \\\normalsize Block2};
\node[blue!60!black!100,align=left,execute at begin node=\setlength{\baselineskip}{0.8em}] () at ($(ACT0) + (-2.35,-0.5)$) {\normalsize IR \\\normalsize Block1};

\node () at (ACT0) {$48 \times H \times W$};
\node () at (ACT1) {$48 \times H \times W$};
\node () at (ACT2) {$24 \times H \times W$};
\node () at (ACT3) {$144 \times H \times W$};
\node () at (ACT4) {$144 \times H \times W$};
\node () at (ACT5) {$32 \times H \times W$};

\draw[line width=0.8mm,inner sep=2pt, black] ($(A) + (-1.7,-0.4)$) rectangle  ($(A) + (1.7,0.4)$);
\draw[line width=0.8mm,inner sep=2pt, black] ($(B) + (-1.2,-0.4)$) 
--($(B) + (-1.7,0.4)$)   
--($(B) + (1.7,0.4)$)
--($(B) + (1.2,-0.4)$)
-- cycle;
\draw[line width=0.8mm,inner sep=2pt, black] ($(C) + (-2.0,-0.4)$) 
--($(C) + (-1.2,0.4)$)   
--($(C) + (1.2,0.4)$)
--($(C) + (2.0,-0.4)$)
-- cycle;
\draw[line width=0.8mm,inner sep=2pt, black] ($(D) + (-2.0,-0.4)$) rectangle  ($(D) + (2.0,0.4)$);
\draw[line width=0.8mm,inner sep=2pt, black] ($(E) + (-1.4,-0.4)$) 
--($(E) + (-2.0,0.4)$)   
--($(E) + (2.0,0.4)$)
--($(E) + (1.4,-0.4)$)
-- cycle;

\node () at (A) {Dw, $3\!\times\!3$, $s\!=\!1$};
\node () at (B) {$1\!\times\!1$, $s\!=\!1$};
\node () at (C) {$1\!\times\!1$, $s\!=\!1$};
\node () at (D) {Dw, $3\!\times\!3$, $s\!=\!2$};
\node () at (E) {$1\!\times\!1$, $s\!=\!1$};
\node () at ($(ACT0) + (0,2.0)$) {MobileNetV2-1.4};
\draw[line width=1mm,->] ($(ACT0) + (0,1)$) -- ($(ACT0) + (0,0.5)$);
\node () at ($(ACT0) + (0,1.5)$) {\LARGE $\vdots$};
\draw[line width=1mm,->] ($(ACT5) + (0,-0.5)$) -- ($(ACT5) + (0,-1)$);
\node () at ($(ACT5) + (0,-1.3)$) {\LARGE $\vdots$};
\node () at ($(ACT5) + (0,-2)$) {$\textcolor{red}{<}$};
\node () at ($(ACT5) + (0,-2.6)$){$\textcolor{blue}{>}$};

\coordinate (A) at (-4.8,0);
\coordinate (ACT0) at ($(A)+(0,1.5)$);
\coordinate (B) at ($(A)+(0,-1.8)$);
\coordinate (ACT1) at ($(B)+(0,0.9)$);
\coordinate (C) at ($(B)+(0,-2.0)$);
\coordinate (ACT2) at ($(C)+(0,1.0)$);
\coordinate (D) at ($(C)+(0,-1.8)$);
\coordinate (ACT3) at ($(D)+(0,0.9)$);
\coordinate (E) at ($(D)+(0,-1.8)$);
\coordinate (ACT4) at ($(E)+(0,0.9)$);
\coordinate (ACT5) at ($(E)+(0,-1.3)$);

\coordinate (X) at ($(ACT1)+(0,0)$);
\coordinate (Y) at ($(D)+(0,0)$);
\draw[line width=0.8mm,inner sep=2pt, black] ($(X) + (-1.2,-0.8)$) 
--($(X) + (-1.7,0.8)$)   
--($(X) + (1.7,0.8)$)
--($(X) + (1.2,-0.8)$)
-- cycle;

\draw[line width=0.8mm,inner sep=2pt, black] ($(Y) + (-1.4,-0.8)$) 
--($(Y) + (-1.2,0.8)$)   
--($(Y) + (1.2,0.8)$)
--($(Y) + (1.4,-0.8)$)
-- cycle;

\node () at (ACT0) {$48 \times H \times W$};
\node () at (ACT2) {$24 \times H \times W$};
\node () at (ACT5) {$32 \times H \times W$};

\node () at (X) {$3\!\times\!3$, $s\!=\!1$};
\node () at (Y) {$3\!\times\!3$, $s\!=\!2$};
\draw[dotted,line width=0.7mm] ($(X) + (1.7,0.8)$)--($(A) + (4.8-1.7,0.4)$);
\draw[dotted,line width=0.7mm] ($(X) + (1.2,-0.8)$)--($(B) + (4.8-1.2,-0.4)$);

\draw[dotted,line width=0.7mm] ($(Y) + (1.2,0.8)$)--($(C) + (4.8-1.2,0.4)$);
\draw[dotted,line width=0.7mm] ($(Y) + (1.4,-0.8)$)--($(E) + (4.8-1.4,-0.4)$);

\node () at ($(ACT0) + (0,2.0)$) {DepthShrinker};
\draw[line width=1mm,->] ($(ACT0) + (0,1)$) -- ($(ACT0) + (0,0.5)$);
\node () at ($(ACT0) + (0,1.5)$) {\LARGE $\vdots$};
\draw[line width=1mm,->] ($(ACT5) + (0,-0.5)$) -- ($(ACT5) + (0,-1)$);
\node () at ($(ACT5) + (0,-1.3)$) {\LARGE $\vdots$};
\node () at ($(X)+(0,1.15)$) {$\textcolor{red}{I=2.67},\textcolor{blue}{T=2.79}$};
\node () at ($(Y)+(0,1.15)$) {$\textcolor{red}{I\!=\!-0.25},\textcolor{blue}{T\!=\!0.86}$};
\node () at ($(ACT5) + (0,-2)$) {$\textcolor{red}{I=2.42}$};
\node () at ($(ACT5) + (0,-2.6)$){$\textcolor{blue}{T=3.65}$};

\coordinate (A) at (4.8,0);
\coordinate (ACT0) at ($(A)+(0,1.5)$);
\coordinate (B) at ($(A)+(0,-1.8)$);
\coordinate (ACT1) at ($(B)+(0,0.9)$);
\coordinate (C) at ($(B)+(0,-2.0)$);
\coordinate (ACT2) at ($(C)+(0,1.0)$);
\coordinate (D) at ($(C)+(0,-1.8)$);
\coordinate (ACT3) at ($(D)+(0,0.9)$);
\coordinate (E) at ($(D)+(0,-1.8)$);
\coordinate (ACT4) at ($(E)+(0,0.9)$);
\coordinate (ACT5) at ($(E)+(0,-1.3)$);

\coordinate (X) at ($(ACT1)+(0,0)$);
\coordinate (Y) at ($($(ACT1)!0.5!(ACT5)$)+(0,-0.3)$);
\draw[line width=0.8mm,inner sep=2pt, black] ($(Y) + (-1.4,-0.8)$) 
--($(Y) + (-1.8,0.8)$)   
--($(Y) + (1.8,0.8)$)
--($(Y) + (1.4,-0.8)$)
-- cycle;

\node () at (ACT0) {$48 \times H \times W$};
\node () at (ACT1) {$48 \times H \times W$};
\node () at (ACT5) {$32 \times H \times W$};

\draw[line width=0.8mm,inner sep=2pt, black] ($(A) + (-1.7,-0.4)$) rectangle  ($(A) + (1.7,0.4)$);
\node () at (A) {Dw, $3\!\times\!3$, $s\!=\!1$};
\node () at (Y) {$3\!\times\!3$, $s\!=\!2$};

\draw[dotted,line width=0.7mm] ($(A) + (-1.7,0.4)$)--($(A) + (-4.8+1.7,0.4)$);
\draw[dotted,line width=0.7mm] ($(A) + (-1.7,-0.4)$)--($(A) + (-4.8+1.7,-0.4)$);
\draw[dotted,line width=0.7mm] ($(Y) + (-1.4,-0.8)$)--($(E) + (-4.8+1.4,-0.4)$);
\draw[dotted,line width=0.7mm] ($(Y) + (-1.8,0.8)$)--($(B) + (-4.8+1.7,+0.4)$);
\node () at ($(ACT0) + (0,2.0)$) {Ours};
\draw[line width=1mm,->] ($(ACT0) + (0,1)$) -- ($(ACT0) + (0,0.5)$);
\node () at ($(ACT0) + (0,1.5)$) {\LARGE $\vdots$};
\draw[line width=1mm,->] ($(ACT5) + (0,-0.5)$) -- ($(ACT5) + (0,-1)$);
\node () at ($(ACT5) + (0,-1.3)$) {\LARGE $\vdots$};
\node () at ($(A)+(0,0.75)$) {$\textcolor{red}{I=2.63},\textcolor{blue}{T=1.63}$};
\node () at ($(Y)+(0,1.15)$) {$\textcolor{red}{I=3.12},\textcolor{blue}{T=1.49}$};
\node () at ($(ACT5) + (0,-2)$) {$\textcolor{red}{I=5.75}$};
\node () at ($(ACT5) + (0,-2.6)$){$\textcolor{blue}{T=3.12}$};
\end{tikzpicture}
}
    \caption{Example of our method finding the network structure that DepthShrinker is unable to find. Our method has a larger search space since it can merge across the blocks while DepthShrinker only considers merging within the Inverted Residual Block. $I[\cdot, \cdot]$ and $T[\cdot,\cdot]$ are evaluated for MobileNetV2-1.4 on ImageNet.
    }
    \label{fig:qual}
\end{figure}
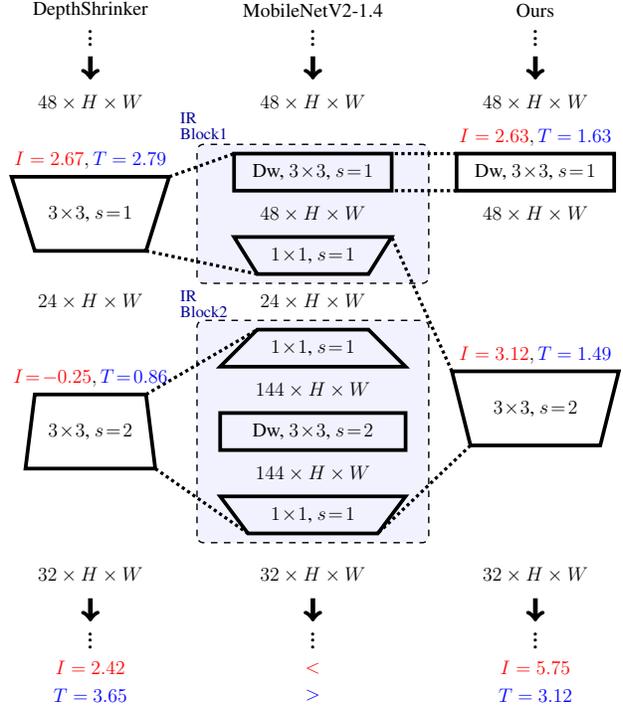

\subsection{Ablation Study on Ordered Set to be Merged}

Recall the definition of $A$ and $S$: $A$ indicates locations where the activation layer is not replaced with an identity function and $S$ indicates indices where we do not merge. The set $S$ always includes $A$ since the activation layers that are not $\mathrm{id}$ cannot be merged. One could argue that we can merge the layers with respect to $A$, without separately computing the optimal merge pattern $S$. In this ablation study, we compare the inference time of the merged network according to $A$ and $S$. \Cref{fig:ablmerge} shows that the network merged according to $S$ is about $30\%$ faster than the network merged according to $A$. This demonstrates that jointly optimizing over $A$ and $S$ simultaneously is crucial for optimal depth compression.

\subsection{An Illustration of the Larger Search Space}

The scope of the DepthShrinker is restricted to the cases where merging operation occurs within the Inverted Residual Block \citep{depthshrinker}. On the other hand, our merging algorithm can handle any series of convolution operations and is agnostic to any specific block structure. For instance, our method finds the architecture that merges across the blocks, which DepthShrinker cannot find as shown in \Cref{fig:qual}. Our method allows us to merge more general series of layers enabling us to discover a more diverse kind of efficient structure.


\section{Conclusion}
We propose an efficient depth compression algorithm to reduce the depth of neural networks for the reduction in run-time memory usage and fast inference latency. Our compression target includes any general convolution operations, whereas existing methods are limited to consecutive depth-wise convolution and point-wise convolution within Inverted Residual Block. We propose a subset selection problem which replaces inefficient activation layers with identity functions and optimally merges consecutive convolution operations into shallow equivalent convolution operations for fast end-to-end inference latency. Since the optimal depth subset selection problem is NP-hard, we formulate a surrogate optimization problem which can be exactly solved via two-stage dynamic programming within a few seconds. We evaluate our methods and baselines by TensorRT for a fair inference latency comparison. Our method outperforms Depthshrinker with a higher accuracy and faster inference speed in MobileNetV2 on the ImageNet dataset. Specifically, we achieve $1.41\times$ speed-up with $0.11$\%p accuracy gain in MobileNetV2-1.0 on the ImageNet.

\section*{Acknowledgements}
This work was supported in part by Samsung Advanced Institute of Technology, Samsung Electronics Co., Ltd. (IO220810-01900-01), Institute of Information \& Communications Technology Planning \& Evaluation (IITP) grant funded by the Korea government (MSIT) ((SW STAR LAB) Development of deployable learning intelligence via self-sustainable and trustworthy machine learning). This material is based upon work supported by the Air Force Office of Scientific Research under award number FA2386-22-1-4010. Yeonwoo Jeong was supported by National Research Foundation of Korea Grant funded by the Korean Government (NRF-2019-Global Ph.D. Fellowship Program). Hyun Oh Song is the corresponding author.


\bibliography{main}
\bibliographystyle{icml2023}

\newpage
\appendix
\onecolumn

\section{Proof}
\label{app:proof}
\begin{proposition}
$A[l,t]$,and $S[l,t]$ computed from the DP recurrence relations, \Cref{eq:dprecur} are the optimal sets $A$ and $S$ of \Cref{eq:opt4}, respectively.
\end{proposition}
\begin{proof}
For given $(l_0, t_0)$, we suppose for all $l<l_0$ and $t<t_0$, $(A[l, t], S[l, t])$ computed from the DP recurrence, \Cref{eq:dprecur} are the optimal $(A,S)$ of \Cref{eq:opt4}, respectively. When $(l,t)=(l_0,t_0)$,  
\begin{align*}
\sum_{s_{i\!-\!1},s_i \in \{0\} \cup S[l_0, t_0] \cup\{l_0\}} T[s_{i\!-\!1}, s_i]
&= \sum_{s_{i\!-\!1},s_i \in \{0\} \cup S[k_0, t_0\!- \!T_\text{opt}[k_0,l_0]] \cup \{k_0\}\cup S_\text{opt}[k_0,l_0]\cup\{l_0\}} T[s_{i\!-\!1}, s_i] \tag{by \Cref{eq:dprecurs}} \\
&= \sum_{s_{i\!-\!1},s_i \in \{0\} \cup S[k_0, t_0\!- \!T_\text{opt}[k_0,l_0]] \cup \{k_0\}} T[s_{i\!-\!1}, s_i] +
\sum_{s_{i\!-\!1},s_i \in \{k_0\}\cup S_\text{opt}[k_0,l_0]\cup\{l_0\}} T[s_{i\!-\!1}, s_i]
\\
&= \sum_{s_{i\!-\!1},s_i \in \{0\} \cup S[k_0, t_0\!- \!T_\text{opt}[k_0,l_0]] \cup \{k_0\}} T[s_{i\!-\!1}, s_i] + T_\text{opt} [k_0,l_0] \tag{by \Cref{eq:timedp2}}\\
&<\left(t_0 - T_\text{opt}[k_0,l_0]\right) + T_\text{opt} [k_0,l_0] = t_0,\tag{by the optimality assumption for $k_0<l_0$ and $t_0-T_{\text{opt}}[k_0,,l_0]<t_0$}
\end{align*}
where
\begin{align*}
 k_0&= \underset{0\leq k'<l}{\mathrm{argmax}} ~ 
    \sum_{a_{j\!-\!1},a_j \in \{0\}\cup A[k', t_0\!-\!T_\text{opt}[k',l_0]]\cup \{k',l\}} I[a_{j\!-\!1}, a_{j}]
\nonumber\\
    &~~\text{subject to } ~~T_\text{opt}[0,k'] +T_\text{opt}[k',l]< t_0.\nonumber\\
\end{align*}

Assume that $(A[l_0,t_0], S[l_0,t_0])$ obtained using \Cref{eq:dprecur} are not optimal $(A, S)$ and  $(A^*, S^*)$ are the optimal $(A, S)$ of  \Cref{eq:opt4} when $(l,t)=(l_0,t_0)$. Then,

\begin{subequations}
\begin{align}
      \sum_{a_{j\!-\!1},a_j \in \{0\}\cup A^*\cup \{l_0\}} I[a_{j\!-\!1}, a_{j}]&> \sum_{a_{j\!-\!1},a_j \in \{0\}\cup A[l_0,t_0]\cup \{l_0\}} I[a_{j\!-\!1}, a_{j}]\label{eq:proofoptimp}\\
   \sum_{s_{i\!-\!1},s_i \in \{0\} \cup S^* \cup\{l_0\}} T[s_{i\!-\!1}, s_i]&<t_0, \label{eq:proofopttime}
\end{align}
\end{subequations}
where $A^* \subseteq S^* \subseteq [l_0-1]$.  

$A^*$ is not an empty set due to \Cref{eq:proofoptimp} and 
\begin{align*} 
\sum_{a_{j\!-\!1},a_j \in \{0\}\cup A[l_0,t_0]\cup \{l_0\}} I[a_{j\!-\!1}, a_{j}]
&= \sum_{a_{j\!-\!1},a_j \in \{0\}\cup A[k_0, t_0\!-\!T_\text{opt}[k_0,l_0]]\cup \{k_0,l_0\}} I[a_{j\!-\!1}, a_{j}]\tag{by \Cref{eq:dprecura}}\\
&\geq \sum_{a_{j\!-\!1},a_j \in \{0\}\cup A[0, t_0\!-\!T_\text{opt}[0,l_0]]\cup \{0,l_0\}} I[a_{j\!-\!1}, a_{j}]\tag{by the definition of $k_0$}\\
&=\sum_{a_{j\!-\!1},a_j \in \{0\}\cup \emptyset\cup \{l_0\}} I[a_{j\!-\!1}, a_{j}]].\tag{by the base case condition}
\end{align*}
Then, let $k^*$ be the maximum value of set $A^*$. 

We define $A' = A^*\setminus\{k^*\}$, $S'_{<k^*}= \{s\in S^*\mid s< k^*\}$, and $S'_{>k^*}= \{s\in S^*\mid s> k^*\}$. The upper bound of $T(S'_{<k^*}, 0, k^*)$ is given as follows:
\begin{align*}
\sum_{s_{i\!-\!1},s_i \in \{0\} \cup S'_{<k^*} \cup\{k^*\}} T[s_{i\!-\!1}, s_i]
&= \sum_{s_{i\!-\!1},s_i \in \{0\} \cup S^* \cup\{l_0\}} T[s_{i\!-\!1}, s_i]
- \sum_{s_{i\!-\!1},s_i \in \{k^*\} \cup S'_{>k^*} \cup\{l_0\}} T[s_{i\!-\!1}, s_i]
\nonumber\\
&\leq \sum_{s_{i\!-\!1},s_i \in \{0\} \cup S^* \cup\{l_0\}} T[s_{i\!-\!1}, s_i] -  T_\text{opt}[k^*, l_0]\tag{by \Cref{eq:timedp1}}\nonumber\\
&< t_0 - T_\text{opt}[k^*, l_0].\tag{by \Cref{eq:proofopttime}}\nonumber
\end{align*}
Therefore, the optimality assumption of $A[k^*, t_0- T_\text{opt}[k^*, l_0]]$ in \Cref{eq:opt4} leads to the inequality:
\begin{align}
\label{eq:proofoptimp3}
\sum_{a_{j\!-\!1},a_j \in \{0\}\cup A[k^*, t_0 - T_\text{opt}[k^*, l_0] ]\cup \{k^*\}} I[a_{j\!-\!1}, a_{j}]\geq \sum_{a_{j\!-\!1},a_j \in \{0\}\cup A'\cup \{k^*\}} I[a_{j\!-\!1}, a_{j}].
\end{align}

Thus,
\begin{align*}
\sum_{a_{j\!-\!1},a_j \in \{0\}\cup A[l_0,t_0]\cup \{l_0\}} I[a_{j\!-\!1}, a_{j}]
&= \sum_{a_{j\!-\!1},a_j \in \{0\}\cup A[k_0, t_0\!-\!T_\text{opt}[k_0,l_0]]\cup \{k_0,l_0\}} I[a_{j\!-\!1}, a_{j}]\tag{by \Cref{eq:dprecura}}\\
&\geq \sum_{a_{j\!-\!1},a_j \in \{0\}\cup A[k^*, t_0\!-\!T_\text{opt}[k^*,l_0]]\cup \{k^*,l_0\}} I[a_{j\!-\!1}, a_{j}]\tag{by the definition of $k_0$}\\
&= \sum_{a_{j\!-\!1},a_j \in \{0\}\cup A[k^*, t_0 - T_\text{opt}[k^*, l_0] ]\cup \{k^*\}} I[a_{j\!-\!1}, a_{j}] + I [k^*, l_0] \\
&\geq \sum_{a_{j\!-\!1},a_j \in \{0\}\cup A'\cup \{k^*\}} I[a_{j\!-\!1}, a_{j}] + I[k^*, l_0] \tag{by \Cref{eq:proofoptimp3}}\\
&= \sum_{a_{j\!-\!1},a_j \in \{0\}\cup A'\cup \{k^*,l_0\}} I[a_{j\!-\!1}, a_{j}]= \sum_{a_{j\!-\!1},a_j \in \{0\}\cup A^*\cup \{l_0\}} I[a_{j\!-\!1}, a_{j}].
\end{align*}

This contradicts with \Cref{eq:proofoptimp}. Therefore, our assumption that $(A[l_0,t_0], S[l_0,t_0])$ obtained using DP recurrence relation are not optimal $(A, S)$ of \Cref{eq:opt4} is false. Thus, $(A[l,t], S[l,t])$ are optimal $(A, S)$ of \Cref{eq:opt4}.
\end{proof}

\begin{proposition}
$S[l,t]$ computed from the DP recurrence relations, \Cref{eq:dprecur} is the optimal $S$ which minimizes the latency of the network when $A[l,t]$ is fixed. Concretely, $S[l,t]$ is the optimal $S$ of the optimization problem:
\begin{align}
    \label{eq:opts}
    \underset{A[l,t]\subseteq S \subseteq [l\!-\!1]}{\mathrm{min}} \sum_{s_{i\!-\!1},s_i \in \{0\}\cup S\cup \{l\}} T[s_{i\!-\!1}, s_{i}] . 
\end{align} 
\end{proposition}
\begin{proof}
When $l\!=\!1$, $S[l,t]=\emptyset$ which satisfies \Cref{eq:opts} by \Cref{eq:dprecurs}. For given $(l_0, t_0)$, we suppose for all $l<l_0$ and $t<t_0$, \Cref{eq:opts} is satisfied. Then, we assume that $S[l_0,t_0]$ obtained using \Cref{eq:dprecurs} is not optimal $S$ and  $S^*$ are the optimal $S$ of  \Cref{eq:opts} when $(l,t)=(l_0,t_0)$. Then, $A[l_0,t_0] \subseteq S^*$ and 
\begin{align}
     \sum_{s_{i\!-\!1},s_i \in \{0\}\cup S[l_0,t_0]\cup \{l_0\}} T[s_{i\!-\!1}, s_{i}] > \sum_{s_{i\!-\!1},s_i \in \{0\}\cup S^*\cup \{l_0\}} T[s_{i\!-\!1}, s_{i}] 
     \label{eq:assump0}
\end{align}
We can divide two cases whether $A[l_0,t_0]$ is an empty set or not.

\paragraph{Case1: $A[l_0,t_0]$ is an empty set} 

$S[l_0,t_0] = S_\text{opt}[0,l_0]$ by \Cref{eq:dprecurs}. Then, $S_\text{opt}[0,l_0]$ is the optimal $S$ of \Cref{eq:opts} when $(l,t)=(l_0,t_0)$ which contradicts with our assumption that $S[l_0,t_0]$ is not optimal $S$ of \Cref{eq:opts} when $(l,t)=(l_0,t_0)$.

\paragraph{Case2: $A[l_0,t_0]$ is not an empty set} 
Let $k_0$ be the maximum value of set $A[l_0,t_0]$. Then, we define $A' = A[l_0, t_0]\setminus\{k_0\}$, $S'_{<k_0}= \{s\in S^*\mid s< k_0\}$, and $S'_{>k_0}= \{s\in S^*\mid s> k_0\}$. By the definition, $A[k_0,t_0-T_\text{opt}[k_0,l_0]] \subseteq S'_{<k_0}$. Then, by the optimality assumption for $k_0<l_0$ and $t_0-T_\text{opt}[k_0,l_0]<t_0$,  
\begin{align}
\sum_{s_{i\!-\!1},s_i \in \{0\} \cup S'_{<k_0} \cup\{k_0\}} T[s_{i\!-\!1}, s_i] \geq \sum_{s_{i\!-\!1},s_i \in \{0\} \cup S[k_0, t_0-T_\text{opt}[k_0,l_0]] \cup\{k_0\}} T[s_{i\!-\!1}, s_i].
\label{eq:assump}
\end{align}

\begin{align*}
\sum_{s_{i\!-\!1},s_i \in \{0\} \cup S^* \cup\{l_0\}} T[s_{i\!-\!1}, s_i]
&= \sum_{s_{i\!-\!1},s_i \in \{0\} \cup S'_{<k_0} \cup\{k_0\}} T[s_{i\!-\!1}, s_i]
+ \sum_{s_{i\!-\!1},s_i \in \{k_0\} \cup S'_{>k_0} \cup\{l_0\}} T[s_{i\!-\!1}, s_i]\\
&\geq \sum_{s_{i\!-\!1},s_i \in \{0\} \cup S[k_0, t_0-T_\text{opt}[k_0,l_0]] \cup\{k_0\}} T[s_{i\!-\!1}, s_i] +  \sum_{s_{i\!-\!1},s_i \in \{k_0\} \cup S'_{>k_0} \cup\{l_0\}} T[s_{i\!-\!1}, s_i]\tag{by \Cref{eq:assump}}\\
&\geq \sum_{s_{i\!-\!1},s_i \in \{0\} \cup S[k_0, t_0-T_\text{opt}[k_0,l_0]] \cup\{k_0\}} T[s_{i\!-\!1}, s_i] +   \sum_{s_{i\!-\!1},s_i \in \{k_0\} \cup S_{\text{opt}} \cup\{l_0\}} T[s_{i\!-\!1}, s_i]
\tag{by \Cref{eq:timedp2}}\\
&\geq \sum_{s_{i\!-\!1},s_i \in \{0\} \cup S[k_0, t_0-T_\text{opt}[k_0,l_0]] \cup\{k_0\}\cup S_{\text{opt}} \cup\{l_0\}} T[s_{i\!-\!1}, s_i]\\
&= \sum_{s_{i\!-\!1},s_i \in \{0\} \cup S[l_0,t_0] \cup\{l_0\}} T[s_{i\!-\!1}, s_i]. \tag{by \Cref{eq:dprecurs}}
\nonumber
\end{align*}
This contradicts with \Cref{eq:assump0}. Therefore, our assumption that $S[l_0,t_0]$ obtained using DP recurrence relation is  not optimal $S$ of \Cref{eq:opts} when $(l,t)=(l_0,t_0)$ is false. Thus, $S[l,t]$ is optimal $S$ of \Cref{eq:opts}.
\end{proof}

\section{Measuring the Importance}
\label{app:imp}
\begin{minipage}{0.49\columnwidth}
\vspace{-3.7em}
\begin{algorithm}[H]
\caption{Finding Optimal Importance with DP}
\label{alg:impDP}
\footnotesize
\begin{algorithmic}
\INPUT $I$ 
\STATE Initialize $I_{\text{opt}}[k, l, a, b] \leftarrow 0, B_{\text{opt}}[k, l] \leftarrow \emptyset \quad \forall k,l,a,b$ 
\STATE $I[k,l,0,b] \leftarrow -\infty \quad \forall k,l,b, \sigma_k \neq \mathrm{id}$. 
\STATE $I[k,l,a,0] \leftarrow -\infty \quad \forall k,l,a, \sigma_l \neq \mathrm{id}$
\STATE $I[k,l,a,0] \leftarrow -\infty \quad \forall k,l,a, \sigma_k = \sigma_l = \mathrm{id}$
\FOR{$l = 1$ {\bfseries to} $L$}
    \FOR{$k = 0$ {\bfseries to} $l\!-\!1$}
        \FOR{$(a, b)$ {\bfseries in} $[(0,0), (0,1),(1,0), (1,1)]$}
            \STATE $m \leftarrow 
            \underset{k\leq m'< l}{ \mathrm{argmax} } 
            \left(I_{\text{opt}}[k, m', a, 0]+ I[m', l, 0, b] \right)$
            \STATE $I_{\text{opt}}[k, l, a, b] \leftarrow I_{\text{opt}}[k, m', a, 0]+ I[m', l, 0, b] $
            \IF{$m \notin \{k\}$}
            \STATE $B_{\text{opt}}[k, l] \leftarrow B_{\text{opt}}[k, m] \cup \{m\}$
            \ENDIF
        \ENDFOR
    \ENDFOR
\ENDFOR
\OUTPUT $I_{\text{opt}}$, $B_{\text{opt}}$
\end{algorithmic}
\end{algorithm}

\end{minipage}
\begin{minipage}{0.49\columnwidth}
\begin{algorithm}[H]
\caption{Solving the Extended Surrogate Objective}
\label{alg:extDP}
\footnotesize
\begin{algorithmic}
\INPUT $T_0, L, T, I$ 
\STATE Initialize $D[l,t,a] \leftarrow  0, A[l, t] \leftarrow \emptyset, S[l, t] \leftarrow \emptyset \quad \forall l, t, a$
\STATE $T_{\text{opt}}, S_{\text{opt}} \leftarrow \text{\Cref{alg:timeDP} }(T, L)$
\STATE $I_{\text{opt}}, B_{\text{opt}} \leftarrow \text{\Cref{alg:impDP} }(I)$
\FOR{$l = 1$ {\bfseries to} $L$}
\FOR{$t=T_{\text{opt}}[0,l]+1$ {\bfseries to} $T_0$}
\FOR{$a=0$ {\bfseries to} $1$}
    \STATE $k, \alpha \leftarrow  \underset{
    \begin{subarray}{c}
  0\leq k' <l \\
  \alpha' \in \{0, 1\}
  \end{subarray}
    }{\mathrm{argmax}} \left(
    D\left[k', t - T_{\text{opt}}[k', l], \alpha' \right] + I_\text{opt}(k', l, \alpha', a) \right)$
    \STATE $\qquad\quad~\text{subject~to~}  ~T_\text{opt}[0,k'] +T_\text{opt}[k',l]< t$
    \STATE $t_{\text{last}} \leftarrow T_{\text{opt}}[k, l]$
    \STATE $D[l, t, a] \leftarrow D[k, t -t_{\text{last}}, \alpha] + I_\text{opt}[k, l, \alpha, a]$
    \STATE $A[l, t, a] \leftarrow A[k, t- t_{\text{last}}, \alpha] \cup \{k:k>0 \, \land \, \alpha = 1\}$
    \STATE $S[l, t] \leftarrow S[k, t- t_{\text{last}}] \cup \{k:k>0\} \cup S_{\text{opt}}[k, l]$
    \STATE $B[l, t] \leftarrow B[k, t- t_{\text{last}}] \cup \{k:k>0\} \cup B_{\text{opt}}[k, l]$
\ENDFOR 
\ENDFOR 
\ENDFOR
\STATE $a_\text{last} \leftarrow \underset{a \in \{0, 1\}}{\mathrm{argmax}} (A[L, T_0, a])$
\OUTPUT $A[L, T_0, a_\text{last}], S[L, T_0], B[L, T_0]$
\end{algorithmic}
\end{algorithm}

\end{minipage}

\subsection{Extension on the Importance}
\label{app:imp-ext}

In \Cref{method}, the $j$-th activation layer of our target network for depth compression is either the $j$-th activation layer in the vanilla network ($\sigma_j$) or an identity function ($\mathrm{id}$). Thus, if $\sigma_j=\mathrm{id}$, the $j$-th activation layer in the target network is inherently an identity function.
For instance, MobileNetV2 has an identity function as an activation layer at the end of each Inverted Residual Block and the corresponding activation layers in our target network are bound to be $\mathrm{id}$ \citep{mobilenet}.
On the other hand, non-linear activation layers at the end of the Inverted Residual Block can improve the performance of the networks compressed from MobileNetV2 \citep{depthshrinker}. 
To this end, we extend the search space of our method by further introducing the network blocks that have a non-linear activation layer at these positions and incorporating them into the DP formulation.

\clearpage
Consider a network block from $i\!+\!1$-th layer to $j$-th layer. 
We introduce discrete variables $d_i, d_j \in \{0,1\}$ to indicate whether the first and the last activation layer of the network block are identity functions or not, respectively.
If $\sigma_i$ and $\sigma_j$ are not identity functions, then we limit $d_i$ and $d_j$ to $1$, respectively.
Then, we redefine the importance of the network block between $i\!+\!1$-th layer to $j$-th layer as $I[i, j, d_i, d_j]$.

Concretely, we redefine the importance as follows:
{\footnotesize
\begin{align} I[i, j, d_i, d_j]  &\coloneqq \max_\theta
          \mathrm{Acc}\left(
          \underbrace{\BigO_{l=j+1}^{L} \sigma_l \circ f_{\theta_l}}_{\text{$j\!+\!1$ to $L$ layers}}
          \circ 
          \underbrace{\left(d_j\sigma + (1-d_j) \mathrm{id} \right)}
          _{\text{$j$-th activation}}
          \circ \underbrace{\BigO_{l=i+1}^{j} f_{\theta_l}}_{\text{$i\!+\!1$ to $j$ layers}}
          \circ 
          \underbrace{\left(d_i\sigma + (1-d_i) \mathrm{id} \right)}_{\text{$i$-th activation}}
          \circ
          \underbrace{f_{\theta_i}
          \circ\BigO_{l=1}^{i\!-\!1}  \sigma_l \circ f_{\theta_l} }_{\text{$1$
        to $i$ layers}}\right)\nonumber\\
        &- \max_\theta
 \mathrm{Acc}\left( 
    \BigO_{l=1}^L \sigma_l \circ f_{\theta_l} \right),
\label{eq:ext-imp}
\end{align}}
\hspace{-0.2em}where $\sigma$ is the activation layer that is not an identity function.


Due to the redefinition of importance, we propose an alternative surrogate for objective in \Cref{eq:opt1_obj} as follows: 
\begin{align}
& \mathcal{I}(A, B) \coloneqq 
\sum_{b_{j\!-\!1},b_j \in \{0\}\cup B\cup \{L\}} I[b_{j\!-\!1}, b_{j}, \mathds{1}_{A} (b_{j-1}), \mathds{1}_{A}(b_j)],
\label{eq:ext-sur}
\end{align}
where $A \subseteq B \subseteq [L-1]$. $A$ denotes the positions of activations which are not identity functions and $B$ denotes the boundary points of the contiguous network blocks for objective approximation. Then, the objective extends to 
\begin{align}
\label{eq:opt-ext}
&\underset{
A\subseteq B, S \subseteq [L-1] }{\mathrm{maximize}}  \sum_{b_{j\!-\!1},b_j \in \{0\}\cup B\cup \{L\}} \!I[b_{j\!-\!1}, b_{j}, \mathds{1}_{A}(b_{j-1}), \mathds{1}_{A}(b_j)] \\ 
&\mathrm{subject\ to\ } ~~ \sum_{s_{i\!-\!1},s_i \in \{0\}\cup S\cup \{L\}} T[s_{i\!-\!1}, s_{i}] < T_0. \nonumber
\end{align}
Note, \Cref{eq:opt-ext} can be exactly solved with DP algorithm analogously by \Cref{alg:extDP}.


\subsection{Possible Combinations of Network Blocks}

In MobileNetV2, we empirically observed that the network blocks with identity functions on both edges unnecessarily degrade the performance by excessively reducing the number of activation functions in the compressed network. To address this issue, we set the importance value of the network blocks to negative infinity if $\sigma_i = \sigma_j = \mathrm{id}, d_j = 0$ and exclude them in the DP algorithm.

Furthermore, we only consider blocks that we can merge into a single layer; thus, the skip-connections in MobileNetV2 considerably reduce the number of possible blocks. 
We also avoid merging in scenarios where a convolutional layer with a kernel size larger than 1 follows the stride 2 convolutional layer since it leads to a significant increase in kernel size \citep{depthshrinker}.
In MobileNetV2, we have 171 different blocks to measure the latency ($T[i, j]$) and 315 different blocks to measure the importance ($I[i, j, d_i, d_j]$).

\subsection{Evaluating and Normalizing the Importance}
\label{subsec:impnorm}

When we evaluate the importance value in \Cref{eq:ext-imp}, we approximate the first term by substituting the activation layers within the block to identity functions and training the network for a few epochs from the pretrained weight.
The second term is considered as the accuracy of the pretrained weight itself. 
In MobileNetV2, we approximate the first term in \Cref{eq:ext-imp} by training the deactivated network for a single epoch.
If the block size is one (\emph{i.e.}, $k-l = 1$), we re-initialize the corresponding block and measure the accuracy drop after training it from the pretrained weight.

When we approximate the first term in \Cref{eq:ext-imp} with the accuracy attained after training it for a few epochs, we tend to calculate a lower importance value than the actual definition of the importance value. This effect is reflected independently for each block; thus, the more block we construct the network with, the more we underestimate the actual importance of the network. Therefore, it is crucial to normalize the importance values by adding an appropriate value to the importance of each block to address this issue. 
To this end, we add the constant multiple of the average importance of the blocks of size one to normalize the importance of each block.

Concretely, we define the set $D$ as 

\begin{align*}
    D =  \bigg\{&\mathrm{Acc}\left(\texttt{one-epoch}\left(
          f\right)\right) - \max_\theta \mathrm{Acc}\left(\BigO_{l=1}^L \sigma_l \circ f_{\theta_l} \right) \bigg|\\ & f = \BigO_{l=i+1}^{L} \sigma_l \circ f_{\theta_l}
          \circ 
          \left(d_{i+1}\sigma + (1-d_{i+1}) \mathrm{id} \right)
          \circ 
          f_{\theta'}
          \circ 
          \left(d_i\sigma + (1-d_i) \mathrm{id} \right)
          \circ\BigO_{l=1}^{i\!-\!1}  \sigma_l \circ f_{\theta_l} \nonumber \\
          & \text{ for } i \in [L-1], \theta' = \texttt{init}(\theta_i), \text{ and } d_i, d_{i+1} \in \{0, 1\} \bigg\}, \nonumber 
\end{align*}
where $\texttt{one-epoch}(\;\cdot\;)$ denotes the network trained for single epoch and $\texttt{init}(\;\cdot\;)$ denotes the initializing function.
Then, we normalize the importance value by
\begin{align}
    I[l, k, a, b] \leftarrow I[l, k, a, b] - \frac{\alpha}{|D|} \sum_{\Delta \mathrm{acc}\in D} \Delta \mathrm{acc}, \nonumber
\end{align}
where $\alpha$ is the hyperparameter.

\begin{table}[b!]
\caption{Accuracy and latency of compressed architectures applied to MobileNetV2-1.0 and MobileNetV2-1.4 on ImageNet-100 dataset. 
Compression methods use the latency information of  \textit{RTX 2080 Ti} and the latency is measured on the \textit{RTX 2080 Ti} with batch size of 128. We report the average accuracy of the three runs of finetuning.}
\begin{subtable}[t]{0.48\columnwidth}
\caption{MobileNetV2-1.0 }
\vspace{0.5em}
\centering
\begin{adjustbox}{max width=1.0\columnwidth}
\begin{tabular}{lcccc}
\toprule
                            && \normalsize{TensorRT}
                            & \footnotesize{\textit{w/o}} \normalsize{TensorRT 
 }\\
                        Network     & Acc (\%)& Lat. (ms) & Lat. (ms)\\
 \cmidrule(r){1-2}\cmidrule(r){3-3} \cmidrule(r){4-4}
    MBV2-1.0  & 87.58 & 19.25 &  40.61 \\
 \cmidrule(r){1-2}\cmidrule(r){3-3} \cmidrule(r){4-4}
    DS-AR-1.0 & 86.89 & 11.86 & 21.14 \\
    Ours& \textbf{86.93}  & \textbf{11.35} & \textbf{20.65} \\
  \cmidrule(r){1-2}\cmidrule(r){3-3} \cmidrule(r){4-4}
    DS-BR-1.0 & 86.45 & 11.73 & \textbf{19.44}\\
    Ours      & \textbf{86.53} & \textbf{11.08} & 19.97 \\
 \cmidrule(r){1-2}\cmidrule(r){3-3} \cmidrule(r){4-4}
    DS-CR-1.0 & 85.38 & 10.47 & \textbf{16.54} \\
    Ours      & \textbf{85.91} & \textbf{9.62} & 16.82 \\
\bottomrule
\end{tabular}
\end{adjustbox}
\label{tab:rpr-in100-mbv2-1.0}
\end{subtable}
\begin{subtable}[t]{0.48\columnwidth}
\caption{MobileNetV2-1.4} 
\vspace{0.5em}
\centering
\begin{adjustbox}{max width=1.0\columnwidth}
\begin{tabular}{lccc}
\toprule
                            && \normalsize{TensorRT}
                            & \footnotesize{\textit{w/o}} \normalsize{TensorRT }\\
                        Network     & Acc (\%)& Lat. (ms)
                            & Lat. (ms) \\
 \cmidrule(r){1-2}\cmidrule(r){3-3} \cmidrule(r){4-4}
    MBV2-1.4  & 88.88 & 29.94 & 61.68 \\
 \cmidrule(r){1-2}\cmidrule(r){3-3} \cmidrule(r){4-4}
    DS-AR-1.4  & 87.57 & 19.55 & 34.58 \\
    Ours& \textbf{88.05}  & \textbf{19.31} & \textbf{33.08} \\
 \cmidrule(r){1-2}\cmidrule(r){3-3} \cmidrule(r){4-4}
    DS-BR-1.4 & 86.23 & 18.22 & 28.77 \\
    Ours& \textbf{87.18}  & \textbf{16.26} & \textbf{27.42} \\
 \cmidrule(r){1-2}\cmidrule(r){3-3} \cmidrule(r){4-4}
    DS-CR-1.4 & 84.85 & 17.21 &  26.07\\
    Ours      & \textbf{85.93} & \textbf{14.65} & \textbf{22.96} \\
\bottomrule
\end{tabular}
\end{adjustbox}
\label{tab:rpr-in100-mbv2-1.4}
\end{subtable}
\end{table}

\section{Additional Experiments}
\subsection{Reproducing the Search Phase of DepthShrinker on ImageNet-100}
\label{app:in100-ds}

We reproduce the search phase of DepthShrinker on top of the ImageNet-100 dataset and search the patterns that match the compression ratio in the original paper \citep{depthshrinker}.
In MobileNetV2-1.0, we sweep through the number of activated blocks among 12, 9, and 7 and denote them `DS-AR-1.0', `DS-BR-1.0', and `DS-CR-1.0', respectively. In MobileNetV2-1.4, we sweep through the number of activated blocks among 11, 8, and 6 and name them `DS-AR-1.4', `DS-BR-1.4', and `DS-CR-1.4', respectively. \Cref{tab:rpr-in100-mbv2-1.0} and \Cref{tab:rpr-in100-mbv2-1.4} summarize the results of comparing our method to the reproduced result of DepthShrinker for MobileNetV2-1.0 and MobileNetV2-1.4 on the ImageNet-100 dataset, respectively. Our method outperforms the baseline performance in TensorRT format regardless of the type of network and compression ratio.

\subsection{Inference Time Transfer Results on Different GPUs}
\label{app:gpus}

\begin{table}[h]
\caption{Accuracy and latency of compressed architectures applied to MobileNetV2-1.0 and MobileNetV2-1.4 on ImageNet-100 dataset. The latency of the compressed network architecture is measured on \textit{TITAN Xp}, \textit{RTX 2080 Ti}, \textit{RTX 3090}, and  \textit{Tesla V100} with batch size of 128. We report the average accuracy of the three runs of finetuning.}
\begin{subtable}[t]{\columnwidth}
\caption{MobileNetV2-1.0}
\centering
\begin{adjustbox}{max width=0.8\columnwidth}
\begin{tabular}{lcccccc}
\toprule
                            & 
                            & \multicolumn{4}{c}{TensorRT Latency (ms)} 
                            & \footnotesize{\textit{w/o} \normalsize{TensorRT (ms)}}\\
                            Network & Acc (\%) & \emph{TITAN Xp} & \emph{RTX 2080 Ti} & \emph{RTX 3090} 
                            & \emph{Tesla V100} 
                            & \emph{RTX 2080 Ti} \\
\cmidrule(r){1-2}\cmidrule(r){3-6} \cmidrule(r){7-7}
    MobileNetV2-1.0  & 87.58 & 26.98 & 19.25 & 13.31 & 15.49 & 40.61 \\
\cmidrule(r){1-2}\cmidrule(r){3-6} \cmidrule(r){7-7}
    MBV2-DS-A  &  87.58  & 19.92 & 14.74 & 9.70 & 11.71 &  27.59  \\
    Ours& \textbf{87.69} & \textbf{17.90} & \textbf{12.53} & \textbf{8.74} & \textbf{10.22} & \textbf{23.02} \\
\cmidrule(r){1-2}\cmidrule(r){3-6} \cmidrule(r){7-7}
    MBV2-DS-B & 87.31 & 18.02 & 12.33 & 8.73 & 10.03 & 22.99  \\
    Ours      & \textbf{87.45} & \textbf{16.93} & \textbf{12.11} & \textbf{8.43} & \textbf{9.93} & \textbf{22.29}  \\
\cmidrule(r){1-2}\cmidrule(r){3-6} \cmidrule(r){7-7}
    MBV2-DS-C & 85.92 & 15.59 & 11.20 & 7.84 & \textbf{9.08} & 20.76  \\
    Ours      & \textbf{86.73} & \textbf{15.27} & \textbf{11.14} & \textbf{7.77} & 9.18 & \textbf{20.62}\\
\cmidrule(r){1-2}\cmidrule(r){3-6} \cmidrule(r){7-7}
    MBV2-DS-D & 85.30 & 14.38 & 10.49 & 7.27 & 8.56 &  18.78 \\
    Ours      & \textbf{85.91} & \textbf{13.45} & \textbf{9.62} & \textbf{6.71} & \textbf{8.06} & \textbf{16.82} \\
\bottomrule
\end{tabular}
\end{adjustbox}
\label{tab:in100-mbv2-1.0}
\end{subtable}
\\
\\
\centering
\begin{subtable}[t]{\columnwidth}
\caption{MobileNetV2-1.4}
\centering
\begin{adjustbox}{max width=.8\columnwidth}
\begin{tabular}{lcccccc}
\toprule
                            & & \multicolumn{4}{c}{TensorRT Latency (ms)}   
                            & \footnotesize{\textit{w/o}} \normalsize{TensorRT (ms)}\\Network
                            & Acc (\%) & \emph{TITAN Xp} & \emph{RTX 2080 Ti} & \emph{RTX 3090} 
                            & \emph{Tesla V100} 
                            & \emph{RTX 2080 Ti} \\
\cmidrule(r){1-2}\cmidrule(r){3-6} \cmidrule(r){7-7}
    MobileNetV2-1.4  & 88.88 & 42.15 & 29.94 & 20.67 & 24.29  & 61.68 \\
\cmidrule(r){1-2}\cmidrule(r){3-6} \cmidrule(r){7-7}
    MBV2-1.4-DS-A  &  88.01  & 26.90 & 19.61 & 13.54 & \textbf{16.05}  &  35.06 \\
    Ours& \textbf{88.41} & \textbf{26.53} & \textbf{19.48} & \textbf{13.28} & 16.45 & \textbf{34.01} \\
\cmidrule(r){1-2}\cmidrule(r){3-6} \cmidrule(r){7-7}
    MBV2-1.4-DS-B  &  86.99  & 25.27 & 19.21 & 13.19 & 15.96 & 31.63 \\
    Ours& \textbf{87.58} & \textbf{24.60} & \textbf{18.22} & \textbf{12.48} & \textbf{15.39} & \textbf{30.77} \\
\cmidrule(r){1-2}\cmidrule(r){3-6} \cmidrule(r){7-7}
    MBV2-1.4-DS-C & 86.73 & 23.64 & 17.47 & 12.08 & 14.52 &  29.73 \\
    MBV2-1.4-DS-D & 86.05 & 22.69 & 17.50 & 12.00 & 14.48 & 27.99  \\
    Ours      & \textbf{87.18} & \textbf{22.03} & \textbf{16.26} & \textbf{11.07} & \textbf{13.50} & \textbf{27.42} \\
\cmidrule(r){1-2}\cmidrule(r){3-6} \cmidrule(r){7-7}
    MBV2-1.4-DS-E & 85.29 & 20.94 & 15.67 & 10.87 & 13.03 &  26.08 \\
    Ours      & \textbf{85.93} & \textbf{19.35} & \textbf{14.65} & \textbf{9.77} & \textbf{12.27} & \textbf{22.96} \\
\bottomrule
\end{tabular}
\end{adjustbox}
\label{tab:in100-mbv2-1.4}
\end{subtable}
\end{table}

\begin{table*}[h!]
\caption{Accuracy and latency of compressed architectures applied to MobileNetV2-1.0 on ImageNet dataset. The latency of the compressed network architecture is measured on \textit{TITAN Xp}, \textit{RTX 2080 Ti}, \textit{RTX 3090}, and  \textit{Tesla V100} with batch size of 128. $\dagger$ denotes the accuracy of the pretrained weight used in DepthShrinker, and we use the same pretrained weight for a fair comparison.}
\vspace{0.5em}
\centering
\begin{adjustbox}{max width=.8\columnwidth}
\begin{tabular}{lcccccccc}
\toprule
    &
                            & \multicolumn{4}{c}{TensorRT Latency (ms)} 
                            & \footnotesize{\textit{w/o} \normalsize{TensorRT (ms)}}\\
                            \cmidrule(r){3-7}\cmidrule(r){8-9}
               Network             & Acc (\%)& \emph{TITAN Xp} & \emph{RTX 2080 Ti} & \emph{RTX 3090} 
                            & \emph{Tesla V100}  
                            & \emph{RTX 2080 Ti}\\
\cmidrule(r){1-2}\cmidrule(r){3-6}\cmidrule(r){7-7}
    MobileNetV2-1.0 & 72.89$^\dagger$ & 27.03 & 19.26 & 13.39 & 15.50 & 40.71 \\
\cmidrule(r){1-2}\cmidrule(r){3-6}\cmidrule(r){7-7}
    MBV2-DS-A  & 72.37  & 20.01 &  14.82 & 9.69 & 11.76 & 27.53 \\
    Ours       & \textbf{72.83} & \textbf{19.53}& \textbf{13.67} & \textbf{9.64} & \textbf{11.16} & \textbf{25.09} \\
\cmidrule(r){1-2}\cmidrule(r){3-6}\cmidrule(r){7-7}
    MBV2-DS-B & 71.96 & \textbf{17.80} & 12.42 & 8.75 & \textbf{10.07} & 22.92 \\
    Ours      & \textbf{72.13} & 18.43 & \textbf{12.38} & \textbf{8.67} & 10.27 & \textbf{21.74} \\
\cmidrule(r){1-2}\cmidrule(r){3-6}\cmidrule(r){7-7}
    MBV2-DS-C & 70.87 & 15.76 & 11.28 & 7.87 & 9.12 & 20.77 \\
    Ours      & \textbf{71.44} & \textbf{15.23} & \textbf{10.90}  & \textbf{7.69} & \textbf{8.98} & \textbf{19.75} \\
\cmidrule(r){1-2}\cmidrule(r){3-6}\cmidrule(r){7-7}
    MBV2-DS-D & 69.43 & 14.38 & 10.53 & 7.27 & 8.56 & 18.82 \\
    Ours      & \textbf{70.65} & \textbf{14.21} & \textbf{9.88} & \textbf{6.99} & \textbf{8.31} & \textbf{16.55} \\
\bottomrule
\end{tabular}
\end{adjustbox}
\vspace{-1.5em}
\label{tab:in-mbv2-1.0}
\end{table*}

In this section, we present the results of measuring the end-to-end inference time across different GPU devices.
The compression of networks utilizes the latency information obtained from the \textit{RTX 2080 Ti} GPU.
We report the latency on \textit{TITAN Xp}, \textit{RTX 2080 Ti}, \textit{RTX 3090}, and \textit{Tesla V100}. 
\Cref{tab:in100-mbv2-1.0} and \Cref{tab:in100-mbv2-1.4} summarize the accuracy and the latency of the networks compressed on the ImageNet-100 dataset. 
We further present the results of compressing MobileNetV2-1.0 on ImageNet dataset in \Cref{tab:in-mbv2-1.0}.
Our method outperforms the baseline in the majority of the settings.
\clearpage

\subsection{Comparison with Channel Pruning Baselines}
\begin{wraptable}[16]{r}{0.51\textwidth} 
\begin{minipage}{\linewidth}
\vspace{-3.5em}
\caption{Accuracy and latency of compressed architectures applied to MobileNetV2-1.0 and MobileNetV2-1.4 on ImageNet dataset. The latency is measured on \textit{RTX 2080 Ti} with batch size of 128.}
\vspace{0.5em}
\begin{tabular}{lccc}
\toprule
                            && \normalsize{TensorRT}
                            & \footnotesize{\textit{w/o}} \normalsize{TensorRT }\\
                        Network     & Acc (\%)& Lat. (ms)
                            & Lat. (ms) \\
 \cmidrule(r){1-2}\cmidrule(r){3-3} \cmidrule(r){4-4}
    MBV2-1.0 & 72.89 & 19.26 & 40.71 \\
    \cmidrule(r){1-2}\cmidrule(r){3-3} \cmidrule(r){4-4}
    Uniform $L^1$& 72.65 & 15.05 & 32.10 \\
    AMC (70\% FLOPs)  & 72.01 & 14.40 & 30.81\\
    Ours & \textbf{72.83} & \textbf{13.67} & \textbf{25.09}\\
\midrule \\ [-3.5ex]
\midrule
    MBV2-1.4 & 76.28 & 29.93 & 61.64 \\
    \cmidrule(r){1-2}\cmidrule(r){3-3} \cmidrule(r){4-4}
    Uniform $L^1$& 74.80 & 20.86 & 42.25 \\
    Ours & \textbf{75.16} & \textbf{19.76} & \textbf{35.07}\\
    \cmidrule(r){1-2}\cmidrule(r){3-3} \cmidrule(r){4-4}
    MetaPruning-1.0$\times$  & 73.69 & 21.75 & 38.70\\
    Ours & \textbf{74.68} & \textbf{18.63} & \textbf{32.35}\\
    \bottomrule[1pt]
\end{tabular}
\label{tab:channel}
\end{minipage}
\end{wraptable}
In this section, we compare our depth compression method with the channel pruning baselines. 
We start from the same pretrained weight and finetune with the identical training protocol described in \Cref{subsec:detail}.
In MobileNetV2-1.0, we compare with uniform $L^1$ pruning and AMC \citep{amc}.
For the uniform $L^1$ pruning, we leave 75\% of the output channels based on $L^1$-norm in the first convolution layer of each Inverted Residual Block and leave the other convolution layers in the block \citep{pfec,rethinking}. 
For AMC, we prune each convolutional layer according to the channel ratio of the AMC network (70\% FLOPs).
In MobileNetV2-1.4, we compare with uniform $L^1$ pruning and MetaPruning \citep{metaprun}.
For the uniform $L^1$ pruning, we leave 65\% of the output channels with the same protocol.
For MetaPruning, we prune each convolutional layer according to the channel ratio of the MetaPruning network (MetaPruning-1.0$\times$).
It is worth noting that we finetune from a pretrained weight pruned based on the $L^1$-norm in reproducing the MetaPruning, while the original method trains the network from scratch.
We choose to reproduce this way since it leads to better accuracy.
\Cref{tab:channel} demonstrates that our method outperforms the channel pruning baselines consistently.

\subsection{Depth Compression Results on VGG19 Network}
\begin{wraptable}[8]{r}{0.43\textwidth} 
\begin{minipage}{\linewidth}
\centering
\vspace{-4em}
\caption{Accuracy and latency of compressed architectures applied to VGG19 on ImageNet dataset. The latency is measured on \textit{RTX 2080 Ti} with batch size of 64.}
\vspace{0.5em}
\begin{tabular}{lccc}
\toprule
                        Network     & Accuracy (\%) & Latency (ms) \\
 \cmidrule(r){1-2}\cmidrule(r){3-3}
    VGG19 & 74.24 & 131\\
    \cmidrule(r){1-2}\cmidrule(r){3-3} 
    Ours& 74.99 & 111 \\
    & 74.33 & 91\\
    & 73.00 & 84 \\
    \bottomrule[1pt]
\end{tabular}
\label{fig:prem}
\end{minipage}
\end{wraptable}
In this section, we present the results of applying our depth compression method to the VGG19 network on the ImageNet dataset \citep{vgg,imagenet}.
We compress the depth of the network utilizing the latency information of \textit{RTX 2080 Ti} and measure the latency on the same \textit{RTX 2080 Ti}.
We finetune the network for 20 epochs using cosine learning rate decay with the SGD optimizer.
As a result, we attain 1.44$\times$ speed-up without losing any accuracy.

\subsection{FLOPs and Run-time Memory Results}
\begin{wraptable}[16]{r}{0.45\textwidth} 
\begin{minipage}{\linewidth}
\vspace{-3.5em}
\centering
\caption{FLOPs and run-time memory usage of compressed architectures applied to MobileNetV2-1.0 on ImageNet dataset. Memory usage is measured with batch size of 128.}
\vspace{0.5em}
\begin{tabular}{lccc}
\toprule
    &  &  & Run-time \\
Network     & Acc (\%) & MFLOPs & Mem. (GB) \\
 \cmidrule(r){1-2}\cmidrule(r){3-3}\cmidrule(r){4-4}
    MBV2-1.0 & 72.89 & 302 & 6.88\\
 \cmidrule(r){1-2}\cmidrule(r){3-3} \cmidrule(r){4-4}
    DS-A-1.0  & 72.37 & 315 &  4.21 \\
    Ours& \textbf{72.83}  & \textbf{291} & \textbf{3.93} \\
 \cmidrule(r){1-2}\cmidrule(r){3-3} \cmidrule(r){4-4}
    DS-B-1.0  & 71.96 & \textbf{258} &  3.63  \\
    Ours& \textbf{72.13}  & 282 & \textbf{3.35} \\
 \cmidrule(r){1-2}\cmidrule(r){3-3} \cmidrule(r){4-4}
    DS-C-1.0 & 70.87 & 248 &  3.31 \\
    Ours      & \textbf{71.44} & \textbf{247} & \textbf{3.16} \\
 \cmidrule(r){1-2}\cmidrule(r){3-3} \cmidrule(r){4-4}
    DS-D-1.0 & 69.43 & \textbf{243} &  2.95 \\
    Ours      & \textbf{70.65} & 247 & \textbf{2.55} \\
    \bottomrule[1pt]
\end{tabular}
\label{fig:prem}
\end{minipage}
\end{wraptable}
In this section, we report the FLOPs and the peak run-time memory usage of our compressed networks compared to the baseline method DepthShrinker \citep{depthshrinker}.
We present the results of applying compression methods to MobileNetV2-1.0 on the ImageNet dataset.
We highlight that our method directly optimizes for the wall clock inference time and therefore did not optimize for the FLOPs.
Although our method does not strictly have fewer FLOPs than the baseline method, our method outperforms the baseline in peak run-time memory, which is related to real-hardware efficiency.
It is worth noting that the FLOPs values we report differ from the baseline works because we report the FLOPs at the test time after fusing the batch normalization layers into the convolutional layers \citep{depthshrinker,mobilenet}.
Furthermore, DepthShrinker's official implementation omits to merge the first Inverted Residual Block in the `DS-A-1.0' network; we measure the test time FLOPs after we merge it following their paper \citep{depthshrinker}.
We report the FLOPs at the test time because the objective of our method is to obtain an efficient network with low latency at the test time.

\clearpage
\subsection{Latency on CPU Device}
\begin{wraptable}[11]{r}{0.4\textwidth} 
\begin{minipage}{\linewidth}
\centering
\vspace{-4em}
\caption{Accuracy and CPU latency of compressed architectures applied to MobileNetV2-1.0 on ImageNet dataset. The latency is measured on 5 Intel Xeon Gold 5220R CPU cores with batch size of 128.}
\vspace{0.5em}
\begin{adjustbox}{max width=.6\columnwidth}
\begin{tabular}{lcc}
\toprule
Network     & Accuracy (\%) & Latency (ms) \\
 \cmidrule(r){1-2}\cmidrule(r){3-3}
    MBV2-1.0 & 72.89 & 1386\\
 \cmidrule(r){1-2}\cmidrule(r){3-3} 
    DS-A-1.0  & 72.37 &  837\\
    Ours& \textbf{72.83}  & \textbf{710}  \\
 \cmidrule(r){1-2}\cmidrule(r){3-3} 
    DS-B-1.0  & 71.96 &  713 \\
    Ours& \textbf{72.13}  &  \textbf{596} \\
 \cmidrule(r){1-2}\cmidrule(r){3-3} 
    DS-C-1.0 & 70.87 &  644 \\
    Ours      & \textbf{71.44} & \textbf{566} \\
 \cmidrule(r){1-2}\cmidrule(r){3-3} 
    DS-D-1.0 & 69.43 &  592 \\
    Ours      & \textbf{70.65} &  \textbf{470} \\
    \bottomrule[1pt]
\end{tabular}
\end{adjustbox}
\end{minipage}
\end{wraptable}
In this section, we present the CPU latency of our compressed networks compared to the baseline method DepthShrinker \citep{depthshrinker}.
We present the results of applying compression methods to MobileNetV2-1.0 on the ImageNet dataset.
We measure the latency on 5 Intel Xeon Gold 5220R CPU cores with batch size of 128.
Our method attains higher accuracy with lower latency compared to DepthShrinker, regardless of the compression ratio.
Specifically, our method attains 1.95$\times$ speed-up with 0.06\%p accuracy drop from the pretrained weight and attains 1.18 $\times$ speed-up with higher accuracy compared to DS-A-1.0.

\subsection{Analysis on the Latency Reduction}
\begin{wraptable}[10]{r}{0.51\textwidth} 
\begin{minipage}{\linewidth}
\vspace{-3.5em}
\centering
\caption{Analysis on the latency reduction from removing activation layers and merging convolutional layers. The latency is measured on \textit{RTX 2080 Ti} with batch size of 128.}
\vspace{0.5em}
\begin{adjustbox}{max width=\columnwidth}
\label{tab:abl-act}
\begin{tabular}{lccc}
\toprule
                            && \normalsize{TensorRT}
                            & \footnotesize{\textit{w/o}} \normalsize{TensorRT }\\
                        Network     & Acc (\%)& Lat. (ms)
                            & Lat. (ms) \\
 \cmidrule(r){1-2}\cmidrule(r){3-3} \cmidrule(r){4-4}
    Original & 72.89 & 19.55 & 41.03 \\
    \cmidrule(r){1-2}\cmidrule(r){3-3} \cmidrule(r){4-4}
    After removing activation & 72.13 & 19.55 &  35.15\\
    After merging convolution  &  & 12.52 & 21.88 \\
    \cmidrule(r){1-2}\cmidrule(r){3-3} \cmidrule(r){4-4}
    After removing activation & 70.65 & 19.55 & 33.69 \\
    After merging convolution &  & 9.92 & 16.60\\
    \bottomrule[1pt]
\end{tabular}
\end{adjustbox}
\label{fig:prem}
\end{minipage}
\end{wraptable}
After finetuning the network, two different factors can contribute to the reduction in latency at the test time.
The first factor involves replacing the activation layer with the identity function, and the second is merging consecutive convolutional layers.
We present the results of the latency reduction incurred by these two factors in \Cref{tab:abl-act}.
While removing activations partially contributes to a latency reduction without TensorRT, its impact becomes negligible in TensorRT format.
This is because TensorRT fuses non-linear activation layers with the preceding convolutional layers \citep{tensorrt}.
In the main paper, we optimize the inference time of the network in the TensorRT implementation and do not consider the latency of the activation layer in our formulation.

\section{Hyperparameters}
\begin{table*}[t]
\caption{Hyperparameters used in our method. We use $\alpha$ in normalizing the importance value and use $T_0$ as the constraint of \Cref{eq:opt1}.}
\vspace{0.5em}
\label{tab:hyper}
\centering
\begin{adjustbox}{max width=0.9\columnwidth}
\begin{tabular}{llcccllccc}
\toprule
Dataset &
Table (Network) & Acc (\%)& $\alpha$ & $T_0$ & 
Dataset &
Table (Network) & Acc (\%)& $\alpha$ & $T_0$ \\
\cmidrule(r){1-1}\cmidrule(r){2-5}\cmidrule(r){6-6}\cmidrule(r){7-10}
ImageNet-100 &
\Cref{tab:small-in100-mbv2-1.4} (MBV2-1.0) & 87.69 & 1.8 & 23.0 & 
ImageNet &
\Cref{tab:small-in-mbv2-1.0} (MBV2-1.0) & 72.83 & 1.6 & 25.0 \\
&     & 87.45 & 1.8 & 22.0 & & & 72.13 & 1.6 & 22.1 \\
&     & 86.73 & 1.8 & 20.5 & & & 71.44 & 1.6 & 20.0 \\
&     & 85.91 & 1.8 & 17.5 & & & 70.65 & 1.6 & 18.0 \\
\cmidrule(r){2-5}\cmidrule(r){7-10}
&
\Cref{tab:small-in100-mbv2-1.4} (MBV2-1.4) &  88.41  & 1.6  & 28.0 & &
\Cref{tab:in-mbv2-1.4} (MBV2-1.4) & 74.68 & 1.2 & 27.0\\
&    & 87.58 & 1.6 & 26.0 & & & 74.19 & 1.2 & 26.0 \\
&    & 87.18 & 1.6 & 23.0 & & & 73.46 & 1.2 & 23.0\\
&    & 85.93 & 1.6 & 20.0 & & & 72.57 & 1.2 & 20.0\\
\bottomrule
\end{tabular}
\end{adjustbox}
\end{table*}

In this section, we present the values of hyperparameters that can reproduce the results of our method in \Cref{tab:hyper}.
Specifically, our method has two hyperparameters $\alpha$ and $T_0$ in optimizing the ordered set $A$ and $S$.
First hyperparameter $\alpha$ works in normalizing importance value of each block. \Cref{subsec:impnorm} describes the detailed process of normalizing the importance value of each block using the hyperparameter $\alpha$.
Second hyperparameter $T_0$ serves the inference time constraint when we solve the \Cref{eq:opt1}.

During finetuning, we finetune the network for 180 epochs using cosine learning rate decay with the SGD optimizer and batch size of 256. 
For the networks compressed on the ImageNet-100 and the compressed MobileNetV2-1.4 on the ImageNet, we finetune using the base learning rate of 0.1, weight decay of 1e-5 and adopt the label smoothing, random erasing and RandAugment following the \citet{depthshrinker} \citep{labelsmoothing,randerase,randaug}. 
For the compressed MobileNetV2-1.0 on ImageNet dataset, we use the base learning rate of 0.05, weight decay of 1e-5 without adopting further improved augmentation techniques, since they did not improve the performance.

\section{Merging Convolutional Layers in Modern CNN}
\label{subsec:mergedetail}
\subsection{Skip Addition}
We address the details to apply the merging for the convolution operations in modern CNNs with skip addition and padding. Consider a skip addition, $f(x) + x$ where $f(\cdot)$ is a network block and $X$ is an input feature map. When $f(\cdot)$ is a single convolution operation, $f(x)\!+\!x$ can be replaced by an equivalent convolution operation \citep{repvgg}. In light of this, our method fuses the skip addition into $f(\cdot)$ only if $f(x)$ is merged into a single convolution operation. 

\subsection{Padding Reordering Technique}
\label{sup:padreorder}
DepthShrinker's scope of merging convolution operations is restricted to cases where the kernel size of at least one of the convolution operations to be merged is 1 \citep{depthshrinker}. To include more general cases of merging where the kernel size of both convolution operations is greater than 1, we need to address the details of padding. In this paper, we limit our considerations to zero padding for the exact merging and apply sufficient zero padding to prevent the computation disparities at the boundaries before and after merging.


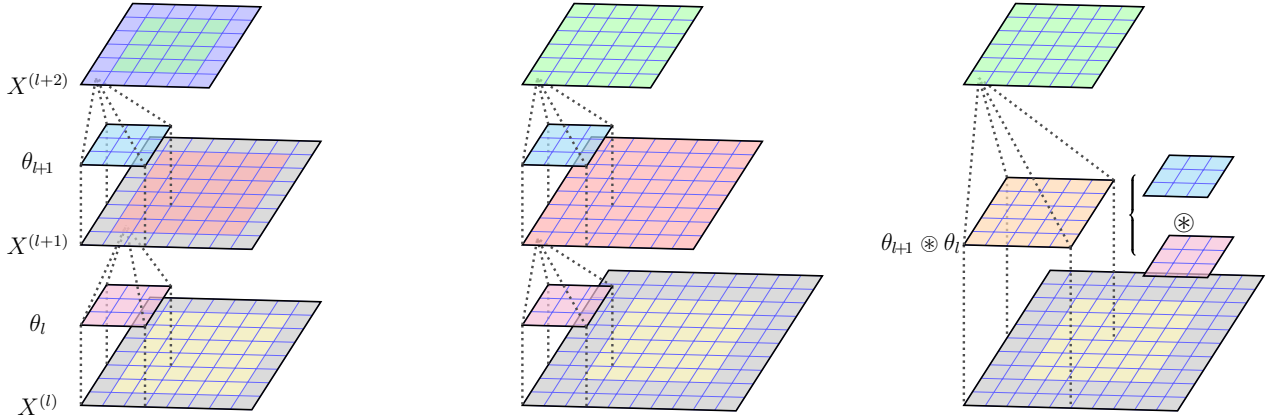
\begin{figure*}[t]
\def\zcolor{black}
\def\fcolor{yellow}
\def\fcolora{red}
\def\fcolorb{green}
\def\fcolorc{red}
\def\falpha{30}
\def\zalpha{20}
\def\kacolor{orange}
\def\kbcolor{magenta}
\def\kccolor{cyan}
\def\dalpha{70}
\def\fxshift{-220}
\def\fyshift{0}
\def\lyshift{40.0}
\def\lxshift{0.0}
\def\nF{6}
\def\kxshift{8}
\def\kyshift{1}
\def\nKa{5}
\def\nKb{3}
\def\nKc{3}
\def\rot{20}
\def\dd{0.3}
\def\xslant{0.9}
\def\yslant{-0.3}
\def\nPa{2}
\def\nPb{1}
\centering %

\resizebox{1.0\columnwidth}{!}{
\centering
\pgfdeclarelayer{fa}\pgfdeclarelayer{ka}\pgfdeclarelayer{kb}\pgfdeclarelayer{kc}
\pgfdeclarelayer{fb}\pgfdeclarelayer{afa}\pgfdeclarelayer{aka}\pgfdeclarelayer{afb}\pgfdeclarelayer{akb}\pgfdeclarelayer{afc}\pgfdeclarelayer{bfa}\pgfdeclarelayer{bka}\pgfdeclarelayer{bfb}\pgfdeclarelayer{bkb}\pgfdeclarelayer{bfc}
\pgfsetlayers{fa,ka,kb,kc,fb, afa, aka, afb, akb, afc, bfa,bka,bfb,bkb,bfc}  
\begin{tikzpicture}[decoration={brace}][scale=1,every node/.style={minimum size=1cm},on grid]
\begin{pgfonlayer}{fa}
       \begin{scope}[  
             every node/.append style={yslant=0,xslant=0,rotate=0},yslant=\yslant,xslant=\xslant,rotate=\rot
          ]
          \coordinate (faA) at (0,0);
          \coordinate (faB) at (0,\dd*\nF+2*\dd*\nPa);
          \coordinate (faC) at (\dd*\nF+2*\dd*\nPa,0);
          \coordinate (faD) at (\dd*\nF+2*\dd*\nPa,\dd*\nF+2*\dd*\nPa);
          \fill[\zcolor!\zalpha,fill, opacity=.7] (faA) rectangle (faD);
          \fill[\fcolor!\falpha,fill, opacity=.7] (faA) +(\nPa*\dd,\nPa*\dd) rectangle ++ (\nF*\dd + \nPa*\dd,\nF*\dd+\nPa*\dd);          
          \draw[step=\dd cm,blue!60] (faA) grid (faD);
          \draw[black!100,thick] (faA) rectangle (faD);
          
    \end{scope}
\end{pgfonlayer}
\begin{pgfonlayer}{kb}
       \begin{scope}[
             xshift=1.5*\lxshift,yshift=1.5*\lyshift,every node/.append style={yslant=0,xslant=0,rotate=0},yslant=\yslant,xslant=\xslant,rotate=\rot
          ]
        \coordinate (kbA) at (\kxshift*\dd ,\kyshift*\dd);
        \coordinate (kbB) at (\kxshift*\dd ,\kyshift*\dd+\dd*\nKb);
        \coordinate (kbC) at (\kxshift*\dd + \dd*\nKb,\kyshift*\dd); 
        \coordinate (kbD) at (\kxshift*\dd + \dd*\nKb,\kyshift*\dd+\dd*\nKb);    
        \coordinate (kbbrace) at (0.5*\kxshift*\dd + 0.6*\dd*\nKa             ,\kyshift*\dd + 0.5*\dd*\nKb); 
        \fill[\kbcolor!30,fill, opacity=.7] (kbA) rectangle (kbD);
        \draw[step=\dd cm,blue!60] (kbA) grid (kbD);
        \draw[black!100,thick] (kbA) rectangle (kbD);
    \end{scope}
\end{pgfonlayer}
\begin{pgfonlayer}{ka}
       \begin{scope}[
             xshift=2*\lxshift,yshift=2*\lyshift,every node/.append style={yslant=0,xslant=0,rotate=0},yslant=\yslant,xslant=\xslant,rotate=\rot
          ]
            \coordinate (kaA) at (0,0);
            \coordinate (kaB) at (0,\dd*\nKa);
            \coordinate (kaC) at (\dd*\nKa,0);
            \coordinate (kaD) at (\dd*\nKa,\dd*\nKa);    
            \draw[very thick,dotted,black!\dalpha] (kaA) -- (faA);
            \draw[very thick,dotted,black!\dalpha] (kaD) -- ($(faA) + (\dd*\nKa, \dd*\nKa)$);
            \draw[very thick,dotted,black!\dalpha] (kaB) -- ($(faA) + (0, \dd*\nKa)$);
            \draw[very thick,dotted,black!\dalpha] (kaC) -- ($(faA) + (\dd*\nKa, 0)$);
            \fill[\kacolor!30,fill, opacity=.7] (kaA) rectangle (kaD);
            \draw[step=\dd cm,blue!60] (kaA) grid (kaD);
            \draw[black!100,thick] (kaA) rectangle (kaD);
            
            \node (convop) at (\kxshift*\dd + 1.5*\dd , \kyshift*\dd +\dd) {\Large$\circledast$};
            \node (text) at (-2*\dd,0) {\large $\theta_{l\!+\!1} \circledast \theta_l$};
    \end{scope}
\end{pgfonlayer}
\begin{pgfonlayer}{kc}
       \begin{scope}[
             xshift=2.5*\lxshift,yshift=2.5*\lyshift,every node/.append style={yslant=0,xslant=0,rotate=0},yslant=\yslant,xslant=\xslant,rotate=\rot
          ]
            \coordinate (kcA) at (\kxshift*\dd ,\kyshift*\dd);
            \coordinate (kcB) at ( \kxshift*\dd ,\kyshift*\dd+\dd*\nKb);
            \coordinate (kcC) at (\kxshift*\dd + \dd*\nKb, \kyshift*\dd);
            \coordinate (kcD) at (\kxshift*\dd + \dd*\nKb,\kyshift*\dd+\dd*\nKb);  
            \coordinate (kcbrace) at (0.5*\kxshift*\dd + 0.6*\dd*\nKa             ,\kyshift*\dd + 0.5*\dd*\nKb); 
            \draw [decorate, decoration = {calligraphic brace}, very thick] (kbbrace) --  (kcbrace);
            \fill[\kccolor!30,fill, opacity=.7] (kcA) rectangle (kcD);
            \draw[step=\dd cm,blue!60] (kcA) grid (kcD);
            \draw[black!100,thick] (kcA) rectangle (kcD);
    \end{scope}
\end{pgfonlayer}
\begin{pgfonlayer}{fb}
       \begin{scope}[
             xshift=4*\lxshift,yshift=4*\lyshift,every node/.append style={yslant=0,xslant=0,rotate=0},yslant=\yslant,xslant=\xslant,rotate=\rot
          ]
            \coordinate (fbA) at (0,0);
            \coordinate (fbB) at (0,\dd*\nF);
            \coordinate (fbC) at (\dd*\nF,0);
            \coordinate (fbD) at (\dd*\nF,\dd*\nF);
            \draw[very thick,dotted,black!\dalpha] (kaB) -- ($(fbA) + (\dd*0.5, \dd*0.5)$);
            \draw[very thick,dotted,black!\dalpha] (kaC) -- ($(fbA) + (\dd*0.5, \dd*0.5)$);
            \draw[very thick,dotted,black!\dalpha] (kaA) -- ($(fbA) + (\dd*0.5, \dd*0.5)$);
            \draw[very thick,dotted,black!\dalpha] (kaD) -- ($(fbA) + (\dd*0.5, \dd*0.5)$);
            \fill[\fcolorb!\falpha,fill, opacity=.7] (fbA) rectangle (fbD);
            \draw[step=\dd cm,blue!60] (fbA) grid (fbD);
            \draw[black!100,thick] (fbA) rectangle (fbD);
    \end{scope}
\end{pgfonlayer}

\begin{pgfonlayer}{afa}
       \begin{scope}[  
       xshift=\fxshift,yshift=\fyshift,
             every node/.append style={yslant=0,xslant=0,rotate=0},yslant=\yslant,xslant=\xslant,rotate=\rot
          ]
          \coordinate (faA) at (0,0);
          \coordinate (faB) at (0,\dd*\nF+2*\nPa*\dd);
          \coordinate (faC) at (\dd*\nF+2*\nPa*\dd,0);
          \coordinate (faD) at (\dd*\nF+2*\nPa*\dd,\dd*\nF+2*\nPa*\dd);
          \fill[\zcolor!\zalpha,fill, opacity=.7] (faA) rectangle (faD);
          \fill[\fcolor!\falpha,fill, opacity=.7] (faA) +(\nPa*\dd,\nPa*\dd) rectangle ++ (\nF*\dd+\nPa*\dd,\nF*\dd+\nPa*\dd);
          \draw[step=\dd cm,blue!60] (faA) grid (faD);
          \draw[black!100,thick] (faA) rectangle (faD);
    \end{scope}
\end{pgfonlayer}
\begin{pgfonlayer}{aka}
       \begin{scope}[
             xshift=\fxshift+\lxshift,yshift=\fyshift+\lyshift,every node/.append style={yslant=0,xslant=0,rotate=0},yslant=\yslant,xslant=\xslant,rotate=\rot
          ]
        \coordinate (kaA) at (0,0);
        \coordinate (kaB) at (0,\dd*\nKb);
        \coordinate (kaC) at (\dd*\nKb,0);
        \coordinate (kaD) at (\dd*\nKb,\dd*\nKb);    
        \draw[very thick,dotted,black!\dalpha] (kaA) -- (faA);
        \draw[very thick,dotted,black!\dalpha] (kaD) -- ($(faA) + (\dd*\nKb, \dd*\nKb)$);
        \draw[very thick,dotted,black!\dalpha] (kaB) -- ($(faA) + (0, \dd*\nKb)$);
        \draw[very thick,dotted,black!\dalpha] (kaC) -- ($(faA) + (\dd*\nKb, 0)$);
        \fill[\kbcolor!30,fill, opacity=.7] (kaA) rectangle (kaD);
        \draw[step=\dd cm,blue!60] (kaA) grid (kaD);
        \draw[black!100,thick] (kaA) rectangle (kaD);
    \end{scope}
\end{pgfonlayer}
\begin{pgfonlayer}{afb}
       \begin{scope}[
        xshift=\fxshift+2*\lxshift,yshift=\fyshift+2*\lyshift,every node/.append style={yslant=0,xslant=0,rotate=0},yslant=\yslant,xslant=\xslant,rotate=\rot
          ]
        \coordinate (fbA) at (0,0);
        \coordinate (fbB) at (0,\dd*\nF+2*\nPb*\dd);
        \coordinate (fbC) at (\dd*\nF+2*\nPb*\dd,0);
        \coordinate (fbD) at (\dd*\nF+2*\nPb*\dd,\dd*\nF+2*\nPb*\dd);
        \draw[very thick,dotted,black!\dalpha] (kaB) -- ($(fbA) + (\dd*0.5, \dd*0.5)$);
        \draw[very thick,dotted,black!\dalpha] (kaC) -- ($(fbA) + (\dd*0.5, \dd*0.5)$);
        \draw[very thick,dotted,black!\dalpha] (kaA) -- ($(fbA) + (\dd*0.5, \dd*0.5)$);
        \draw[very thick,dotted,black!\dalpha] (kaD) -- ($(fbA) + (\dd*0.5, \dd*0.5)$);
        \fill[\fcolorc!\falpha,fill, opacity=.7] (fbA) rectangle (fbD);
        \draw[step=\dd cm,blue!60] (fbA) grid (fbD);
        \draw[black!100,thick] (fbA) rectangle (fbD);
    \end{scope}
\end{pgfonlayer}

\begin{pgfonlayer}{akb}
       \begin{scope}[
             xshift=\fxshift+3*\lxshift,yshift=\fyshift+3*\lyshift,every node/.append style={yslant=0,xslant=0,rotate=0},yslant=\yslant,xslant=\xslant,rotate=\rot
          ]
        \coordinate (kbA) at (0,0);
        \coordinate (kbB) at (0,\dd*\nKc);
        \coordinate (kbC) at (\dd*\nKc,0);
        \coordinate (kbD) at (\dd*\nKc,\dd*\nKc);    
        \draw[very thick,dotted,black!\dalpha] (kbA) -- (fbA);
        \draw[very thick,dotted,black!\dalpha] (kbD) -- ($(fbA) + (\dd*\nKc, \dd*\nKc)$);
        \draw[very thick,dotted,black!\dalpha] (kbB) -- ($(fbA) + (0, \dd*\nKc)$);
        \draw[very thick,dotted,black!\dalpha] (kbC) -- ($(fbA) + (\dd*\nKc, 0)$);  
        
        \fill[\kccolor!30,fill, opacity=.7] (kbA) rectangle (kbD);
        \draw[step=\dd cm,blue!60] (kbA) grid (kbD);
        \draw[black!100,thick] (kbA) rectangle (kbD);
    \end{scope}
\end{pgfonlayer}

\begin{pgfonlayer}{afc}
       \begin{scope}[
             xshift=\fxshift+4*\lxshift,yshift=\fyshift+4*\lyshift,every node/.append style={yslant=0,xslant=0,rotate=0},yslant=\yslant,xslant=\xslant,rotate=\rot
          ]
            \coordinate (fcA) at (0,0);
            \coordinate (fcB) at (0,\dd*\nF);
            \coordinate (fcC) at (\dd*\nF,0);
            \coordinate (fcD) at (\dd*\nF,\dd*\nF);
            \draw[very thick,dotted,black!\dalpha] (kbB) -- ($(fcA) + (\dd*0.5, \dd*0.5)$);
            \draw[very thick,dotted,black!\dalpha] (kbC) -- ($(fcA) + (\dd*0.5, \dd*0.5)$);
            \draw[very thick,dotted,black!\dalpha] (kbA) -- ($(fcA) + (\dd*0.5, \dd*0.5)$);
            \draw[very thick,dotted,black!\dalpha] (kbD) -- ($(fcA) + (\dd*0.5, \dd*0.5)$);
            \fill[\fcolorb!\falpha,fill, opacity=.7] (fcA) rectangle (fcD);
            \draw[step=\dd cm,blue!60] (fcA) grid (fcD);
            \draw[black!100,thick] (fcA) rectangle (fcD);

    \end{scope}
\end{pgfonlayer}
\begin{pgfonlayer}{bfa}
       \begin{scope}[  
       xshift=2*\fxshift,yshift=\fyshift,
             every node/.append style=          {yslant=0,xslant=0,rotate=0},yslant=\yslant,xslant=\xslant,rotate=\rot
          ]
          \coordinate (faA) at (0,0);
          \coordinate (faB) at (0,\dd*\nF+2*\nPb*\dd);
          \coordinate (faC) at (\dd*\nF+2*\nPb*\dd,0);
          \coordinate (faD) at (\dd*\nF+2*\nPb*\dd,\dd*\nF+2*\nPb*\dd);
          \fill[\zcolor!\zalpha,fill, opacity=.7] (faA) rectangle (faD);
          \fill[\fcolor!\falpha,fill, opacity=.7] (faA) +(\nPb*\dd,\nPb*\dd) rectangle ++ (\nF*\dd+\nPb*\dd,\nF*\dd+\nPb*\dd);
          \draw[step=\dd cm,blue!60] (faA) grid (faD);
          \draw[black!100,thick] (faA) rectangle (faD);
          \node (text) at (-2*\dd,0) {\large $X^{(l)}$};
    \end{scope}
\end{pgfonlayer}
\begin{pgfonlayer}{bka}
       \begin{scope}[
             xshift=2*\fxshift+\lxshift,yshift=\fyshift+\lyshift,every node/.append style={yslant=0,xslant=0,rotate=0},yslant=\yslant,xslant=\xslant,rotate=\rot
          ]
            \coordinate (kaA) at (0,0);
            \coordinate (kaB) at (0,\dd*\nKb);
            \coordinate (kaC) at (\dd*\nKb,0);
            \coordinate (kaD) at (\dd*\nKb,\dd*\nKb);    
            
            \draw[very thick,dotted,black!\dalpha] (kaA) -- (faA);
            \draw[very thick,dotted,black!\dalpha] (kaD) -- ($(faA) + (\dd*\nKb, \dd*\nKb)$);
            \draw[very thick,dotted,black!\dalpha] (kaB) -- ($(faA) + (0, \dd*\nKb)$);
            \draw[very thick,dotted,black!\dalpha] (kaC) -- ($(faA) + (\dd*\nKb, 0)$);
            \fill[\kbcolor!30,fill, opacity=.7] (kaA) rectangle (kaD);
            \draw[step=\dd cm,blue!60] (kaA) grid (kaD);
            \draw[black!100,thick] (kaA) rectangle (kaD);
            
            \node (text) at (-2*\dd, 0){\large $\theta_l$};
    \end{scope}
\end{pgfonlayer}
\begin{pgfonlayer}{bfb}
       \begin{scope}[
             xshift=2*\fxshift+2*\lxshift,yshift=\fyshift+2*\lyshift,every node/.append style={yslant=0,xslant=0,rotate=0},yslant=\yslant,xslant=\xslant,rotate=\rot
          ]
            \coordinate (fbA) at (0,0);
            \coordinate (fbB) at (0,\dd*\nF+2*\nPb*\dd);
            \coordinate (fbC) at (\dd*\nF+2*\nPb*\dd,0);
            \coordinate (fbD) at (\dd*\nF+2*\nPb*\dd,\dd*\nF+2*\nPb*\dd);
            
            \draw[very thick,dotted,black!\dalpha] (kaB) -- ($(fbA) + (\dd*1.5, \dd*1.5)$);
            \draw[very thick,dotted,black!\dalpha] (kaC) -- ($(fbA) + (\dd*1.5, \dd*1.5)$);
            \draw[very thick,dotted,black!\dalpha] (kaA) -- ($(fbA) + (\dd*1.5, \dd*1.5)$);
            \draw[very thick,dotted,black!\dalpha] (kaD) -- ($(fbA) + (\dd*1.5, \dd*1.5)$);
            \fill[\zcolor!\zalpha,fill, opacity=.7] (fbA) rectangle (fbD);
            \fill[\fcolora!\falpha,fill, opacity=.7] (fbA) +(\nPb*\dd,\nPb*\dd) rectangle ++ (\nF*\dd+\nPb*\dd,\nF*\dd+\nPb*\dd);
            \draw[step=\dd cm,blue!60] (fbA) grid (fbD);
            \draw[black!100,thick] (fbA) rectangle (fbD);
            \node (text) at (-2*\dd,0) {\large $X^{(l+1)}$};
    \end{scope}
\end{pgfonlayer}

\begin{pgfonlayer}{bkb}
       \begin{scope}[
             xshift=2*\fxshift+3*\lxshift,yshift=\fyshift+3*\lyshift,every node/.append style={yslant=0,xslant=0,rotate=0},yslant=\yslant,xslant=\xslant,rotate=\rot
          ]
          \coordinate (kbA) at (0,0);
          \coordinate (kbB) at (0,\dd*\nKc);
          \coordinate (kbC) at (\dd*\nKc,0);
          \coordinate (kbD) at (\dd*\nKc,\dd*\nKc);    
        \draw[very thick,dotted,black!\dalpha] (kbA) -- (fbA);
        \draw[very thick,dotted,black!\dalpha] (kbD) -- ($(fbA) + (\dd*\nKc, \dd*\nKc)$);
        \draw[very thick,dotted,black!\dalpha] (kbB) -- ($(fbA) + (0, \dd*\nKc)$);
         \draw[very thick,dotted,black!\dalpha] (kbC) -- ($(fbA) + (\dd*\nKc, 0)$);
       
          \fill[\kccolor!30,fill, opacity=.7] (kbA) rectangle (kbD);
          \draw[step=\dd cm,blue!60] (kbA) grid (kbD);
        \draw[black!100,thick] (kbA) rectangle (kbD);
          \node (text) at (-2*\dd, 0) {\large $\theta_{l\!+\!1}$};
    \end{scope}
\end{pgfonlayer}

\begin{pgfonlayer}{bfc}
       \begin{scope}[
             xshift=2*\fxshift+4*\lxshift,yshift=\fyshift+4*\lyshift,every node/.append style={yslant=0,xslant=0,rotate=0},yslant=\yslant,xslant=\xslant,rotate=\rot
          ]
          \coordinate (fcA) at (0,0);
          \coordinate (fcB) at (0,\dd*\nF);
          \coordinate (fcC) at (\dd*\nF,0);
          \coordinate (fcD) at (\dd*\nF,\dd*\nF);      
        \draw[very thick,dotted,black!\dalpha] (kbB) -- ($(fcA) + (\dd*0.5, \dd*0.5)$);
        \draw[very thick,dotted,black!\dalpha] (kbC) -- ($(fcA) + (\dd*0.5, \dd*0.5)$);
        \draw[very thick,dotted,black!\dalpha] (kbA) -- ($(fcA) + (\dd*0.5, \dd*0.5)$);
         \draw[very thick,dotted,black!\dalpha] (kbD) -- ($(fcA) + (\dd*0.5, \dd*0.5)$);
          
          \fill[blue!\falpha,fill, opacity=.7] (fcA) rectangle (fcD);
          \fill[\fcolorb!\falpha,fill, opacity=.7] (fcA) + (\dd, \dd) rectangle ++ (\nF*\dd-\dd,\nF*\dd-\dd);
          \draw[step=\dd cm,blue!60] (fcA) grid (fcD);
          \draw[black!100,thick] (fcA) rectangle (fcD);

         \node (text) at (-2*\dd,0) {\large $X^{(l+2)}$};
    \end{scope}
\end{pgfonlayer}
\end{tikzpicture}
}
\caption{Illustration comparing the output from two different types of padding applied to two consecutive $3\times 3$ convolution operations with the output from a merged $5\times 5$ convolution operation. The boundary of the output feature map obtained from applying zero padding of size $1$ before each $3\times 3$ convolution is distinct from that of the output feature map obtained from the merged $5\times 5$ convolution. Conversely, if zero padding of size $2$ is applied to the first $3\times 3$ convolution, the output feature map is equivalent to the output feature map obtained from the merged $5\times 5$ convolution.
} 
\label{fig:convmerge}
\end{figure*}

Consider a feature map $X^{(l)}$, upon which two consecutive $3\times 3$ convolution operations, utilizing kernels $\theta_l$ and $\theta_{l\!+\!1}$, are applied to produce an output feature map $X^{(l+2)}$. The output generated by the first convolution operation utilizing kernel $\theta_l$ is denoted as $X^{(l+1)}$. As shown in \Cref{fig:convmerge}, when zero padding of size $1$ is applied prior to each of the $3\times3$ convolution operations, the boundary of the output resulting from the merged $5\times 5$ convolution operation, utilizing kernel $\theta_{l\!+\!1} \circledast \theta_l$ differs from that of $X^{(l+2)}$. Insufficient zero padding results in a computation skip at the boundary of $X^{(l+1)}$, which in turn leads to a discrepancy between the computation at the boundary of $X^{(l+2)}$ and the output feature map of the merged $5\times 5$ convolution operation. Conversely, when zero padding of size $2$ is applied prior to the first $3\times 3$ convolution operation, the output feature map of the two consecutive $3\times 3$ convolution operations is equivalent to the output feature map of the merged $5\times 5$ convolution operation.

In light of this, after we optimize the optimal ordered set $A$ and $S$, we fix the activation layers following $A$ and reorder the padding of the convolutional layers according to $S$ before the finetuning process. After finetuning, we merge the network at the test time without losing any accuracy.



\end{document}